\documentclass{article}

\PassOptionsToPackage{numbers}{natbib}
\usepackage[preprint]{neurips_2026}
\usepackage[numbers]{natbib}
\usepackage{amsmath,amssymb}
\usepackage{enumitem}
\usepackage{booktabs}
\usepackage{multirow}
\usepackage{adjustbox}

\usepackage[utf8]{inputenc} 
\usepackage[T1]{fontenc}    
\usepackage{hyperref}       
\usepackage{url}            
\usepackage{booktabs}       
\usepackage{amsfonts}       
\usepackage{nicefrac}       
\usepackage{microtype}      
\usepackage{xcolor}         
\usepackage{booktabs}
\usepackage{graphicx}
\usepackage{algorithm}
\usepackage{algpseudocode}
\usepackage{graphicx}
\usepackage{caption}
\usepackage{algorithm}
\usepackage{algpseudocode}
\usepackage{wrapfig}
\usepackage{booktabs}
\usepackage{multirow}
\usepackage{graphicx}
\usepackage{xcolor}
\usepackage{makecell}
\usepackage{booktabs}
\usepackage{wrapfig}

\newcommand{\score}[2]{\ensuremath{#1_{\scriptscriptstyle(#2)}}}
\newcommand{\posval}[1]{\textcolor{green!60!black}{\ensuremath{#1}}}
\newcommand{\negval}[1]{\textcolor{red!70!black}{\ensuremath{#1}}}
\newcommand{\neuval}[1]{\ensuremath{#1}}

\title{Pretraining Multiple Instance Learning Networks with Multi-Teacher Distillation from Pathology Slide Foundation Models}

\author{
Mingxi Fu$^{1,*}$ $\quad$
Jiawen Li$^{1,*}$ $\quad$
Renao Yan$^{2}$ $\quad$
Jiali Hu$^{3}$ \\
\textbf{
Qiehe Sun$^{4}$ $\quad$
Tian Guan$^{1,\dagger}$ $\quad$
Yonghong He$^{1,4,5,\dagger}$
}\\
  $^1$Tsinghua Shenzhen International Graduate School, Tsinghua University\\
  $^2$University of Washington$\ \ $
  $^3$City University of Hong Kong (Dongguan) $\ \ $ \\
  $^4$Medical Optical Technology R\&D Center, Research Institute of Tsinghua $\ \ $\\
  $^5$Jinfeng Laboratory\\
  \texttt{guantian@sz.tsinghua.edu.cn} $\quad$ \texttt{heyh@sz.tsinghua.edu.cn}\\
$^{*}$Equal contribution $\quad$ $^{\dagger}$Corresponding authors
}

\begin{document}

\maketitle

\begin{abstract}
  Multiple instance learning (MIL) has become the main paradigm for whole-slide image (WSI) analysis in computational pathology. However, existing MIL aggregators are still typically trained from scratch for each downstream task, relying on limited slide-level labels to learn both aggregation mechanisms and downstream discriminative representations simultaneously. As a result, they often suffer from unstable optimization, overfitting, and limited transferability. Similar to pretrained ResNet and Vision Transformer models in natural image learning, MIL also requires reusable pretrained initialization. However, high-quality slide-level pretraining data remain scarce, and MIL models are usually lightweight and weakly supervised, making large-scale pretraining difficult in practice. To address this challenge, we propose a distillation-based pretraining framework for MIL, which leverages two slide-level foundation models, TITAN and CARE, as teachers to transfer their representational knowledge into a diverse set of MIL architectures. To effectively balance supervision from different teachers, we further introduce an angular dispersion normalized distillation loss. The distilled weights are then used as initialization for downstream adaptation. We conduct systematic evaluations on 15 benchmark datasets under both linear probing and full-parameter fine-tuning, and further validate its advantages in few-shot scenarios. Experimental results show that pretraining generally improves MIL aggregators over from scratch training, especially in linear-probing and few-shot settings, while maintaining the computational efficiency of lightweight MIL models. Code is available at \url{https://github.com/fu0201/MIL_Pretrained}.
\end{abstract}

\section{Introduction}
\label{sec:intro}

Multiple instance learning (MIL) has been widely adopted as the mainstream framework for whole-slide image (WSI) analysis in computational pathology. Under this framework, patch features produced by a frozen tile encoder are treated as instances, and the entire slide is modeled as a bag whose label supervises learning~\cite{xu2025mil_foundation}.
Driven by this formulation, a rich spectrum of MIL aggregators has emerged, including attention-based pooling~\cite{ilse2018attention,lu2021clam}, Transformer-based architectures~\cite{shao2021transmil}, graph neural networks~\cite{li2024wikg,fu2025dagrl}, and state-space models~\cite{zhang20252dmamba}. These methods have continuously advanced instance aggregation and slide-level generalization. Despite this architectural diversity, a fundamental limitation persists: existing MIL aggregators are invariably trained from scratch on each downstream task, relying solely on a handful of slide-level labels to simultaneously learn both discriminative instance representations and an effective aggregation strategy.
Under such weak supervision and limited data, aggregators frequently overfit to spurious correlations and fail to acquire stable, transferable slide representations~\cite{xu2025mil_foundation,shao2025miltransfer}.

\begin{wrapfigure}{r}{0.5\textwidth}
    \vspace{-0.5em}
    \centering
    \includegraphics[width=0.5\textwidth]{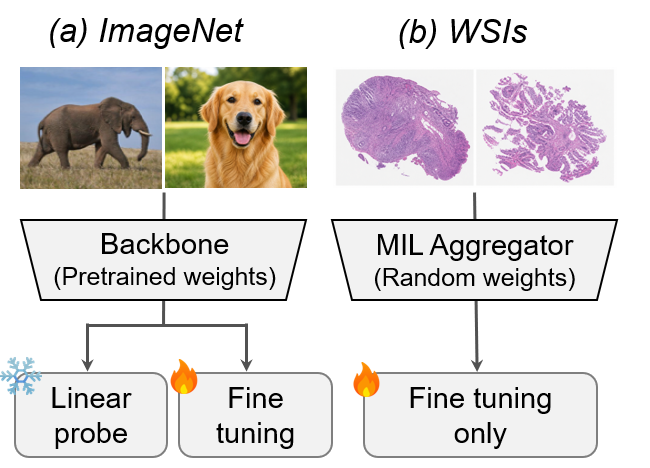}
    \caption{Pretrained backbones support freezing or fine-tuning in natural image analysis, whereas MIL models for WSI are typically trained from scratch.}
    \label{fig:picfirst}
    \vspace{-0.8em}
\end{wrapfigure}

A natural solution is to pretrain MIL aggregators before adapting them to downstream WSI tasks. In modern computer vision, pretraining followed by downstream fine-tuning has become the standard paradigm for visual representation learning. Representative backbones such as ResNet~\cite{he2016deep} and Vision Transformer (ViT)~\cite{dosovitskiy2021image} are typically pretrained on large-scale image corpora and then adapted to downstream tasks through fine-tuning or linear-probing as shown in Figure~\ref{fig:picfirst}. Studies have shown that such pretrained initialization substantially improves transferability, convergence speed, and data efficiency, thereby providing a strong and reusable starting point for a wide range of vision tasks~\cite{kornblith2019better,kolesnikov2020big}. This paradigm has increasingly reshaped computational pathology as well. At the tile level, self-supervised and vision--language foundation models, such as UNI~\cite{chen2024uni}, Virchow~\cite{vorontsov2024virchow}, CONCH~\cite{lu2024conch}, and MUSK~\cite{xiang2025musk}, have dramatically improved patch-level feature quality through pretraining on large-scale histopathology images. More recently, at the slide level, models such as TITAN~\cite{ding2025titan}, Prov-GigaPath~\cite{xu2024provgigapath}, and CARE~\cite{zhang2026care} have further demonstrated that pretraining entire slide encoders on large-scale high-resolution pathology images can yield powerful and general-purpose slide representations. Beyond foundation models themselves, Shao et al.~\cite{shao2025miltransfer} also provided direct evidence that pretrained MIL aggregators consistently outperform their randomly initialized counterparts in downstream transfer.

However, directly pretraining MIL aggregators still faces several challenges. 
First, large-scale slide-level datasets remain intrinsically scarce. Unlike low-resolution tile images, which can be densely cropped from WSIs and thereby scaled to hundreds of millions or even billions of samples, publicly available WSI cohorts are typically limited to thousands or tens of thousands of slides, far below the data scale required for robust foundation-model pretraining~\cite{ding2025titan,xu2024provgigapath}. Second, conventional MIL aggregators are usually designed as lightweight task-specific modules rather than high-capacity pretraining targets. Most MIL aggregators contain only a few million parameters and are mainly supervised by bag-level labels, providing insufficient optimization signals for learning generalizable slide-level representations~\cite{jing2021understanding}. Recent studies have attempted to mitigate this issue by scaling up slide encoders and pretraining them on large WSI cohorts~\cite{shaikovski2024prism}. However, in digital pathology, where downstream tasks are highly heterogeneous and annotated data are often limited, directly fine-tuning large slide encoders can be computationally expensive and optimization-unstable, thereby restricting their practical applicability across diverse clinical tasks.

To make MIL pretraining feasible, in this work we propose a multi-teacher distillation framework for pretraining MIL networks from pathology slide foundation models. Specifically, we employ TITAN~\cite{ding2025titan} and CARE~\cite{zhang2026care}, two state-of-the-art pretrained slide encoders as complementary expert teachers. Guided by a angular dispersion normalized distillation loss, we distill their slide-level representational knowledge into 9 MIL architectures, covering attention-based, Transformer-based, graph-based, and state-space designs. To validate the effectiveness of the pretrained MIL aggregators, we initialize downstream fine-tuning with the distillation-pretrained weights. Across six benchmark datasets and more than fifteen clinical tasks, including cancer subtyping, biomarker prediction, and mutation detection, we compare this paradigm with from-scratch MIL training and direct teacher-model fine-tuning. Our results demonstrate that, compared with from-scratch training, distillation-based pretraining enhances MIL aggregators across fine-tuning, linear-probing, and few-shot evaluation settings, and further enables these lightweight models to outperform their teacher models in many scenarios.

\section{Related Work}
\label{sec:related work}

\subsection{Multiple Instance Learning for Weakly Supervised WSI Analysis}

Multiple instance learning (MIL) has long been the dominant paradigm for weakly supervised WSI analysis. In this framework, pretrained patch encoders are used to extract local visual features, while an aggregator further integrates instance-level information to generate slide-level predictions \cite{laleh2022benchmarking,althelaya2023survey}. Early MIL methods mainly focused on permutation-invariant aggregation designs, such as the attention-based pooling in ABMIL~\cite{ilse2018attention} and the clustering-constrained attention mechanism in CLAM~\cite{lu2021clam}. To capture dependencies among instances, TransMIL~\cite{shao2021transmil} introduced self-attention together with pyramidal positional encoding, whereas graph-based methods such as WiKG~\cite{li2024wikg} and DAGMIL~\cite{fu2025dagrl} construct spatial or feature graphs to propagate contextual information across patches. More recently, state space models have also been introduced into MIL aggregation, leveraging their linear-complexity sequence modeling capability to efficiently handle the thousands of instances typically encountered in WSIs~\cite{zhang20252dmamba}. Nevertheless, most existing MIL methods still treat the aggregator as a task-specific module that must be trained from scratch on each downstream dataset using limited slide-level labels, and its cross-task transferability remains largely underexplored.

\subsection{Pretraining from patch-level to slide-level in computational pathology}

With the rapid development of self-supervised learning, pathology image pretraining has achieved substantial progress at the patch level. Methods such as SimCLR~\cite{chen2020simclr}, MoCo~\cite{he2020moco}, and DINO~\cite{caron2021dino} have been applied to millions of tiles cropped from hundreds of thousands of WSIs, leading to a series of pretrained models with different scales and capabilities \cite{kang2023benchmarking,li2021dsmil,wang2022ctranspath}. Vision Transformer backbones trained under the DINOv2~\cite{oquab2024dinov2} framework, such as UNI~\cite{chen2024uni} and Virchow~\cite{vorontsov2024virchow}, have become widely adopted image encoders in digital pathology. Meanwhile, works such as CONCH~\cite{lu2024conch} and MUSK~\cite{xiang2025musk} further explored vision-language pretraining, offering pathology models with stronger cross-modal capabilities. Building upon these powerful pathology-specific encoders, recent studies have further investigated slide-level pretraining to learn general-purpose WSI representations. For example, CHIEF~\cite{wang2024pathology} combines tile-level representation learning with weakly supervised slide-level pretraining to capture generalizable WSI patterns. MADELEINE~\cite{jaume2024madeleine} introduces a multi-stain pretraining strategy, where matched slides from different staining modalities are aligned through global-local cross-stain objectives. TITAN~\cite{ding2025titan} further extends slide-level pretraining to the multimodal setting by aligning WSI representations with pathology reports and synthetic captions. CARE~\cite{zhang2026care} incorporates molecular guidance by first learning morphological representations from WSIs and then refining adaptive region representations through RNA and protein alignment. However, MIL aggregators are typically lightweight, which limits their capacity to benefit from conventional supervised pretraining in the same way as large-scale encoders. Recent work such as FEATHER~\cite{shao2025miltransfer} has attempted to pretrain MIL models with supervised slide-level classification labels, but this paradigm requires large-scale labeled datasets, making MIL pretraining costly and difficult to scale to broader pathology tasks.

\subsection{Knowledge Distillation and Multi-teacher Knowlodege distillation}

Knowledge distillation provides an effective mechanism for transferring knowledge from well-trained teacher models to compact student models, commonly through soft-logit supervision or feature-level representation alignment~\cite{hinton2015distilling,wei2022featuredistillation,ahn2019vid,beyer2022patient,mirzadeh2020teacherassistant}. Multi-teacher distillation further extends this paradigm by allowing a student model to inherit complementary knowledge from multiple heterogeneous teachers~\cite{liu2020amtmlkd,park2020feed}, which is particularly useful for foundation models with diverse architectures and representation spaces. Recent vision foundation models have validated the effectiveness of this strategy. AM-RADIO~\cite{ranzinger2024amradio} distills multiple powerful vision models, including CLIP~\cite{radford2021clip}, DINOv2~\cite{oquab2024dinov2}, and SAM~\cite{kirillov2023sam}, into a unified student representation, while C-RADIOv4~\cite{ranzinger2026cradiov4} further improves representation compatibility and transferability. In computational pathology, GPFM~\cite{ma2026gpfm} similarly distills multiple expert encoders into a unified pathology foundation model to improve generalization across downstream tasks. Despite these advances, existing pathology distillation studies mainly focus on low-resolution images encoders, while high-resolution whole-slide representation learning remain largely underexplored.

\section{Method}
\label{sec:method}

\begin{figure}[t]
    \centering
    \includegraphics[width=\linewidth]{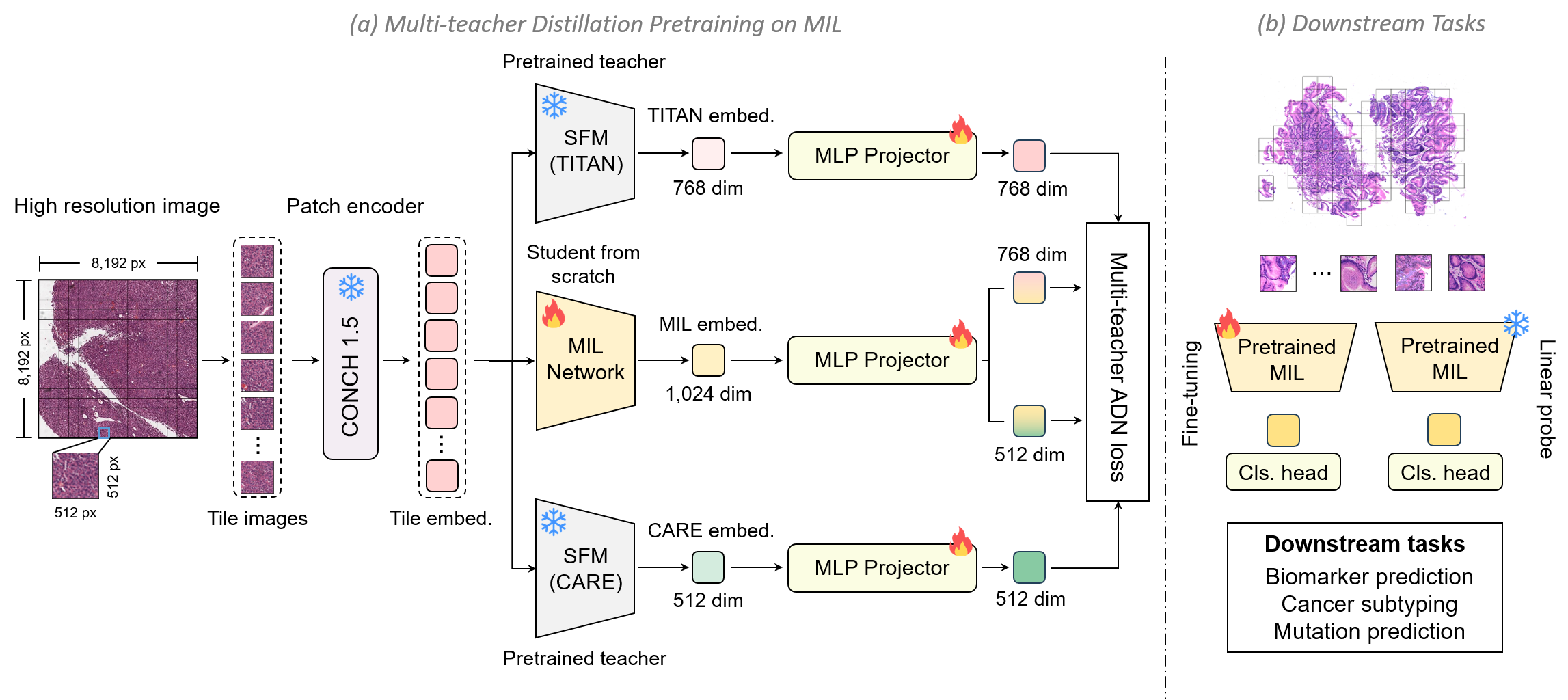}
    \caption{Overview of the proposed framework.
    Left: ROI-level pretraining of the MIL network via multi-teacher distillation from TITAN and CARE.
    Right: Transfer of the pretrained MIL aggregator to downstream WSI-level evaluation tasks.}
    \label{fig:main_figure}
\end{figure}

We propose a multi-teacher distillation pretraining method on MIL, a general framework for pretraining lightweight MIL aggregators through slide-level representation transfer. 
Figure~\ref{fig:main_figure} provides an overview of our method. We first introduce the multi-teacher distillation pretraining procedure, where MIL aggregators are pretrained on high-resolution pathology regions of interest (ROIs) under the guidance of pretrained whole-slide foundation models. We then describe how the pretrained MIL weights are transferred to downstream WSI analysis through full-parameter fine-tuning or linear probing. 

\subsection{Multi-teacher Distillation Pretraining on MIL}
\label{sec:multi_teacher_distill}

In WSI analysis, each slide is commonly represented as a bag of patch-level features. 
Given a WSI or a high-resolution pathology region, we divide it into $N$ non-overlapping patches and use a frozen patch encoder $f_{\mathrm{patch}}$ to extract patch embeddings:
\begin{equation}
    \mathbf{X} = \{\mathbf{x}_i\}_{i=1}^{N} \in \mathbb{R}^{N \times D},
\end{equation}
where $\mathbf{x}_i \in \mathbb{R}^{D}$ denotes the feature vector of the $i$-th patch. 
When spatial information is available, we also record the corresponding patch coordinates:
\begin{equation}
    \mathbf{C} = \{(r_i, c_i)\}_{i=1}^{N}.
\end{equation}

A MIL aggregator $g_\theta$ maps the bag of patch features, optionally together with spatial coordinates, into a slide-level representation:
\begin{equation}
    \mathbf{z} = g_\theta(\mathbf{X}, \mathbf{C}) \in \mathbb{R}^{d}.
\end{equation}
For downstream classification, $\mathbf{z}$ is further passed into a task-specific prediction head $h_\psi$:
\begin{equation}
    \hat{\mathbf{y}} = h_\psi(\mathbf{z}).
\end{equation}

Large-scale data are essential for pretraining MIL aggregators. However, directly using full WSIs for batch-level MIL pretraining is computationally inconvenient because different WSIs contain highly variable numbers of patches, resulting in irregular bag sizes and unstable memory consumption. Inspired by the high-resolution image pretraining strategy used in TITAN~\citep{ding2025titan}, we instead construct the pretraining data from extremely high-resolution pathology ROIs. Specifically, we use TCGA-UT-8K~\citep{ding2025titan} as pretraining data. Each ROI is divided into non-overlapping $512 \times 512$ patches, yielding a regular $16 \times 16$ grid of 256 patches per ROI. For each patch, we extract a 768-dimensional feature vector using CONCH v1.5~\citep{lu2024conch}. 

\paragraph{Teacher models.}
We select TITAN~\citep{ding2025titan} and CARE~\citep{zhang2026care} as teacher models because they provide complementary sources of slide-level knowledge. 
TITAN~\citep{ding2025titan} offers general-purpose and transferable whole-slide semantics learned from large-scale multimodal pretraining, whereas CARE~\citep{zhang2026care} provides morphology-aware and molecularly informed regional representations. Their combination provides diverse supervisory signals and reduces the bias introduced by relying on a single teacher representation.

Given patch features $\mathbf{X}$ and spatial coordinates $\mathbf{C}$, TITAN~\citep{ding2025titan} first arranges patch features into a 2D spatial grid and produces a pooled slide-level embedding:
\begin{equation}
    \mathbf{t}^{\mathrm{titan}} = T_{\mathrm{titan}}(\mathbf{X}, \mathbf{C}) 
    \in \mathbb{R}^{D_{\mathrm{t}}},
    \qquad D_{\mathrm{t}} = 768.
\end{equation}
Similarly, CARE~\citep{zhang2026care} constructs morphologically relevant adaptive regions from the input bag and outputs a slide-level embedding:
\begin{equation}
    \mathbf{t}^{\mathrm{care}} = T_{\mathrm{care}}(\mathbf{X}, \mathbf{C})
    \in \mathbb{R}^{D_{\mathrm{c}}},
    \qquad D_{\mathrm{c}} = 512.
\end{equation}
During distillation pretraining, both teacher models are kept frozen.

\paragraph{Student models.}
The student is a MIL aggregator $g_\theta$ with its classification head removed, so that it only outputs a slide-level embedding:
\begin{equation}
    \mathbf{z} = g_\theta(\mathbf{X}, \mathbf{C}) \in \mathbb{R}^{d}.
\end{equation}
Since TITAN~\citep{ding2025titan} and CARE~\citep{zhang2026care} have different representation dimensions, we attach two learnable teacher-specific projection heads to map the student embedding into each teacher space:
\begin{equation}
    \hat{\mathbf{t}}^{\mathrm{titan}} 
    = \phi_{\mathrm{titan}}(\mathbf{z})
    \in \mathbb{R}^{D_{\mathrm{t}}},
    \qquad
    \hat{\mathbf{t}}^{\mathrm{care}} 
    = \phi_{\mathrm{care}}(\mathbf{z})
    \in \mathbb{R}^{D_{\mathrm{c}}},
\end{equation}
where 
$\phi_{\mathrm{titan}}: \mathbb{R}^{d} \rightarrow \mathbb{R}^{D_{\mathrm{t}}}$ 
and 
$\phi_{\mathrm{care}}: \mathbb{R}^{d} \rightarrow \mathbb{R}^{D_{\mathrm{c}}}$ 
are learnable linear projection heads. 
The projected student embeddings are then aligned with the corresponding frozen teacher embeddings during multi-teacher distillation.

\paragraph{Angular Dispersion Normalized Distillation Loss.}

A central challenge in multi-teacher distillation is that different teachers may exhibit different concentration levels in their representation distributions on the hypersphere. If one teacher produces more tightly clustered embeddings, the scale of its supervision signal may differ systematically from that of another teacher, leading to imbalanced optimization across teachers. To address this issue, inspired by c-RADIOv4~\cite{ranzinger2026cradiov4}, we propose the Angular Dispersion Normalized Distillation Loss (ADN Loss), whose key idea is to normalize each teacher-specific alignment error by the running angular dispersion of that teacher's representation distribution.

For a batch of $B$ slides, let $\hat{\mathbf{t}}_i$ and $\mathbf{t}_i$ denote the $\ell_2$-normalized student prediction and teacher target for teacher $k$, respectively. The student-teacher discrepancy is measured by angular distance:
\begin{equation}
    \Theta_i^{(k)}
    =
    \arccos\!\left(
    \mathrm{clip}\!\left(
    (\hat{\mathbf{t}}_i^{(k)})^\top {\mathbf{t}}_i^{(k)},
    -1+\epsilon,
    1-\epsilon
    \right)
    \right),
\end{equation}
where $\epsilon$ is a small constant for numerical stability.

Different teachers may exhibit different angular variances in their representation spaces, which may cause one teacher to dominate the optimization. 
We therefore introduce a simple teacher-wise normalization strategy. 
For each teacher branch $k$, we estimate a running angular dispersion $\sigma_k$ from the standard deviation of teacher embeddings around their update center direction on the unit hypersphere. 
The distillation loss for teacher $k$ is computed as
\begin{equation}
    \mathcal{L}_{\mathrm{distill}}^{(k)}
    =
    \frac{1}{B}
    \sum_{i=1}^{B}
    \frac{
    \Theta_i^{(k)}
    }{
    \sigma_k + \epsilon
    },
\end{equation}
where $\sigma^{(k)}$ is the running angular dispersion. It is clamped to $[\sigma_{\min}, \infty)$ with $\sigma_{\min} = 0.05$ to prevent numerical instability.

For multi-teacher distillation, the final objective is the weighted sum over all teacher branches:
\begin{equation}
\mathcal{L}_{\text{ADN}} =
\sum_{k \in \{\text{titan}, \text{care}\}} 
w_k \cdot \mathcal{L}_{\text{distill}}^{(k)}
\end{equation}
where $w_k$ is the teacher weight (set to 1 for both teachers) and $\epsilon = 10^{-6}$. By dividing each teacher's error by its angular dispersion, a teacher whose embeddings are more tightly clustered (smaller $\sigma^{(k)}$) will \emph{not} dominate the gradient, since its normalized error is rescaled to a comparable magnitude as that of the more dispersed teacher. Detailed information of algorithm are provided in Appendix~\ref{sec:appendix2}.

\subsection{Using Pretrained Weights for Downstream Tasks}
\label{sec:downstream_use}

After multi-teacher distillation pretraining, we discard the teacher-specific projection heads and retain only the pretrained MIL aggregator weights $\theta^\ast$. 
These weights serve as reusable initialization for downstream WSI analysis. 
Given a downstream dataset, we load $\theta^\ast$ into the corresponding MIL architecture and attach a task-specific linear classification head $h_\psi$. 
The pretrained MIL model can then be used in two standard adaptation settings.

\paragraph{Full-parameter fine-tuning.}
In this setting, the pretrained aggregator is used as initialization, and both the MIL aggregator and the task-specific classification head are updated with downstream slide-level labels:
\begin{equation}
    \min_{\theta,\psi}
    \mathcal{L}_{\mathrm{CE}}
    \left(
    h_\psi(g_\theta(\mathbf{X}, \mathbf{C})),
    \mathbf{y}
    \right).
\end{equation}
This protocol evaluates whether the pretrained MIL weights provide a better optimization starting point than random initialization.

\paragraph{Linear probing.}
In this setting, the pretrained MIL aggregator is frozen, and only the task-specific linear head is trained:
\begin{equation}
    \min_{\psi}
    \mathcal{L}_{\mathrm{CE}}
    \left(
    h_\psi(g_{\theta^\ast}(\mathbf{X}, \mathbf{C})),
    \mathbf{y}
    \right).
\end{equation}
This protocol directly evaluates the transferability of the learned slide-level representation.

\section{Experiments}

\subsection{Dataset}
\label{sec:dataset}
\paragraph{Pretraining dataset.}
During distillation pretraining, we use TCGA-UT-8K~\citep{ding2025titan}, a large-scale region-level pan-cancer pathology dataset.
It contains 25,495 ROIs of size $8{,}192 \times 8{,}192$ pixels, derived from 9,662 H\&E-stained FFPE diagnostic whole-slide images (WSIs) from TCGA.

\paragraph{Evaluation datasets.}
We evaluate our method on six public WSI cohorts covering 15 downstream tasks, including biomarker prediction, cancer subtyping, pan-cancer classification, and mutation prediction. BCNB~\cite{xu2021bcnb} is used for ER, PR, and HER2 prediction; BRACS~\cite{brancati2022bracs} for coarse- and fine-grained breast lesion subtyping; CPTAC~\cite{cptac_portal} for pan-cancer and NSCLC classification; EBRAINS~\cite{roetzer2022digital} for brain tumor subtyping and IDH mutation prediction; KIDRARE~\cite{he2026boosting} for pediatric tumor classification; and MUT-HET-RCC~\cite{acosta2022intratumoral} for BAP1, PBRM1, and SETD2 mutation prediction. For all tasks, we construct label-stratified train/validation/test splits following the same experimental protocol. Detailed dataset information is provided in Appendix~\ref{sec:appendix1}.

\subsection{Implementation Details}
\label{sec:impl}

\paragraph{Pretraining setup.}
Each ROI is partitioned into non-overlapping $512 \times 512$ patches, and each patch is encoded into a 768-dimensional feature vector using CONCH v1.5~\citep{lu2024conch}. Teacher embeddings are pre-computed and cached before training to avoid repeated inference. We jointly optimize the student aggregator and the two projection heads using Adam~\cite{kingma2015adam} with a learning rate of $5\times10^{-5}$ and weight decay of $10^{-4}$ for 500 epochs. Training is conducted on 4 NVIDIA L20 48GB GPUs using PyTorch DistributedDataParallel (DDP), with a per-GPU batch size of 1024.

\paragraph{Downstream fine-tuning setup.}
For downstream tasks, we initialize the student aggregator with the pretrained weights $\theta^*$ and attach a task-specific linear classification head. We fine-tune the whole model end-to-end using Adam~\cite{kingma2015adam} with a learning rate of $10^{-4}$ and weight decay of $10^{-4}$, 1 batch size. We apply early stopping with a patience of 5 based on validation balanced accuracy (bacc), and train for at most 50 epochs. Experiments are run on a single NVIDIA L20 48GB GPU and all baseline methods are trained under the same hyperparameters and settings.

\subsection{Linear-probing Comparison between Teacher Models and Pretrained MIL Aggregators}

Table~\ref{tab:merged_linear_probe_bacc} summarizes the test balanced accuracy of linear-probing trained on frozen slide-level representations produced by pretrained nine MIL aggregators and two teacher slide-embedding models, CARE~\citep{zhang2026care} and TITAN~\citep{ding2025titan} on six merged dataset groups covering 15 tasks. Detailed results for all 15 individual downstream tasks are provided in the Appendix~\ref{sec:appendix-linear}. Overall, the pretrained MIL aggregators achieve competitive performance under the linear-probing setting and, in many cases, outperform the teacher-level slide encoders. In particular, ABMIL~\cite{ilse2018attention}, CLAM~\cite{lu2021clam}, TransMIL~\citep{shao2021transmil}, WiKG~\cite{li2024wikg}, DAGMIL~\cite{fu2025dagrl}, and AMDMIL~\cite{ling2024agent} outperform the corresponding teacher models across all dataset groups. This suggests that, after knowledge distillation, attention-based, Transformer-based, and graph-based MIL aggregators can effectively preserve transferable slide-level representational knowledge, which can be readily exploited by a simple linear classifier. Meanwhile, GDFMIL~\cite{zhang2026rethinking} and 2DMamba~\cite{zhang20252dmamba} show relatively weaker performance on several dataset, indicating that the transferability of distilled representations is still affected by architecture-specific inductive biases. Under the frozen-feature linear-probing protocol, their representational advantages may not be fully activated.

\begin{table*}[t]
  \centering
  \caption{
  Merged-dataset linear-probing comparison between teacher models and pretrained MIL aggregators.
  Results are reported as test balanced accuracy (bacc, \%) with standard deviation shown in the lower-right subscript.
  $\Delta$ denotes the difference between each pretrained MIL aggregator and the better teacher model on the same dataset.
  }
  \label{tab:merged_linear_probe_bacc}
  \scriptsize
  \setlength{\tabcolsep}{3pt}
  \renewcommand{\arraystretch}{1.05}
  \resizebox{\linewidth}{!}{%
  \begin{tabular}{lcclccccccccc}
    \toprule
    \multirow{2}{*}{Dataset}
    & \multicolumn{2}{c}{Teacher Models}
    & \multirow{2}{*}{Metric}
    & \multicolumn{9}{c}{Pretrained Student MIL Aggregators} \\
    \cmidrule(lr){2-3} \cmidrule(lr){5-13}
    & TITAN & CARE
    & & ABMIL & CLAM & TransMIL & WiKG & DAGMIL & GDFMIL & 2DMamba & AMDMIL & AEMMIL \\
    \midrule

    \multirow{2}{*}{\shortstack[l]{\texttt{BCNB}\\{\scriptsize(3 tasks)}}}
    & \multirow{2}{*}{\score{51.7}{2.3}}
    & \multirow{2}{*}{\score{50.0}{0.0}}
    & BACC
    & \score{69.7}{1.3} & \score{69.4}{2.4} & \score{70.6}{2.3} & \score{67.9}{2.3} & \score{68.0}{1.9} & \score{50.0}{0.0} & \score{50.2}{0.4} & \score{66.8}{1.3} & \score{63.9}{2.5} \\
    & & &
    $\Delta$
    & \posval{+18.0} & \posval{+17.7} & \posval{+18.9} & \posval{+16.2} & \posval{+16.3} & \negval{-1.7} & \negval{-1.5} & \posval{+15.1} & \posval{+12.2} \\

    \midrule

    \multirow{2}{*}{\shortstack[l]{\texttt{BRACS}\\{\scriptsize(2 tasks)}}}
    & \multirow{2}{*}{\score{40.3}{5.0}}
    & \multirow{2}{*}{\score{39.6}{1.2}}
    & BACC
    & \score{45.2}{3.4} & \score{46.0}{3.5} & \score{46.9}{4.7} & \score{47.5}{4.0} & \score{45.0}{3.1} & \score{33.5}{3.8} & \score{32.3}{1.9} & \score{53.1}{4.8} & \score{38.3}{2.2} \\
    & & &
    $\Delta$
    & \posval{+4.9} & \posval{+5.7} & \posval{+6.6} & \posval{+7.2} & \posval{+4.7} & \negval{-6.8} & \negval{-8.0} & \posval{+12.8} & \negval{-2.0} \\

    \midrule

    \multirow{2}{*}{\shortstack[l]{\texttt{CPTAC}\\{\scriptsize(2 tasks)}}}
    & \multirow{2}{*}{\score{92.9}{2.7}}
    & \multirow{2}{*}{\score{86.7}{4.0}}
    & BACC
    & \score{93.3}{0.9} & \score{94.4}{1.3} & \score{94.8}{0.2} & \score{95.5}{1.2} & \score{94.3}{0.9} & \score{91.1}{1.4} & \score{90.1}{1.2} & \score{94.2}{0.4} & \score{94.2}{1.1} \\
    & & &
    $\Delta$
    & \posval{+0.4} & \posval{+1.5} & \posval{+1.9} & \posval{+2.6} & \posval{+1.4} & \negval{-1.8} & \negval{-2.8} & \posval{+1.3} & \posval{+1.3} \\

    \midrule

    \multirow{2}{*}{\shortstack[l]{\texttt{KIDRARE}\\{\scriptsize(2 tasks)}}}
    & \multirow{2}{*}{\score{51.5}{5.5}}
    & \multirow{2}{*}{\score{44.9}{1.2}}
    & BACC
    & \score{61.0}{1.8} & \score{62.7}{1.6} & \score{61.6}{2.7} & \score{60.6}{1.6} & \score{53.8}{2.4} & \score{41.9}{3.3} & \score{42.3}{2.5} & \score{61.1}{2.9} & \score{55.4}{3.0} \\
    & & &
    $\Delta$
    & \posval{+9.5} & \posval{+11.2} & \posval{+10.1} & \posval{+9.1} & \posval{+2.3} & \negval{-9.6} & \negval{-9.2} & \posval{+9.6} & \posval{+3.9} \\

    \midrule

    \multirow{2}{*}{\shortstack[l]{\texttt{EBRAINS}\\{\scriptsize(3 tasks)}}}
    & \multirow{2}{*}{\score{56.8}{11.4}}
    & \multirow{2}{*}{\score{49.0}{3.7}}
    & BACC
    & \score{79.9}{1.0} & \score{80.8}{1.7} & \score{81.6}{1.7} & \score{81.5}{1.9} & \score{79.8}{1.2} & \score{71.7}{3.2} & \score{75.8}{2.0} & \score{82.2}{0.9} & \score{79.6}{1.7} \\
    & & &
    $\Delta$
    & \posval{+23.1} & \posval{+24.0} & \posval{+24.8} & \posval{+24.7} & \posval{+23.0} & \posval{+14.9} & \posval{+19.0} & \posval{+25.4} & \posval{+22.8} \\

    \midrule

    \multirow{2}{*}{\shortstack[l]{\texttt{MUT\_HET\_RCC}\\{\scriptsize(3 tasks)}}}
    & \multirow{2}{*}{\score{52.4}{1.5}}
    & \multirow{2}{*}{\score{51.2}{1.3}}
    & BACC
    & \score{62.6}{2.3} & \score{64.0}{2.3} & \score{62.0}{1.7} & \score{62.6}{1.6} & \score{59.7}{1.6} & \score{52.8}{0.6} & \score{52.8}{1.4} & \score{60.9}{2.8} & \score{56.1}{1.9} \\
    & & &
    $\Delta$
    & \posval{+10.2} & \posval{+11.6} & \posval{+9.6} & \posval{+10.2} & \posval{+7.3} & \posval{+0.4} & \posval{+0.4} & \posval{+8.5} & \posval{+3.7} \\



    \bottomrule
  \end{tabular}%
  }
\end{table*}

\subsection{Fine-tuning Comparison between From-scratch and Pretrained MIL Aggregators}

We evaluate distillation-pretrained initialization for downstream WSI classification using nine MIL aggregators under the same training protocol in Table~\ref{tab:merged_dataset_bacc}. Detailed results for all 15 individual downstream tasks are provided in the Appendix~\ref{sec:appendix-fine}. Overall, pretrained initialization improves most MIL models over from-scratch training, indicating that teacher-guided pretraining provides useful slide-level priors. At the dataset level, BRACS~\cite{brancati2022bracs} shows the largest average gain, suggesting that pretrained initialization is especially useful for tasks requiring subtle morphology and cross-patch aggregation. The pretrained MIL aggregators also achieve competitive performance compared with the teacher slide encoders in several cohorts, suggesting that knowledge from large slide foundation models can be effectively transferred into lightweight MIL architectures. Notably, although GDFMIL~\cite{zhang2026rethinking} and 2DMamba~\cite{zhang20252dmamba} show relatively weaker transferability under the frozen linear-probing setting, their performance is substantially improved after full-parameter fine-tuning. These results show that slide-level knowledge from large teachers can be effectively compressed into lightweight MIL aggregators for better downstream adaptation.

\begin{table*}[h]
  \centering
  \caption{Merged-dataset fine-tuning comparison between teacher models, from-scratch MIL, and pretrained MIL. Results are reported as test balanced accuracy (bacc, \%) with standard deviation shown in the lower-right subscript. $\Delta$ denotes pretrained minus scratch and is reported using mean values only.}
  \label{tab:merged_dataset_bacc}
  \scriptsize
  \setlength{\tabcolsep}{3pt}
  \renewcommand{\arraystretch}{1.05}
  \resizebox{\linewidth}{!}{%
  \begin{tabular}{lcclccccccccc}
    \toprule
    \multirow{2}{*}{Dataset}
    & \multicolumn{2}{c}{Teacher Models}
    & \multirow{2}{*}{Setting}
    & \multicolumn{9}{c}{Student MIL Aggregators} \\
    \cmidrule(lr){2-3} \cmidrule(lr){5-13}
    & TITAN & CARE
    & & ABMIL & CLAM & TransMIL & WiKG & DAGMIL & GDFMIL & 2DMamba & AMDMIL & AEMMIL \\
    \midrule

    \multirow{3}{*}{\shortstack[l]{\texttt{BCNB}\\{\scriptsize(3 tasks)}}}
    & \multirow{3}{*}{\score{69.3}{2.6}}
    & \multirow{3}{*}{\score{71.3}{2.7}}
    & Scratch & \score{70.4}{1.6} & \score{70.0}{2.6} & \score{65.2}{3.3} & \score{69.6}{2.7} & \score{67.4}{3.1} & \score{70.0}{5.2} & \score{69.2}{2.3} & \score{70.4}{2.1} & \score{72.3}{1.4} \\
    & & & Pretrain & \score{70.5}{1.8} & \score{71.3}{2.2} & \score{70.0}{3.3} & \score{69.4}{3.6} & \score{72.1}{1.9} & \score{71.5}{1.2} & \score{70.5}{2.0} & \score{72.5}{2.3} & \score{71.1}{1.8} \\
    & & & $\Delta$ & \posval{+0.1} & \posval{+1.3} & \posval{+4.7} & \negval{-0.2} & \posval{+4.7} & \posval{+1.5} & \posval{+1.3} & \posval{+2.1} & \negval{-1.2} \\
    \midrule

    \multirow{3}{*}{\shortstack[l]{\texttt{BRACS}\\{\scriptsize(2 tasks)}}}
    & \multirow{3}{*}{\score{52.2}{2.8}}
    & \multirow{3}{*}{\score{52.8}{2.3}}
    & Scratch & \score{50.9}{5.9} & \score{52.9}{2.3} & \score{47.2}{2.2} & \score{47.4}{6.6} & \score{55.1}{1.5} & \score{50.7}{3.3} & \score{57.9}{2.7} & \score{55.5}{4.7} & \score{54.0}{4.4} \\
    & & & Pretrain & \score{54.6}{2.4} & \score{52.9}{3.8} & \score{54.0}{3.6} & \score{53.0}{4.0} & \score{57.1}{5.1} & \score{56.8}{5.7} & \score{59.8}{2.9} & \score{54.6}{3.5} & \score{56.9}{2.1} \\
    & & & $\Delta$ & \posval{+3.6} & \neuval{0.0} & \posval{+6.8} & \posval{+5.6} & \posval{+1.9} & \posval{+6.1} & \posval{+2.0} & \negval{-0.8} & \posval{+2.9} \\
    \midrule

    \multirow{3}{*}{\shortstack[l]{\texttt{CPTAC}\\{\scriptsize(2 tasks)}}}
    & \multirow{3}{*}{\score{95.5}{0.6}}
    & \multirow{3}{*}{\score{94.9}{0.9}}
    & Scratch & \score{92.8}{0.6} & \score{93.5}{1.2} & \score{94.5}{1.5} & \score{91.9}{1.0} & \score{92.4}{2.3} & \score{93.2}{1.8} & \score{93.4}{1.1} & \score{94.9}{0.9} & \score{94.5}{1.5} \\
    & & & Pretrain & \score{94.1}{1.0} & \score{93.4}{1.2} & \score{94.0}{1.1} & \score{94.4}{2.0} & \score{95.4}{0.8} & \score{94.2}{0.5} & \score{92.9}{1.3} & \score{94.5}{1.1} & \score{93.2}{0.9} \\
    & & & $\Delta$ & \posval{+1.2} & \negval{-0.1} & \negval{-0.6} & \posval{+2.5} & \posval{+3.0} & \posval{+1.0} & \negval{-0.5} & \negval{-0.4} & \negval{-1.3} \\
    \midrule

    \multirow{3}{*}{\shortstack[l]{\texttt{KIDRARE}\\{\scriptsize(2 tasks)}}}
    & \multirow{3}{*}{\score{71.0}{5.4}}
    & \multirow{3}{*}{\score{69.7}{4.4}}
    & Scratch & \score{67.9}{2.3} & \score{68.6}{1.9} & \score{70.0}{3.4} & \score{69.7}{4.2} & \score{70.0}{3.1} & \score{69.6}{3.8} & \score{70.8}{3.4} & \score{75.3}{3.0} & \score{68.1}{2.8} \\
    & & & Pretrain & \score{69.6}{1.7} & \score{67.9}{2.4} & \score{72.9}{2.2} & \score{68.7}{3.2} & \score{67.4}{2.8} & \score{71.2}{3.6} & \score{71.1}{3.8} & \score{73.5}{2.7} & \score{70.1}{2.8} \\
    & & & $\Delta$ & \posval{+1.7} & \negval{-0.7} & \posval{+2.9} & \negval{-1.0} & \negval{-2.6} & \posval{+1.6} & \posval{+0.3} & \negval{-1.8} & \posval{+2.0} \\
    \midrule

    \multirow{3}{*}{\shortstack[l]{\texttt{EBRAINS}\\{\scriptsize(3 tasks)}}}
    & \multirow{3}{*}{\score{79.3}{3.7}}
    & \multirow{3}{*}{\score{82.0}{1.9}}
    & Scratch & \score{81.2}{1.5} & \score{81.4}{1.3} & \score{79.6}{2.1} & \score{79.6}{2.4} & \score{80.7}{2.0} & \score{79.9}{2.6} & \score{82.7}{1.3} & \score{81.9}{1.9} & \score{82.4}{1.4} \\
    & & & Pretrain & \score{81.1}{2.0} & \score{81.1}{1.4} & \score{80.2}{1.8} & \score{80.3}{2.2} & \score{81.0}{2.8} & \score{79.8}{1.8} & \score{81.6}{2.0} & \score{81.9}{2.2} & \score{82.9}{1.4} \\
    & & & $\Delta$ & \negval{-0.2} & \negval{-0.3} & \posval{+0.6} & \posval{+0.7} & \posval{+0.3} & \negval{-0.1} & \negval{-1.1} & \neuval{0.0} & \posval{+0.5} \\
    \midrule

    \multirow{3}{*}{\shortstack[l]{\texttt{MUT\_HET\_RCC}\\{\scriptsize(3 tasks)}}}
    & \multirow{3}{*}{\score{70.1}{2.8}}
    & \multirow{3}{*}{\score{67.2}{2.5}}
    & Scratch & \score{63.0}{2.9} & \score{63.6}{2.9} & \score{59.3}{4.0} & \score{62.3}{3.8} & \score{60.6}{4.9} & \score{59.0}{2.6} & \score{64.9}{3.3} & \score{65.3}{2.9} & \score{67.0}{3.6} \\
    & & & Pretrain & \score{63.5}{3.6} & \score{67.1}{3.1} & \score{64.9}{2.0} & \score{60.8}{6.3} & \score{60.9}{4.9} & \score{63.2}{5.3} & \score{67.1}{3.4} & \score{66.4}{3.3} & \score{66.4}{3.1} \\
    & & & $\Delta$ & \posval{+0.4} & \posval{+3.5} & \posval{+5.6} & \negval{-1.4} & \posval{+0.2} & \posval{+4.2} & \posval{+2.3} & \posval{+1.1} & \negval{-0.6} \\

    \bottomrule
  \end{tabular}%
  }
\end{table*}

\subsection{Few-shot transfer to downstream MIL tasks}
\begin{figure}[h]
    \centering
    \includegraphics[width=\linewidth]{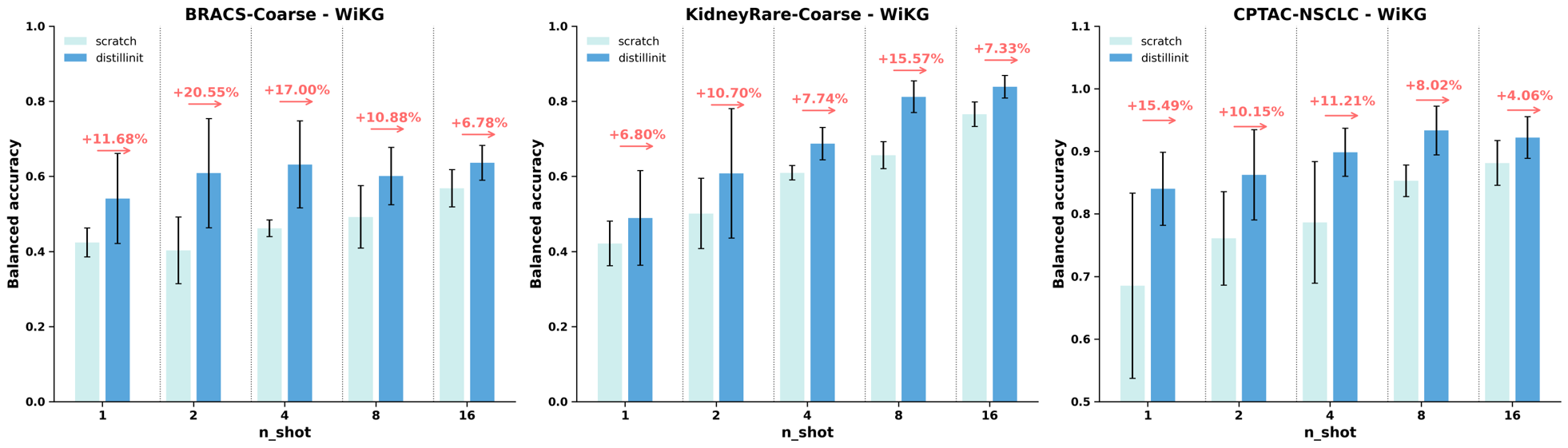}
    \caption{Few-shot comparison of from-scratch and distillation-initialized WiKG across three datasets.}
    \label{fig:few-shot}
\end{figure}

To evaluate the data efficiency of the proposed initialization, we further assess model performance under few-shot slide classification settings with the number of shots per class $K \in \{1,2,4,8,16\}$. For each setting, we randomly sample $K$ slides from each class to form the support set. Balanced accuracy is used as the evaluation metric. Figure~\ref{fig:few-shot} shows the results of WiKG~\cite{li2024wikg} on BRACS-Coarse, CPTAC-NSCLC, and KidneyRare-Coarse. Complete index results on other models and tasks are provided in Appendix~\ref{sec:appendix5}. Across all three datasets, distillation initialization consistently outperforms scratch training at every evaluated shot number. On BRACS-Coarse, the gain ranges from 6.78\% to 20.55\% percentage points; on CPTAC-NSCLC, from 4.06\% to 15.49\% percentage points; and on KidneyRare-Coarse, from 6.80\% to 15.57\% percentage points. The largest improvements are generally observed in lower-shot settings, indicating that distillation is particularly beneficial when labeled supervision is scarce. These results suggest that distillation transfers a useful prior for downstream MIL, improving both optimization stability and label efficiency.

\subsection{Ablation}

\paragraph{Effectiveness of multi-teacher distillation.}

\begin{wrapfigure}{r}{0.6\textwidth}
    \vspace{-0.5em}
    \centering
    \includegraphics[width=0.6\textwidth]{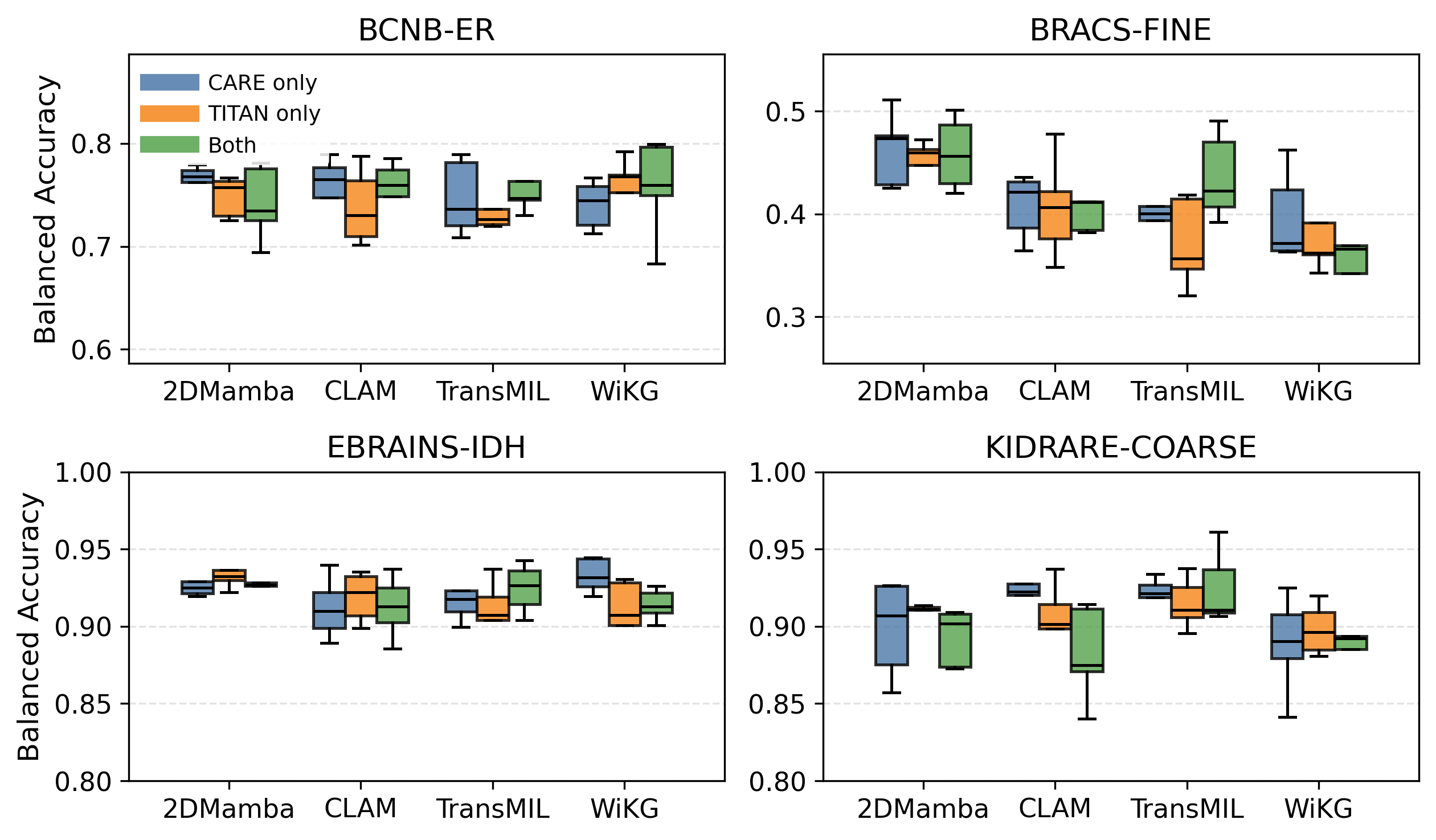}
    \caption{Teacher-source ablation for distillation pretraining. Results compare TITAN-only, CARE-only, and joint distillation across downstream datasets and MIL architectures.}
    \label{fig:ablation-teacher}
    \vspace{-0.5em}
\end{wrapfigure}
We ablate the teacher source used for distillation pretraining by comparing TITAN-only, CARE-only, and joint distillation from both teachers. The results in Figure~\ref{fig:ablation-teacher} show that the optimal teacher configuration varies across downstream datasets and MIL architectures, indicating that the transferred supervision from different slide-level teachers is architecture- and task-dependent. Specifically, CARE-only achieves strong results for 2DMamba~\cite{zhang20252dmamba} on BCNB-ER, while TITAN-only performs competitively for WiKG~\cite{li2024wikg} on BCNB-ER. Joint distillation from both teachers also brings advantages in several settings, such as Transmil~\citep{shao2021transmil} on KIDRARE-COARSE, 2DMamba~\cite{zhang20252dmamba} on BRACS-FINE and EBRAINS-IDH. These results suggest that CARE~\citep{zhang2026care} and TITAN~\citep{ding2025titan} encode complementary but non-uniform supervisory signals, and that combining them can be beneficial when their inductive biases are well aligned with the target task and MIL architecture. Detailed results for all 15 individual downstream tasks are provided in the Appendix~\ref{sec:appendixteacher}.


\paragraph{Effectiveness of distillation loss.}

\begin{wraptable}{r}{0.50\textwidth}
    \vspace{-0.8em}
    \centering
    \caption{
    Averaged balanced accuracy (bacc, \%) with standard deviation of different type loss across dataset groups.}
    \label{tab:losstype_group_avg_bacc}
    \scriptsize
    \setlength{\tabcolsep}{2.8pt}
    \renewcommand{\arraystretch}{1.08}
    \resizebox{\linewidth}{!}{%
    \begin{tabular}{lccc}
    \toprule
    Dataset & ADN & MSE & Cosine \\
    \midrule
    \makecell[l]{BCNB (3)} 
    & $70.5_{\scriptscriptstyle(1.8)}$
    & $\mathbf{70.8}_{\scriptscriptstyle(1.7)}$
    & $70.7_{\scriptscriptstyle(1.5)}$ \\

    \makecell[l]{BRACS (2)} 
    & $\mathbf{54.6}_{\scriptscriptstyle(2.4)}$
    & $53.5_{\scriptscriptstyle(2.7)}$
    & $53.8_{\scriptscriptstyle(3.2)}$ \\

    \makecell[l]{CPTAC (2)} 
    & $\mathbf{94.1}_{\scriptscriptstyle(1.0)}$
    & $93.4_{\scriptscriptstyle(0.8)}$
    & $93.6_{\scriptscriptstyle(1.3)}$ \\

    \makecell[l]{KIDRARE (2)} 
    & $\mathbf{69.6}_{\scriptscriptstyle(1.7)}$
    & $69.0_{\scriptscriptstyle(2.3)}$
    & $68.4_{\scriptscriptstyle(1.6)}$ \\

    \makecell[l]{EBRAINS (3)} 
    & $81.1_{\scriptscriptstyle(2.0)}$
    & $81.2_{\scriptscriptstyle(1.8)}$
    & $\mathbf{81.7}_{\scriptscriptstyle(1.5)}$ \\

    \makecell[l]{MUT-HET-RCC (3)} 
    & $63.5_{\scriptscriptstyle(3.6)}$
    & $63.1_{\scriptscriptstyle(3.7)}$
    & $\mathbf{64.4}_{\scriptscriptstyle(3.4)}$ \\
    \bottomrule
    \end{tabular}%
    }
    \vspace{-1.0em}
\end{wraptable}

We further compare three distillation objectives: ADN loss, mean squared error (MSE) loss, and cosine loss. Since different slide-level teachers may have distinct feature scales and embedding distributions, these objectives provide complementary ways to align student and teacher representations. Table~\ref{tab:losstype_group_avg_bacc} reports the averaged bacc within each dataset group. Overall, the three losses achieve comparable performance, while angle loss shows clear advantages on several dataset groups, including BRACS~\cite{brancati2022bracs}, CPTAC~\cite{cptac_portal}, and KIDRARE~\cite{he2026boosting}. This suggests that angular supervision can provide an effective geometric constraint for transferring slide-level teacher knowledge into lightweight MIL aggregators, especially when preserving relative representation structure is important. Complete results on all 15 downstream tasks are provided in Appendix~\ref{sec:appendixloss}.

\paragraph{Memory scalability analysis.}

We evaluate memory scalability on BRACS-coarse by increasing the number of patches from 5k to 30k for a single WSI. Compared with representative MIL models, CARE~\citep{zhang2026care} and TITAN~\citep{ding2025titan} show substantially faster memory growth. TITAN~\citep{ding2025titan} already exceeds 35 GiB at 10k patches and becomes OOM beyond 15k patches, while CARE~\citep{zhang2026care} reaches over 30 GiB at 30k patches. In contrast, MIL aggregators remain stable at 30k patches with much lower memory usage, including WiKG~\cite{li2024wikg} at about 5.5 GiB, TransMIL~\citep{shao2021transmil} at about 3.3 GiB, and ABMIL~\cite{ilse2018attention} at about 0.24 GiB. These results highlight the practical scalability of lightweight MIL aggregators for large-patch WSI analysis and efficient downstream fine-tuning.

\begin{figure}[h]
    \centering
    \includegraphics[width=0.9\linewidth]{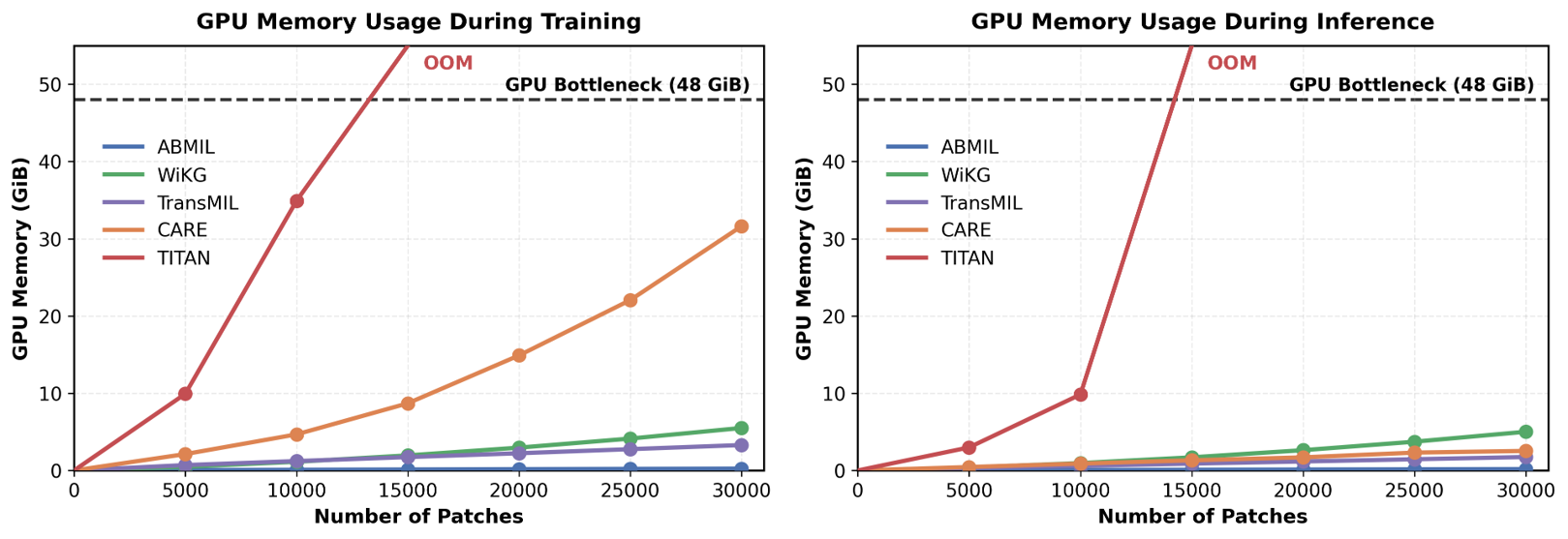}
    \caption{GPU memory scalability during training and testing under increasing patch numbers.}
    \label{fig:pic1}
\end{figure}

\section{Conclusion and Future Work}

In this work, we demonstrate that lightweight MIL aggregators can be effectively pretrained through slide-level knowledge distillation. Using CARE and TITAN as teacher models, we transfer their slide-level representations into diverse MIL architectures with an angular dispersion normalized distillation loss. The resulting pretrained aggregators provide reusable initialization for downstream adaptation, offering a practical alternative to conventional from-scratch MIL training. Across multiple computational pathology benchmarks, distilled initialization improves most MIL architectures and brings clear benefits in few-shot settings. Overall, distilled MIL initialization enables transferable slide-level priors while remaining efficient to adapt for WSI tasks with limited labels or computational resources. Future work will scale up the pretraining data and evaluate the proposed framework on a broader range of downstream tasks.

\bibliographystyle{plain}
\bibliography{references}

@article{laleh2022benchmarking,
  title   = {Benchmarking weakly-supervised deep learning pipelines for whole slide classification in computational pathology},
  author  = {Ghaffari Laleh, Narmin and Muti, Hannah Sophie and Loeffler, Chiara Maria Lavinia and Echle, Amelie and Saldanha, Oliver Lester and Mahmood, Faisal and Lu, Ming Y. and Trautwein, Christian and Langer, Rupert and Dislich, Bastian and Buelow, Roman D. and Grabsch, Heike Irmgard and Brenner, Hermann and Chang-Claude, Jenny and Alwers, Elizabeth and Brinker, Titus J. and Khader, Firas and Truhn, Daniel and Gaisa, Nadine T. and Boor, Peter and Schulz, Volkmar and Krisam, Johannes and Kather, Jakob Nikolas},
  journal = {Medical Image Analysis},
  volume  = {79},
  pages   = {102474},
  year    = {2022},
  doi     = {10.1016/j.media.2022.102474}
}

@article{althelaya2023survey,
  title   = {Applications of discriminative and deep learning feature extraction methods for whole slide image analysis: A survey},
  author  = {Al-Thelaya, Khaled and Gilal, Nauman Ullah and Alzubaidi, Mahmood and Majeed, Fahad and Agus, Marco and Schneider, Jens and Househ, Mowafa},
  journal = {Journal of Pathology Informatics},
  volume  = {14},
  pages   = {100335},
  year    = {2023},
  doi     = {10.1016/j.jpi.2023.100335}
}

@inproceedings{chen2020simclr,
  title     = {A Simple Framework for Contrastive Learning of Visual Representations},
  author    = {Chen, Ting and Kornblith, Simon and Norouzi, Mohammad and Hinton, Geoffrey},
  booktitle = {Proceedings of the 37th International Conference on Machine Learning},
  series    = {Proceedings of Machine Learning Research},
  volume    = {119},
  pages     = {1597--1607},
  year      = {2020},
  publisher = {PMLR},
  url       = {https://proceedings.mlr.press/v119/chen20j.html}
}

@inproceedings{he2020moco,
  title     = {Momentum Contrast for Unsupervised Visual Representation Learning},
  author    = {He, Kaiming and Fan, Haoqi and Wu, Yuxin and Xie, Saining and Girshick, Ross},
  booktitle = {Proceedings of the IEEE/CVF Conference on Computer Vision and Pattern Recognition (CVPR)},
  pages     = {9729--9738},
  year      = {2020},
  url       = {https://openaccess.thecvf.com/content_CVPR_2020/html/He_Momentum_Contrast_for_Unsupervised_Visual_Representation_Learning_CVPR_2020_paper.html}
}

@inproceedings{caron2021dino,
  title     = {Emerging Properties in Self-Supervised Vision Transformers},
  author    = {Caron, Mathilde and Touvron, Hugo and Misra, Ishan and J{\'e}gou, Herv{\'e} and Mairal, Julien and Bojanowski, Piotr and Joulin, Armand},
  booktitle = {Proceedings of the IEEE/CVF International Conference on Computer Vision (ICCV)},
  pages     = {9650--9660},
  year      = {2021},
  url       = {https://openaccess.thecvf.com/content/ICCV2021/html/Caron_Emerging_Properties_in_Self-Supervised_Vision_Transformers_ICCV_2021_paper.html}
}

@inproceedings{kang2023benchmarking,
  title     = {Benchmarking Self-Supervised Learning on Diverse Pathology Datasets},
  author    = {Kang, Mingu and Song, Heon and Park, Seonwook and Yoo, Donggeun and Pereira, S{\'e}rgio},
  booktitle = {Proceedings of the IEEE/CVF Conference on Computer Vision and Pattern Recognition (CVPR)},
  pages     = {3344--3354},
  year      = {2023},
  url       = {https://openaccess.thecvf.com/content/CVPR2023/html/Kang_Benchmarking_Self-Supervised_Learning_on_Diverse_Pathology_Datasets_CVPR_2023_paper.html}
}

@inproceedings{li2021dsmil,
  title     = {Dual-Stream Multiple Instance Learning Network for Whole Slide Image Classification with Self-supervised Contrastive Learning},
  author    = {Li, Bin and Li, Yin and Eliceiri, Kevin W.},
  booktitle = {Proceedings of the IEEE/CVF Conference on Computer Vision and Pattern Recognition (CVPR)},
  pages     = {14318--14328},
  year      = {2021},
  url       = {https://openaccess.thecvf.com/content/CVPR2021/html/Li_Dual-Stream_Multiple_Instance_Learning_Network_for_Whole_Slide_Image_Classification_CVPR_2021_paper.html}
}

@article{wang2022ctranspath,
  title   = {Transformer-based unsupervised contrastive learning for histopathological image classification},
  author  = {Wang, Xiyue and Yang, Sen and Zhang, Jun and Wang, Minghui and Zhang, Jing and Yang, Wei and Huang, Junzhou and Han, Xiao},
  journal = {Medical Image Analysis},
  volume  = {81},
  pages   = {102559},
  year    = {2022},
  doi     = {10.1016/j.media.2022.102559}
}

@inproceedings{jaume2024madeleine,
  title     = {Multistain Pretraining for Slide Representation Learning in Pathology},
  author    = {Jaume, Guillaume and Vaidya, Anurag and Zhang, Andrew and Song, Andrew H. and Chen, Richard J. and Sahai, Sharifa and Mo, Dandan and Madrigal, Emilio and Le, Long Phi and Mahmood, Faisal},
  booktitle = {Computer Vision -- ECCV 2024},
  pages     = {19--37},
  year      = {2024},
  publisher = {Springer},
  doi       = {10.1007/978-3-031-73414-4_2}
}

@article{wei2022featuredistillation,
  title         = {Contrastive Learning Rivals Masked Image Modeling in Fine-tuning via Feature Distillation},
  author        = {Wei, Yixuan and Hu, Han and Xie, Zhenda and Zhang, Zheng and Cao, Yue and Bao, Jianmin and Chen, Dong and Guo, Baining},
  journal       = {arXiv preprint arXiv:2205.14141},
  year          = {2022},
  eprint        = {2205.14141},
  archivePrefix = {arXiv},
  primaryClass  = {cs.CV},
  url           = {https://arxiv.org/abs/2205.14141}
}

@inproceedings{ahn2019vid,
  title     = {Variational Information Distillation for Knowledge Transfer},
  author    = {Ahn, Sungsoo and Hu, Shell Xu and Damianou, Andreas and Lawrence, Neil D. and Dai, Zhenwen},
  booktitle = {Proceedings of the IEEE/CVF Conference on Computer Vision and Pattern Recognition (CVPR)},
  pages     = {9155--9163},
  year      = {2019},
  doi       = {10.1109/CVPR.2019.00938}
}

@inproceedings{beyer2022patient,
  title     = {Knowledge Distillation: A Good Teacher Is Patient and Consistent},
  author    = {Beyer, Lucas and Zhai, Xiaohua and Royer, Am{\'e}lie and Markeeva, Larisa and Anil, Rohan and Kolesnikov, Alexander},
  booktitle = {Proceedings of the IEEE/CVF Conference on Computer Vision and Pattern Recognition (CVPR)},
  pages     = {10925--10934},
  year      = {2022},
  url       = {https://openaccess.thecvf.com/content/CVPR2022/html/Beyer_Knowledge_Distillation_A_Good_Teacher_Is_Patient_and_Consistent_CVPR_2022_paper.html}
}

@inproceedings{mirzadeh2020teacherassistant,
  title     = {Improved Knowledge Distillation via Teacher Assistant},
  author    = {Mirzadeh, Seyed Iman and Farajtabar, Mehrdad and Li, Ang and Levine, Nir and Matsukawa, Akihiro and Ghasemzadeh, Hassan},
  booktitle = {Proceedings of the AAAI Conference on Artificial Intelligence},
  volume    = {34},
  number    = {04},
  pages     = {5191--5198},
  year      = {2020},
  url       = {https://ojs.aaai.org/index.php/AAAI/article/view/5963}
}

@article{ranzinger2026cradiov4,
  title         = {C-RADIOv4 (Tech Report)},
  author        = {Ranzinger, Mike and Heinrich, Greg and McCarthy, Collin and Kautz, Jan and Tao, Andrew and Catanzaro, Bryan and Molchanov, Pavlo},
  journal       = {arXiv preprint arXiv:2601.17237},
  year          = {2026},
  eprint        = {2601.17237},
  archivePrefix = {arXiv},
  primaryClass  = {cs.CV},
  url           = {https://arxiv.org/abs/2601.17237}
}

@article{liu2020amtmlkd,
  title   = {Adaptive multi-teacher multi-level knowledge distillation},
  author  = {Liu, Yuang and Zhang, Wei and Wang, Jun},
  journal = {Neurocomputing},
  volume  = {415},
  pages   = {106--113},
  year    = {2020},
  doi     = {10.1016/j.neucom.2020.07.048}
}

@inproceedings{park2020feed,
  title     = {Feature-Level Ensemble Knowledge Distillation for Aggregating Knowledge from Multiple Networks},
  author    = {Park, SeongUk and Kwak, Nojun},
  booktitle = {ECAI 2020},
  pages     = {1411--1418},
  year      = {2020},
  publisher = {IOS Press},
  doi       = {10.3233/FAIA200246}
}

@article{xu2025mil_foundation,
  title={When Multiple Instance Learning Meets Foundation Models: Advancing Histological Whole Slide Image Analysis},
  author={Xu, Hongming and Wang, Mingkang and Shi, Duanbo and Qin, Huamin and Zhang, Yunpeng and Liu, Zaiyi and Madabhushi, Anant and Gao, Peng and Cong, Fengyu and Lu, Cheng},
  journal={Medical Image Analysis},
  volume={101},
  pages={103456},
  year={2025},
  doi={10.1016/j.media.2025.103456}
}

@inproceedings{ilse2018attention,
  title={Attention-based Deep Multiple Instance Learning},
  author={Ilse, Maximilian and Tomczak, Jakub M. and Welling, Max},
  booktitle={Proceedings of the 35th International Conference on Machine Learning},
  series={Proceedings of Machine Learning Research},
  volume={80},
  pages={2127--2136},
  year={2018},
  publisher={PMLR},
  url={https://proceedings.mlr.press/v80/ilse18a.html}
}

@article{lu2021clam,
  title={Data-efficient and Weakly Supervised Computational Pathology on Whole-Slide Images},
  author={Lu, Ming Y. and Williamson, Drew F. K. and Chen, Tiffany Y. and Chen, Richard J. and Barbieri, Matteo and Mahmood, Faisal},
  journal={Nature Biomedical Engineering},
  volume={5},
  number={6},
  pages={555--570},
  year={2021},
  doi={10.1038/s41551-020-00682-w}
}

@inproceedings{shao2021transmil,
  title={TransMIL: Transformer based Correlated Multiple Instance Learning for Whole Slide Image Classification},
  author={Shao, Zhuchen and Bian, Hao and Chen, Yang and Wang, Yifeng and Zhang, Jian and Ji, Xiangyang and Zhang, Yongbing},
  booktitle={Advances in Neural Information Processing Systems},
  volume={34},
  pages={2136--2147},
  year={2021}
}

@inproceedings{li2024wikg,
  title={Dynamic Graph Representation with Knowledge-aware Attention for Histopathology Whole Slide Image Analysis},
  author={Li, Jiawen and Chen, Yuxuan and Chu, Hongbo and Sun, Qiehe and Guan, Tian and Han, Anjia and He, Yonghong},
  booktitle={Proceedings of the IEEE/CVF Conference on Computer Vision and Pattern Recognition},
  pages={11323--11332},
  year={2024}
}

@article{fu2025dagrl,
  title={Deformable Attention Graph Representation Learning for Histopathology Whole Slide Image Analysis},
  author={Fu, Mingxi and Ling, Xitong and Chen, Yuxuan and Li, Jiawen and Fu, Fanglei and Yuan, Huaitian and Guan, Tian and He, Yonghong and Zhu, Lianghui},
  journal={arXiv preprint arXiv:2508.05382},
  year={2025},
  url={https://arxiv.org/abs/2508.05382}
}

@article{oquab2024dinov2,
  title={DINOv2: Learning Robust Visual Features without Supervision},
  author={Oquab, Maxime and Darcet, Timoth{\'e}e and Moutakanni, Th{\'e}o and Vo, Huy and Szafraniec, Marc and Khalidov, Vasil and Fernandez, Pierre and Haziza, Daniel and Massa, Francisco and El-Nouby, Alaaeldin and Assran, Mahmoud and Ballas, Nicolas and Galuba, Wojciech and Howes, Russell and Huang, Po-Yao and Li, Shang-Wen and Misra, Ishan and Rabbat, Michael and Sharma, Vasu and Synnaeve, Gabriel and Xu, Hu and J{\'e}gou, Herv{\'e} and Mairal, Julien and Labatut, Patrick and Joulin, Armand and Bojanowski, Piotr},
  journal={arXiv preprint arXiv:2304.07193},
  year={2024},
  url={https://arxiv.org/abs/2304.07193}
}

@article{chen2024uni,
  title={Towards a General-Purpose Foundation Model for Computational Pathology},
  author={Chen, Richard J. and Ding, Tong and Lu, Ming Y. and Williamson, Drew F. K. and Jaume, Guillaume and Chen, Bowen and Zhang, Andrew and Shao, Daniel and Song, Andrew H. and Shaban, Muhammad and Williams, Mane and Vaidya, Anurag and Sahai, Sharifa and Oldenburg, Lukas and Weishaupt, Luca L. and Wang, Judy J. and Williams, Walt and Le, Long Phi and Gerber, Georg and Mahmood, Faisal},
  journal={Nature Medicine},
  volume={30},
  number={3},
  pages={850--862},
  year={2024},
  doi={10.1038/s41591-024-02857-3}
}

@article{xu2024provgigapath,
  title={A Whole-Slide Foundation Model for Digital Pathology from Real-World Data},
  author={Xu, Hanwen and Usuyama, Naoto and Bagga, Jaume and others},
  journal={Nature},
  volume={630},
  number={8015},
  pages={181--188},
  year={2024},
  doi={10.1038/s41586-024-07441-w}
}

@article{vorontsov2024virchow,
  title={A Foundation Model for Clinical-Grade Computational Pathology and Rare Cancers Detection},
  author={Vorontsov, Eugene and Bozkurt, Alican and Casson, Adam and Shaikovski, George and Zelechowski, Michal and Severson, Kristen and Zimmermann, Eric and Hall, James and Tenenholtz, Noam and Fusi, Nicolo and others},
  journal={Nature Medicine},
  volume={30},
  number={10},
  pages={2924--2935},
  year={2024},
  doi={10.1038/s41591-024-03141-0}
}

@article{ding2025titan,
  title={A Multimodal Whole-Slide Foundation Model for Pathology},
  author={Ding, Tong and Jaume, Guillaume and others},
  journal={Nature Medicine},
  volume={31},
  number={11},
  pages={3749--3761},
  year={2025},
  doi={10.1038/s41591-025-03982-3}
}

@article{lu2024conch,
  title={A Visual-Language Foundation Model for Computational Pathology},
  author={Lu, Ming Y. and Chen, Bowen and Williamson, Drew F. K. and Chen, Richard J. and Liang, Ivy and Ding, Tong and Jaume, Guillaume and Odintsov, Igor and Le, Long Phi and Gerber, Georg and Parwani, Anil V. and Zhang, Andrew and Mahmood, Faisal},
  journal={Nature Medicine},
  volume={30},
  number={3},
  pages={863--874},
  year={2024},
  doi={10.1038/s41591-024-02856-4}
}

@article{xiang2025musk,
  title={A Vision-Language Foundation Model for Precision Oncology},
  author={Xiang, Jinxi and Wang, Xiyue and Zhang, Xiaoming and Xi, Yinghua and Eweje, Feyisope and Chen, Yijiang and Li, Yuchen and Bergstrom, Colin and Gopaulchan, Matthew and Kim, Ted and Yu, Kun-Hsing and others},
  journal={Nature},
  volume={638},
  number={8051},
  pages={769--778},
  year={2025},
  doi={10.1038/s41586-024-08378-w}
}

@article{hinton2015distilling,
  title={Distilling the Knowledge in a Neural Network},
  author={Hinton, Geoffrey and Vinyals, Oriol and Dean, Jeff},
  journal={arXiv preprint arXiv:1503.02531},
  year={2015},
  url={https://arxiv.org/abs/1503.02531}
}

@inproceedings{radford2021clip,
  title={Learning Transferable Visual Models From Natural Language Supervision},
  author={Radford, Alec and Kim, Jong Wook and Hallacy, Chris and Ramesh, Aditya and Goh, Gabriel and Agarwal, Sandhini and Sastry, Girish and Askell, Amanda and Mishkin, Pamela and Clark, Jack and Krueger, Gretchen and Sutskever, Ilya},
  booktitle={Proceedings of the 38th International Conference on Machine Learning},
  series={Proceedings of Machine Learning Research},
  volume={139},
  pages={8748--8763},
  year={2021},
  publisher={PMLR},
  url={https://proceedings.mlr.press/v139/radford21a.html}
}

@inproceedings{kirillov2023sam,
  title={Segment Anything},
  author={Kirillov, Alexander and Mintun, Eric and Ravi, Nikhila and Mao, Hanzi and Rolland, Chloe and Gustafson, Laura and Xiao, Tete and Whitehead, Spencer and Berg, Alexander C. and Lo, Wan-Yen and Dollar, Piotr and Girshick, Ross},
  booktitle={Proceedings of the IEEE/CVF International Conference on Computer Vision},
  pages={4015--4026},
  year={2023}
}

@inproceedings{ranzinger2024amradio,
  title={AM-RADIO: Agglomerative Vision Foundation Model Reduce All Domains Into One},
  author={Ranzinger, Mike and Heinrich, Greg and Kautz, Jan and Molchanov, Pavlo},
  booktitle={Proceedings of the IEEE/CVF Conference on Computer Vision and Pattern Recognition},
  pages={12490--12500},
  year={2024},
  doi={10.1109/CVPR52733.2024.01187}
}

@article{ma2026gpfm,
  title={A Generalizable Pathology Foundation Model Using a Unified Knowledge Distillation Pretraining Framework},
  author={Ma, Jiabo and Guo, Zhengrui and Zhou, Fengtao and Wang, Yihui and Xu, Yingxue and Li, Jinbang and Yan, Fang and Cai, Yu and Zhu, Zhengjie and Jin, Cheng and others},
  journal={Nature Biomedical Engineering},
  volume={10},
  number={3},
  pages={545--564},
  year={2026},
  doi={10.1038/s41551-025-01488-4}
}

@inproceedings{shao2025miltransfer,
  title={Do Multiple Instance Learning Models Transfer?},
  author={Shao, Daniel and Chen, Richard J. and Song, Andrew H. and Runevic, Joel and Lu, Ming Y. and Ding, Tong and Mahmood, Faisal},
  booktitle={Proceedings of the 42nd International Conference on Machine Learning},
  series={Proceedings of Machine Learning Research},
  volume={267},
  pages={54219--54238},
  year={2025},
  publisher={PMLR},
  url={https://proceedings.mlr.press/v267/shao25a.html}
}

@article{zhang2026care,
  title={CARE: A Molecular-Guided Foundation Model with Adaptive Region Modeling for Whole Slide Image Analysis},
  author={Zhang, Di and Gong, Zhangpeng and Pang, Xiaobo and Liu, Jiashuai and Lu, Junbo and Cui, Hao and Ge, Jiusong and Zeng, Zhi and Yi, Kai and Li, Yinghua and Liu, Si and Yu, Tingsong and Wang, Haoran and Crispin-Ortuzar, Mireia and Yu, Weimiao and Li, Chen and Gao, Zeyu},
  journal={arXiv preprint arXiv:2602.21637},
  year={2026},
  url={https://arxiv.org/abs/2602.21637}
}

@inproceedings{zhang20252dmamba,
  title={2dmamba: Efficient state space model for image representation with applications on giga-pixel whole slide image classification},
  author={Zhang, Jingwei and Nguyen, Anh Tien and Han, Xi and Trinh, Vincent Quoc-Huy and Qin, Hong and Samaras, Dimitris and Hosseini, Mahdi S},
  booktitle={Proceedings of the Computer Vision and Pattern Recognition Conference},
  pages={3583--3592},
  year={2025}
}

@article{wang2024pathology,
  title={A pathology foundation model for cancer diagnosis and prognosis prediction},
  author={Wang, Xiyue and Zhao, Junhan and Marostica, Eliana and Yuan, Wei and Jin, Jietian and Zhang, Jiayu and Li, Ruijiang and Tang, Hongping and Wang, Kanran and Li, Yu and others},
  journal={Nature},
  volume={634},
  number={8035},
  pages={970--978},
  year={2024},
  publisher={Nature Publishing Group UK London}
}

@inproceedings{he2016deep,
  title={Deep Residual Learning for Image Recognition},
  author={He, Kaiming and Zhang, Xiangyu and Ren, Shaoqing and Sun, Jian},
  booktitle={Proceedings of the IEEE Conference on Computer Vision and Pattern Recognition (CVPR)},
  pages={770--778},
  year={2016}
}

@inproceedings{dosovitskiy2021image,
  title={An Image is Worth 16x16 Words: Transformers for Image Recognition at Scale},
  author={Dosovitskiy, Alexey and Beyer, Lucas and Kolesnikov, Alexander and Weissenborn, Dirk and Zhai, Xiaohua and Unterthiner, Thomas and Dehghani, Mostafa and Minderer, Matthias and Heigold, Georg and Gelly, Sylvain and others},
  booktitle={International Conference on Learning Representations (ICLR)},
  year={2021}
}

@inproceedings{kornblith2019better,
  title={Do Better ImageNet Models Transfer Better?},
  author={Kornblith, Simon and Shlens, Jonathon and Le, Quoc V.},
  booktitle={Proceedings of the IEEE/CVF Conference on Computer Vision and Pattern Recognition (CVPR)},
  pages={2661--2671},
  year={2019}
}

@inproceedings{kolesnikov2020big,
  title={Big Transfer (BiT): General Visual Representation Learning},
  author={Kolesnikov, Alexander and Beyer, Lucas and Zhai, Xiaohua and Puigcerver, Joan and Yung, Jessica and Gelly, Sylvain and Houlsby, Neil},
  booktitle={European Conference on Computer Vision (ECCV)},
  pages={491--507},
  year={2020}
}

@inproceedings{kingma2015adam,
  author    = {Kingma, Diederik P. and Ba, Jimmy},
  title     = {Adam: A Method for Stochastic Optimization},
  booktitle = {International Conference on Learning Representations (ICLR)},
  year      = {2015},
  url       = {https://arxiv.org/abs/1412.6980}
}

@article{xu2021bcnb,
  author    = {Feng Xu and Chuang Zhu and Wenqi Tang and Ying Wang and Yu Zhang and Jie Li and Hongchuan Jiang and Zhongyue Shi and Jun Liu and Mulan Jin},
  title     = {Predicting Axillary Lymph Node Metastasis in Early Breast Cancer Using Deep Learning on Primary Tumor Biopsy Slides},
  journal   = {Frontiers in Oncology},
  volume    = {11},
  pages     = {759007},
  year      = {2021},
  doi       = {10.3389/fonc.2021.759007}
}

@article{brancati2022bracs,
  author    = {Nadia Brancati and Maria Frucci and Daniela Riccio and Giuseppe De Pietro and others},
  title     = {BRACS: A Dataset for BReAst Carcinoma Subtyping in H\&E Histology Images},
  journal   = {Database},
  year      = {2022},
  pages     = {baac093},
  doi       = {10.1093/database/baac093}
}

@misc{cptac_portal,
  author       = {{National Cancer Institute Genomic Data Commons}},
  title        = {Clinical Proteomic Tumor Analysis Consortium (CPTAC)},
  year         = {2026},
  howpublished = {\url{https://gdc.cancer.gov/about-gdc/contributed-genomic-data-cancer-research/clinical-proteomic-tumor-analysis-consortium-cptac}},
  note         = {Accessed: 2026-04-21}
}

@article{roetzer2022digital,
  author    = {Theresa Roetzer-Pejrimovsky and others},
  title     = {The Digital Brain Tumour Atlas, an Open Histopathology Resource},
  journal   = {Scientific Data},
  volume    = {9},
  year      = {2022},
  doi       = {10.1038/s41597-022-01157-0}
}

@article{he2026boosting,
  author    = {Dexuan He and Xiao Zhou and Wenbin Guan and Liyuan Zhang and Xiaoman Zhang and Sinuo Xu and Ge Wang and Lifeng Wang and Xiaojun Yuan and Jing Ma and Xin Sun and Yanfeng Wang and Kun Sun and Ya Zhang and Weidi Xie},
  title     = {Boosting Pathology Foundation Models via Few-shot Prompt-tuning for Rare Cancer Subtyping},
  journal   = {Nature Communications},
  year      = {2026},
  doi       = {10.1038/s41467-026-71715-2}
}

@article{acosta2022intratumoral,
  author    = {P. H. Acosta and others},
  title     = {Intratumoral Resolution of Driver Gene Mutation Heterogeneity in Renal Cancer Using Deep Learning},
  journal   = {Cancer Research},
  volume    = {82},
  number    = {15},
  pages     = {2792--2806},
  year      = {2022},
  doi       = {10.1158/0008-5472.CAN-21-4097}
}

@inproceedings{zhang2026rethinking,
  title={Rethinking Multi-Instance Learning Through Graph-Driven Fusion: A Dual-Path Approach to Adaptive Representation},
  author={Zhang, Yu-Xuan and Zhou, Zhengchun and Liu, Weisha and Zhang, Mingxing},
  booktitle={Proceedings of the AAAI Conference on Artificial Intelligence},
  volume={40},
  number={34},
  pages={28510--28518},
  year={2026}
}

@inproceedings{ling2024agent,
  title={Agent aggregator with mask denoise mechanism for histopathology whole slide image analysis},
  author={Ling, Xitong and Ouyang, Minxi and Wang, Yizhi and Chen, Xinrui and Yan, Renao and Chu, Hongbo and Cheng, Junru and Guan, Tian and Tian, Sufang and Liu, Xiaoping and others},
  booktitle={Proceedings of the 32nd ACM International Conference on Multimedia},
  pages={2795--2803},
  year={2024}
}

@article{jing2021understanding,
  title={Understanding dimensional collapse in contrastive self-supervised learning},
  author={Jing, Li and Vincent, Pascal and LeCun, Yann and Tian, Yuandong},
  journal={arXiv preprint arXiv:2110.09348},
  year={2021}
}

@article{shaikovski2024prism,
  title={Prism: A multi-modal generative foundation model for slide-level histopathology},
  author={Shaikovski, George and Casson, Adam and Severson, Kristen and Zimmermann, Eric and Wang, Yi Kan and Kunz, Jeremy D and Retamero, Juan A and Oakley, Gerard and Klimstra, David and Kanan, Christopher and others},
  journal={arXiv preprint arXiv:2405.10254},
  year={2024}
}


\appendix

\section{Technical appendices and supplementary material}

\subsection{Algorithm: Dual-teacher Slide-level Distillation Pretraining.}
\label{sec:appendix2}

For completeness, we provide the detailed pseudocode of the proposed dual-teacher slide-level distillation pretraining procedure in Algorithm~\ref{alg:dual_teacher_distill}. The algorithm summarizes the full training pipeline, including offline teacher cache construction, student forward propagation, teacher-specific projection and normalization, EMA-based angular dispersion estimation, and the final weighted dual-teacher distillation objective. This appendix is intended to clarify the implementation details of the pretraining stage and to improve reproducibility of our framework.

\begin{algorithm}[h]
\caption{Dual-teacher slide-level distillation pretraining}
\label{alg:dual_teacher_distill}
\footnotesize
\begin{algorithmic}[1]
\Require Unlabeled slides $\mathcal{D}=\{(\mathbf{X}_s,\mathbf{C}_s)\}_{s=1}^{S}$
\Require Frozen teachers $T_{\mathrm{titan}}, T_{\mathrm{care}}$
\Require Student MIL aggregator $g_\theta$
\Require Projection heads $\phi_{\mathrm{titan}}, \phi_{\mathrm{care}}$
\Require Teacher weights $w_{\mathrm{titan}}, w_{\mathrm{care}}$
\State Let $\mathcal{K}=\{\mathrm{titan}, \mathrm{care}\}$
\State \textbf{Build teacher cache}
\ForAll{slides $(\mathbf{X}_s,\mathbf{C}_s)\in\mathcal{D}$}
    \State $\mathbf{t}_s^{\mathrm{titan}} \gets T_{\mathrm{titan}}(\mathbf{X}_s,\mathbf{C}_s)$
    \State $\mathbf{t}_s^{\mathrm{care}} \gets T_{\mathrm{care}}(\mathbf{X}_s,\mathbf{C}_s)$
    \State cache $(\mathbf{t}_s^{\mathrm{titan}},\mathbf{t}_s^{\mathrm{care}})$
\EndFor
\State Initialize EMA statistics $(\boldsymbol{\mu}^{(k)},\sigma^{(k)})$ for all $k\in\mathcal{K}$
\For{epoch $=1$ to $E$}
    \ForAll{mini-batches $\mathcal{B}=\{(\mathbf{X}_i,\mathbf{C}_i,\mathbf{t}_i^{\mathrm{titan}},\mathbf{t}_i^{\mathrm{care}})\}_{i=1}^{B}$}
        \State $\mathbf{z}_i \gets g_\theta(\mathbf{X}_i,\mathbf{C}_i)$ for all $i$
        \State $\hat{\mathbf{t}}_i^{\mathrm{titan}} \gets \phi_{\mathrm{titan}}(\mathbf{z}_i)$ for all $i$
        \State $\hat{\mathbf{t}}_i^{\mathrm{care}} \gets \phi_{\mathrm{care}}(\mathbf{z}_i)$ for all $i$
        \ForAll{$k\in\mathcal{K}$}
            \State $\hat{\mathbf{t}}_i^{(k)} \gets \ell_2\text{-norm}(\hat{\mathbf{t}}_i^{(k)})$ for all $i$
            \State $\mathbf{t}_i^{(k)} \gets \ell_2\text{-norm}(\mathbf{t}_i^{(k)})$ for all $i$
            \State $e_i^{(k)} \gets 1-\cos(\hat{\mathbf{t}}_i^{(k)},\mathbf{t}_i^{(k)})$ for all $i$
            \State update $\boldsymbol{\mu}^{(k)}$ via EMA
            \State update $\sigma^{(k)}$ via EMA and clamp to $\sigma_{\min}$
            \State $\mathcal{L}^{(k)} \gets \frac{1}{B}\sum_{i=1}^{B}\frac{e_i^{(k)}}{\sigma^{(k)}+\epsilon}$
        \EndFor
        \State $\mathcal{L} \gets w_{\mathrm{titan}}\mathcal{L}^{\mathrm{titan}} + w_{\mathrm{care}}\mathcal{L}^{\mathrm{care}}$
        \State update $\theta,\phi_{\mathrm{titan}},\phi_{\mathrm{care}}$ with AdamW on $\mathcal{L}$
    \EndFor
\EndFor
\State \Return distilled student weights $\theta^\ast$
\end{algorithmic}
\end{algorithm}

\subsection{Detailed description of evaluation datasets.}
\label{sec:appendix1}

We evaluate our method on six public whole-slide image (WSI) cohorts covering biomarker prediction, subtype classification, pan-cancer classification, and mutation prediction. Below, we provide detailed descriptions of all evaluation datasets, including task definitions, class compositions, and train/validation/test splits.

\paragraph{BCNB~\cite{xu2021bcnb}.}
The BCNB cohort consists of 1,058 WSIs of early breast cancer core needle biopsies. On this cohort, we construct three biomarker prediction tasks, namely ER positivity prediction, PR positivity prediction, and HER2 positivity prediction. For all three tasks, we adopt label-stratified train/validation/test splits with 677, 169, and 212 slides, respectively.

\paragraph{BRACS~\cite{brancati2022bracs}.}
The BRACS cohort consists of 546 WSIs of different subtypes of breast cancer. We construct two subtype classification tasks on this cohort. The first is a coarse-grained 3-class classification task with three classes: benign, atypical, and malignant. The second is a fine-grained 7-class classification task with seven classes: Normal, benign, udh, fea, adh, dcis, and ic. For both tasks, we use the same label-stratified split of 394 training slides, 65 validation slides, and 87 test slides.

\paragraph{CPTAC~\cite{cptac_portal}.}
The CPTAC cohort consists of 2,059 pan-cancer WSIs from the following nine cancer types: Lung adenocarcinoma (LUAD), Head and neck squamous cell carcinoma (HNSC), Clear cell renal cell carcinoma (CCRCC), Colon adenocarcinoma (COAD), Lung squamous cell carcinoma (LSCC), Ovarian cancer (OV), Pancreatic ductal adenocarcinoma (PDA), Glioblastoma (GBM), and Breast cancer (BRCA). We construct two tasks on this cohort. The first is a 9-class pan-cancer classification task over the nine cancer types listed above, with a label-stratified split of 1313/340/406 slides for train/validation/test. The second is a binary NSCLC classification task, with a label-stratified split of 402/108/120 slides.

\paragraph{EBRAINS~\cite{roetzer2022digital}.}
The EBRAINS cohort consists of 2,310 WSIs of different brain tumour subtypes. We construct three tasks on this cohort. The first is a coarse-grained 12-class subtype classification task with the following classes: Metastatic tumours, Meningiomas, Embryonal tumours, Circumscribed astrocytic gliomas, Ependymal tumours, Mesenchymal, non-meningothelial tumours involving the CNS, Glioneuronal and neuronal tumours, Adult-type diffuse gliomas, Paediatric-type diffuse low-grade gliomas, Haematolymphoid tumours involving the CNS, Tumours of the sellar region, and Cranial and paraspinal nerve tumours. The corresponding train/validation/test split is 1470/359/481.

The second is a fine-grained 30-class subtype classification task with the following classes: Schwannoma, Ependymoma, Fibrous meningioma, Ganglioglioma, Anaplastic astrocytoma, IDH-mutant, Transitional meningioma, Meningothelial meningioma, Anaplastic ependymoma, Glioblastoma, IDH-wildtype, Atypical meningioma, Oligodendroglioma, IDH-mutant and 1p-19q codeleted, Lipoma, Diffuse large B-cell lymphoma of the CNS, Metastatic tumours, Haemangiopericytoma, Pilocytic astrocytoma, Adamantinomatous craniopharyngioma, Anaplastic meningioma, Medulloblastoma, non-WNT-non-SHH, Pituitary adenoma, Angiomatous meningioma, Diffuse astrocytoma, IDH-mutant, Glioblastoma, IDH-mutant, Haemangioblastoma, Anaplastic oligodendroglioma, IDH-mutant and 1p-19q codeleted, Haemangioma, Langerhans cell histiocytosis, Secretory meningioma, Gliosarcoma, and Anaplastic astrocytoma, IDH-wildtype. The corresponding split is 1455/363/492.

The third is an IDH biomarker mutation prediction task. We use a label-stratified train/validation/test split of 540/130/173 slides for this task.

\paragraph{KidRare~\cite{he2026boosting}.}
The KidRare cohort consists of 1,283 WSIs of rare pediatric tumours. We construct two classification tasks on this cohort. The first is a coarse-grained 4-class classification task over the following pediatric cancer types: Nephroblastoma, Neuroblastoma, Hepatoblastoma, and Medulloblastoma. The second is a fine-grained 13-class classification task with the following classes: Mixed epithelial and mesenchymal hepatoblastoma, Epithelial macrotrabecular pattern of hepatoblastoma, Tumor, Ganglioneuroblastoma, intermixed, Large Cell/Anaplastic medulloblastoma, Normal, Desmoplastic nodular medulloblastoma, unknown, Differentiating neuroblastoma, Poorly differentiated neuroblastoma, Epithelial mixed fetal and embryonal hepatoblastoma, Classic medulloblastoma, and Pure fetal hepatoblastoma with low mitotic activity. For both tasks, we use the same label-stratified split of 821/206/256 slides for train/validation/test.

\paragraph{MUT-HET-RCC~\cite{acosta2022intratumoral}.}
The MUT-HET-RCC cohort consists of 1,291 WSIs of clear cell renal cell carcinoma. On this cohort, we construct three mutation prediction tasks targeting BAP1, PBRM1, and SETD2. For all three tasks, we use label-stratified train/validation/test splits with 825, 207, and 259 slides, respectively.

\subsection{Results of linear probe}
\label{sec:appendix-linear}

\begin{table*}[h]
\centering
\caption{Linear-probe performance of teacher models on the 15 downstream datasets. Values are reported as mean with standard deviation in subscript.}
\label{tab:teacher_linear_probe_all_metrics}
\resizebox{\textwidth}{!}{%
\begin{tabular}{lcccccccc}
\toprule
Dataset 
& \multicolumn{4}{c}{CARE} 
& \multicolumn{4}{c}{TITAN} \\
\cmidrule(lr){2-5} \cmidrule(lr){6-9}
& Acc. & bacc & F1 & AUC 
& Acc. & bacc & F1 & AUC \\
\midrule
bcnb\_er & 78.3$_{0.0}$ & 50.0$_{0.0}$ & 43.9$_{0.0}$ & 80.8$_{0.8}$ & 79.1$_{1.1}$ & 51.7$_{2.5}$ & 47.3$_{4.8}$ & 80.1$_{1.8}$ \\
bcnb\_pr & 74.5$_{0.0}$ & 50.0$_{0.0}$ & 42.7$_{0.0}$ & 74.6$_{0.9}$ & 75.5$_{1.3}$ & 51.8$_{2.5}$ & 46.3$_{5.0}$ & 70.1$_{1.8}$ \\
bcnb\_her2 & 74.1$_{0.0}$ & 50.0$_{0.0}$ & 42.5$_{0.0}$ & 63.1$_{2.8}$ & 74.7$_{0.9}$ & 51.4$_{1.9}$ & 45.4$_{3.9}$ & 63.4$_{5.6}$ \\
bracs\_coarse & 56.5$_{1.9}$ & 51.2$_{1.7}$ & 44.0$_{1.5}$ & 74.9$_{2.3}$ & 57.5$_{11.0}$ & 52.1$_{10.0}$ & 43.5$_{13.2}$ & 83.8$_{5.3}$ \\
bracs\_fine & 40.5$_{1.3}$ & 28.0$_{0.8}$ & 18.9$_{0.1}$ & 70.4$_{2.3}$ & 41.4$_{0.0}$ & 28.6$_{0.0}$ & 18.9$_{0.1}$ & 79.2$_{1.6}$ \\
cptac\_nsclc & 87.0$_{7.2}$ & 86.8$_{7.3}$ & 86.5$_{8.1}$ & 99.4$_{0.5}$ & 94.3$_{5.0}$ & 94.2$_{5.1}$ & 94.2$_{5.1}$ & 99.9$_{0.2}$ \\
cptac\_pancancer & 89.0$_{0.5}$ & 86.7$_{0.6}$ & 87.2$_{0.4}$ & 98.8$_{0.0}$ & 92.4$_{0.3}$ & 91.5$_{0.2}$ & 91.9$_{0.4}$ & 99.6$_{0.0}$ \\
ebrains\_idh & 64.5$_{7.2}$ & 54.3$_{9.6}$ & 44.9$_{15.4}$ & 93.2$_{0.9}$ & 68.8$_{8.0}$ & 59.7$_{10.3}$ & 54.2$_{16.4}$ & 96.3$_{0.4}$ \\
ebrains\_coarse & 79.0$_{0.2}$ & 51.4$_{0.4}$ & 52.5$_{0.6}$ & 97.6$_{0.0}$ & 86.7$_{1.2}$ & 67.4$_{3.4}$ & 70.7$_{4.0}$ & 98.7$_{0.1}$ \\
ebrains\_fine & 56.9$_{0.9}$ & 41.3$_{1.1}$ & 40.0$_{1.3}$ & 96.8$_{0.0}$ & 56.3$_{18.7}$ & 43.3$_{20.4}$ & 42.8$_{20.4}$ & 97.4$_{1.1}$ \\
kidrare\_coarse & 77.3$_{1.3}$ & 66.9$_{2.1}$ & 65.9$_{2.8}$ & 96.0$_{0.1}$ & 83.9$_{2.9}$ & 73.5$_{6.4}$ & 73.3$_{8.4}$ & 98.1$_{0.2}$ \\
kidrare\_fine & 53.7$_{0.4}$ & 22.9$_{0.3}$ & 18.9$_{0.3}$ & 91.2$_{0.2}$ & 60.4$_{2.7}$ & 29.4$_{4.5}$ & 26.8$_{5.5}$ & 93.4$_{0.3}$ \\
mut\_het\_rcc\_bap1\_mutation & 87.3$_{0.0}$ & 50.0$_{0.0}$ & 46.6$_{0.0}$ & 80.0$_{1.3}$ & 87.3$_{0.0}$ & 50.0$_{0.0}$ & 46.6$_{0.0}$ & 79.6$_{1.0}$ \\
mut\_het\_rcc\_pbrm1\_mutation & 55.1$_{3.5}$ & 53.6$_{3.8}$ & 42.8$_{9.3}$ & 69.3$_{1.3}$ & 58.0$_{3.6}$ & 56.9$_{4.1}$ & 51.0$_{10.0}$ & 70.3$_{0.5}$ \\
mut\_het\_rcc\_setd2\_mutation & 73.0$_{0.0}$ & 50.0$_{0.0}$ & 42.2$_{0.0}$ & 68.8$_{4.9}$ & 73.0$_{0.2}$ & 50.2$_{0.5}$ & 42.8$_{1.2}$ & 67.6$_{5.8}$ \\
\bottomrule
\end{tabular}%
}
\end{table*}

\begin{table*}[h]
\centering
\caption{Balanced accuracy (bacc) of teacher models and distillation-initialized MIL models under the linear-probe setting. For each dataset, $\Delta$ denotes the difference between each MIL model and the best teacher result on the same dataset. Values are reported as mean with standard deviation in subscript.}
\label{tab:linear_probe_teacher_distill_bacc_delta}
\resizebox{\textwidth}{!}{%
\begin{tabular}{lcclccccccccc}
\toprule
\multirow{2}{*}{Dataset}
& \multicolumn{2}{c}{Teacher Models}
& \multirow{2}{*}{Metric}
& \multicolumn{9}{c}{Pretrained Student MIL Aggregators} \\
\cmidrule(lr){2-3} \cmidrule(lr){5-13}
& TITAN & CARE
& & ABMIL & CLAM & TransMIL & WiKG & DAGMIL & GDFMIL & 2DMamba & AMDMIL & AEMMIL \\
\midrule

\multirow{2}{*}{bcnb\_er}
& \multirow{2}{*}{\score{51.7}{2.5}}
& \multirow{2}{*}{\score{50.0}{0.0}}
& BACC
& \score{74.0}{1.2} & \score{74.4}{2.2} & \score{74.6}{2.8} & \score{71.5}{1.9} & \score{70.6}{1.7} & \score{50.0}{0.0} & \score{50.0}{0.0} & \score{66.3}{1.1} & \score{66.2}{3.2} \\
& & & $\Delta$
& \posval{+22.3} & \posval{+22.7} & \posval{+22.9} & \posval{+19.8} & \posval{+18.9} & \negval{-1.7} & \negval{-1.7} & \posval{+14.6} & \posval{+14.5} \\

\multirow{2}{*}{bcnb\_pr}
& \multirow{2}{*}{\score{51.8}{2.5}}
& \multirow{2}{*}{\score{50.0}{0.0}}
& BACC
& \score{71.3}{1.4} & \score{70.5}{3.5} & \score{72.5}{2.0} & \score{69.5}{3.1} & \score{68.8}{2.0} & \score{50.0}{0.0} & \score{50.0}{0.0} & \score{68.3}{1.4} & \score{64.8}{1.7} \\
& & & $\Delta$
& \posval{+19.5} & \posval{+18.7} & \posval{+20.7} & \posval{+17.7} & \posval{+17.0} & \negval{-1.8} & \negval{-1.8} & \posval{+16.5} & \posval{+13.0} \\

\multirow{2}{*}{bcnb\_her2}
& \multirow{2}{*}{\score{51.4}{1.9}}
& \multirow{2}{*}{\score{50.0}{0.0}}
& BACC
& \score{64.0}{1.3} & \score{63.2}{1.5} & \score{64.7}{2.0} & \score{62.8}{1.9} & \score{64.6}{2.0} & \score{50.0}{0.0} & \score{50.6}{1.2} & \score{65.6}{1.5} & \score{60.8}{2.7} \\
& & & $\Delta$
& \posval{+12.6} & \posval{+11.8} & \posval{+13.3} & \posval{+11.4} & \posval{+13.2} & \negval{-1.4} & \negval{-0.8} & \posval{+14.2} & \posval{+9.4} \\

\multirow{2}{*}{bracs\_coarse}
& \multirow{2}{*}{\score{52.1}{10.0}}
& \multirow{2}{*}{\score{51.2}{1.7}}
& BACC
& \score{58.1}{3.4} & \score{61.5}{2.3} & \score{56.2}{1.1} & \score{62.6}{1.5} & \score{54.6}{1.9} & \score{39.4}{7.0} & \score{43.8}{3.1} & \score{65.0}{2.4} & \score{52.1}{3.4} \\
& & & $\Delta$
& \posval{+6.0} & \posval{+9.4} & \posval{+4.1} & \posval{+10.5} & \posval{+2.5} & \negval{-12.7} & \negval{-8.3} & \posval{+12.9} & 0.0 \\

\multirow{2}{*}{bracs\_fine}
& \multirow{2}{*}{\score{28.6}{0.0}}
& \multirow{2}{*}{\score{28.0}{0.8}}
& BACC
& \score{32.3}{3.3} & \score{30.4}{4.7} & \score{37.5}{8.2} & \score{32.4}{6.6} & \score{35.4}{4.3} & \score{27.7}{0.5} & \score{20.9}{0.7} & \score{41.2}{7.2} & \score{24.6}{1.0} \\
& & & $\Delta$
& \posval{+3.7} & \posval{+1.8} & \posval{+8.9} & \posval{+3.8} & \posval{+6.8} & \negval{-0.9} & \negval{-7.7} & \posval{+12.6} & \negval{-4.0} \\

\multirow{2}{*}{cptac\_nsclc}
& \multirow{2}{*}{\score{94.2}{5.1}}
& \multirow{2}{*}{\score{86.8}{7.3}}
& BACC
& \score{94.3}{0.6} & \score{96.7}{0.5} & \score{95.8}{0.0} & \score{98.5}{0.6} & \score{96.5}{0.6} & \score{93.5}{2.6} & \score{91.9}{1.7} & \score{93.8}{0.4} & \score{97.0}{1.0} \\
& & & $\Delta$
& \posval{+0.1} & \posval{+2.5} & \posval{+1.6} & \posval{+4.3} & \posval{+2.3} & \negval{-0.7} & \negval{-2.3} & \negval{-0.4} & \posval{+2.8} \\

\multirow{2}{*}{cptac\_pancancer}
& \multirow{2}{*}{\score{91.5}{0.2}}
& \multirow{2}{*}{\score{86.7}{0.6}}
& BACC
& \score{92.2}{1.3} & \score{92.2}{2.1} & \score{93.8}{0.4} & \score{92.4}{1.8} & \score{92.1}{1.1} & \score{88.6}{0.3} & \score{88.3}{0.7} & \score{94.5}{0.4} & \score{91.3}{1.2} \\
& & & $\Delta$
& \posval{+0.7} & \posval{+0.7} & \posval{+2.3} & \posval{+0.9} & \posval{+0.6} & \negval{-2.9} & \negval{-3.2} & \posval{+3.0} & \negval{-0.2} \\

\multirow{2}{*}{ebrains\_idh}
& \multirow{2}{*}{\score{59.7}{10.3}}
& \multirow{2}{*}{\score{54.3}{9.6}}
& BACC
& \score{91.4}{0.8} & \score{91.9}{1.8} & \score{92.0}{1.8} & \score{92.7}{1.5} & \score{92.2}{1.9} & \score{89.1}{3.0} & \score{90.2}{0.5} & \score{93.5}{0.4} & \score{92.2}{1.8} \\
& & & $\Delta$
& \posval{+31.7} & \posval{+32.2} & \posval{+32.3} & \posval{+33.0} & \posval{+32.5} & \posval{+29.4} & \posval{+30.5} & \posval{+33.8} & \posval{+32.5} \\

\multirow{2}{*}{ebrains\_coarse}
& \multirow{2}{*}{\score{67.4}{3.4}}
& \multirow{2}{*}{\score{51.4}{0.4}}
& BACC
& \score{83.8}{0.8} & \score{85.1}{2.0} & \score{85.9}{1.5} & \score{85.1}{1.5} & \score{84.2}{1.5} & \score{70.7}{2.3} & \score{79.8}{3.6} & \score{86.2}{1.0} & \score{82.4}{1.0} \\
& & & $\Delta$
& \posval{+16.4} & \posval{+17.7} & \posval{+18.5} & \posval{+17.7} & \posval{+16.8} & \posval{+3.3} & \posval{+12.4} & \posval{+18.8} & \posval{+15.0} \\

\multirow{2}{*}{ebrains\_fine}
& \multirow{2}{*}{\score{43.3}{20.4}}
& \multirow{2}{*}{\score{41.3}{1.1}}
& BACC
& \score{64.6}{1.5} & \score{65.5}{1.4} & \score{67.0}{1.7} & \score{66.8}{2.7} & \score{63.2}{0.2} & \score{55.4}{4.3} & \score{57.3}{1.8} & \score{66.9}{1.4} & \score{64.3}{2.5} \\
& & & $\Delta$
& \posval{+21.3} & \posval{+22.2} & \posval{+23.7} & \posval{+23.5} & \posval{+19.9} & \posval{+12.1} & \posval{+14.0} & \posval{+23.6} & \posval{+21.0} \\

\multirow{2}{*}{kidrare\_coarse}
& \multirow{2}{*}{\score{73.5}{6.4}}
& \multirow{2}{*}{\score{66.9}{2.1}}
& BACC
& \score{84.4}{1.8} & \score{85.8}{1.2} & \score{83.9}{1.6} & \score{82.6}{0.7} & \score{74.1}{2.2} & \score{61.6}{3.3} & \score{63.0}{2.4} & \score{84.2}{3.5} & \score{77.6}{3.0} \\
& & & $\Delta$
& \posval{+10.9} & \posval{+12.3} & \posval{+10.4} & \posval{+9.1} & \posval{+0.6} & \negval{-11.9} & \negval{-10.5} & \posval{+10.7} & \posval{+4.1} \\

\multirow{2}{*}{kidrare\_fine}
& \multirow{2}{*}{\score{29.4}{4.5}}
& \multirow{2}{*}{\score{22.9}{0.3}}
& BACC
& \score{37.6}{1.8} & \score{39.5}{2.1} & \score{39.3}{3.7} & \score{38.6}{2.4} & \score{33.5}{2.5} & \score{22.3}{3.4} & \score{21.6}{2.7} & \score{37.9}{2.2} & \score{33.2}{3.0} \\
& & & $\Delta$
& \posval{+8.2} & \posval{+10.1} & \posval{+9.9} & \posval{+9.2} & \posval{+4.1} & \negval{-7.1} & \negval{-7.8} & \posval{+8.5} & \posval{+3.8} \\

\multirow{2}{*}{mut\_het\_rcc\_bap1\_mutation}
& \multirow{2}{*}{\score{50.0}{0.0}}
& \multirow{2}{*}{\score{50.0}{0.0}}
& BACC
& \score{61.2}{3.2} & \score{63.2}{3.1} & \score{60.9}{1.8} & \score{61.3}{1.8} & \score{55.2}{2.8} & \score{50.0}{0.0} & \score{50.0}{0.0} & \score{57.6}{4.0} & \score{53.6}{2.1} \\
& & & $\Delta$
& \posval{+11.2} & \posval{+13.2} & \posval{+10.9} & \posval{+11.3} & \posval{+5.2} & 0.0 & 0.0 & \posval{+7.6} & \posval{+3.6} \\

\multirow{2}{*}{mut\_het\_rcc\_pbrm1\_mutation}
& \multirow{2}{*}{\score{56.9}{4.1}}
& \multirow{2}{*}{\score{53.6}{3.8}}
& BACC
& \score{66.4}{1.0} & \score{66.8}{0.5} & \score{62.5}{1.5} & \score{64.3}{2.6} & \score{64.6}{1.0} & \score{58.3}{1.7} & \score{58.6}{3.9} & \score{64.7}{2.4} & \score{63.3}{2.2} \\
& & & $\Delta$
& \posval{+9.5} & \posval{+9.9} & \posval{+5.6} & \posval{+7.4} & \posval{+7.7} & \posval{+1.4} & \posval{+1.7} & \posval{+7.8} & \posval{+6.4} \\

\multirow{2}{*}{mut\_het\_rcc\_setd2\_mutation}
& \multirow{2}{*}{\score{50.2}{0.5}}
& \multirow{2}{*}{\score{50.0}{0.0}}
& BACC
& \score{60.2}{2.7} & \score{61.9}{3.3} & \score{62.5}{1.8} & \score{62.1}{0.3} & \score{59.4}{1.0} & \score{50.0}{0.0} & \score{49.8}{0.2} & \score{60.4}{1.9} & \score{51.3}{1.4} \\
& & & $\Delta$
& \posval{+10.0} & \posval{+11.7} & \posval{+12.3} & \posval{+11.9} & \posval{+9.2} & \negval{-0.2} & \negval{-0.4} & \posval{+10.2} & \posval{+1.1} \\

\bottomrule
\end{tabular}%
}
\end{table*}

\begin{table*}[h]
\centering
\caption{Accuracy of teacher models and distillation-initialized MIL models under the linear-probe setting. For each dataset, $\Delta$ denotes the difference between each MIL model and the best teacher result on the same dataset. Values are reported as mean with standard deviation in subscript.}
\label{tab:linear_probe_teacher_distill_acc_delta}
\resizebox{\textwidth}{!}{%
\begin{tabular}{lcclccccccccc}
\toprule
\multirow{2}{*}{Dataset}
& \multicolumn{2}{c}{Teacher Models}
& \multirow{2}{*}{Metric}
& \multicolumn{9}{c}{Pretrained Student MIL Aggregators} \\
\cmidrule(lr){2-3} \cmidrule(lr){5-13}
& TITAN & CARE
& & ABMIL & CLAM & TransMIL & WiKG & DAGMIL & GDFMIL & 2DMamba & AMDMIL & AEMMIL \\
\midrule

\multirow{2}{*}{bcnb\_er}
& \multirow{2}{*}{\score{79.1}{1.1}}
& \multirow{2}{*}{\score{78.3}{0.0}}
& ACC
& \score{81.6}{2.0} & \score{83.5}{0.5} & \score{82.5}{1.0} & \score{83.1}{1.1} & \score{83.8}{1.0} & \score{78.3}{0.0} & \score{78.3}{0.0} & \score{81.4}{1.6} & \score{82.0}{0.8} \\
& & & $\Delta$
& \posval{+2.5} & \posval{+4.4} & \posval{+3.4} & \posval{+4.0} & \posval{+4.7} & \negval{-0.8} & \negval{-0.8} & \posval{+2.3} & \posval{+2.9} \\

\multirow{2}{*}{bcnb\_pr}
& \multirow{2}{*}{\score{75.5}{1.3}}
& \multirow{2}{*}{\score{74.5}{0.0}}
& ACC
& \score{79.3}{1.0} & \score{79.0}{2.6} & \score{79.1}{1.9} & \score{78.5}{1.4} & \score{80.0}{0.9} & \score{74.5}{0.0} & \score{74.5}{0.0} & \score{77.2}{1.8} & \score{77.4}{0.3} \\
& & & $\Delta$
& \posval{+3.8} & \posval{+3.5} & \posval{+3.6} & \posval{+3.0} & \posval{+4.5} & \negval{-1.0} & \negval{-1.0} & \posval{+1.7} & \posval{+1.9} \\

\multirow{2}{*}{bcnb\_her2}
& \multirow{2}{*}{\score{74.7}{0.9}}
& \multirow{2}{*}{\score{74.1}{0.0}}
& ACC
& \score{68.0}{3.9} & \score{69.5}{4.2} & \score{69.2}{6.9} & \score{67.9}{7.0} & \score{73.6}{0.8} & \score{73.9}{0.0} & \score{74.1}{0.4} & \score{73.6}{1.5} & \score{72.8}{1.6} \\
& & & $\Delta$
& \negval{-6.7} & \negval{-5.2} & \negval{-5.5} & \negval{-6.8} & \negval{-1.1} & \negval{-0.8} & \negval{-0.6} & \negval{-1.1} & \negval{-1.9} \\

\multirow{2}{*}{bracs\_coarse}
& \multirow{2}{*}{\score{57.5}{11.0}}
& \multirow{2}{*}{\score{56.5}{1.9}}
& ACC
& \score{63.4}{3.3} & \score{66.4}{2.0} & \score{62.1}{1.3} & \score{67.6}{1.3} & \score{60.2}{2.1} & \score{43.5}{7.7} & \score{48.3}{3.4} & \score{70.6}{2.1} & \score{57.5}{3.8} \\
& & & $\Delta$
& \posval{+5.9} & \posval{+8.9} & \posval{+4.6} & \posval{+10.1} & \posval{+2.7} & \negval{-14.0} & \negval{-9.2} & \posval{+13.1} & 0.0 \\

\multirow{2}{*}{bracs\_fine}
& \multirow{2}{*}{\score{41.4}{0.0}}
& \multirow{2}{*}{\score{40.5}{1.3}}
& ACC
& \score{40.5}{2.0} & \score{39.3}{3.7} & \score{43.9}{6.3} & \score{40.9}{3.9} & \score{41.5}{3.9} & \score{35.9}{0.8} & \score{29.9}{0.7} & \score{48.3}{6.2} & \score{34.9}{1.6} \\
& & & $\Delta$
& \negval{-0.9} & \negval{-2.1} & \posval{+2.5} & \negval{-0.5} & \posval{+0.1} & \negval{-5.5} & \negval{-11.5} & \posval{+6.9} & \negval{-6.5} \\

\multirow{2}{*}{cptac\_nsclc}
& \multirow{2}{*}{\score{94.3}{5.0}}
& \multirow{2}{*}{\score{87.0}{7.2}}
& ACC
& \score{94.3}{0.6} & \score{96.6}{0.5} & \score{95.8}{0.0} & \score{98.5}{0.6} & \score{96.5}{0.6} & \score{93.5}{2.7} & \score{91.8}{1.7} & \score{93.8}{0.4} & \score{97.0}{1.0} \\
& & & $\Delta$
& 0.0 & \posval{+2.3} & \posval{+1.5} & \posval{+4.2} & \posval{+2.2} & \negval{-0.8} & \negval{-2.5} & \negval{-0.5} & \posval{+2.7} \\

\multirow{2}{*}{cptac\_pancancer}
& \multirow{2}{*}{\score{92.4}{0.3}}
& \multirow{2}{*}{\score{89.0}{0.5}}
& ACC
& \score{93.0}{0.9} & \score{92.8}{1.8} & \score{94.4}{0.6} & \score{93.1}{1.4} & \score{92.6}{0.9} & \score{89.5}{0.4} & \score{90.1}{0.6} & \score{95.3}{0.3} & \score{92.4}{1.0} \\
& & & $\Delta$
& \posval{+0.6} & \posval{+0.4} & \posval{+2.0} & \posval{+0.7} & \posval{+0.2} & \negval{-2.9} & \negval{-2.3} & \posval{+2.9} & 0.0 \\

\multirow{2}{*}{ebrains\_idh}
& \multirow{2}{*}{\score{68.8}{8.0}}
& \multirow{2}{*}{\score{64.5}{7.2}}
& ACC
& \score{91.4}{1.1} & \score{91.5}{2.3} & \score{91.5}{2.3} & \score{92.3}{1.3} & \score{92.3}{1.7} & \score{88.9}{3.7} & \score{90.3}{0.5} & \score{93.6}{0.5} & \score{92.8}{1.4} \\
& & & $\Delta$
& \posval{+22.6} & \posval{+22.7} & \posval{+22.7} & \posval{+23.5} & \posval{+23.5} & \posval{+20.1} & \posval{+21.5} & \posval{+24.8} & \posval{+24.0} \\

\multirow{2}{*}{ebrains\_coarse}
& \multirow{2}{*}{\score{86.7}{1.2}}
& \multirow{2}{*}{\score{79.0}{0.2}}
& ACC
& \score{87.5}{1.0} & \score{87.4}{1.0} & \score{88.8}{1.7} & \score{87.6}{1.5} & \score{88.4}{0.6} & \score{84.8}{0.8} & \score{87.5}{1.6} & \score{90.2}{0.6} & \score{88.3}{0.7} \\
& & & $\Delta$
& \posval{+0.8} & \posval{+0.7} & \posval{+2.1} & \posval{+0.9} & \posval{+1.7} & \negval{-1.9} & \posval{+0.8} & \posval{+3.5} & \posval{+1.6} \\

\multirow{2}{*}{ebrains\_fine}
& \multirow{2}{*}{\score{56.3}{18.7}}
& \multirow{2}{*}{\score{56.9}{0.9}}
& ACC
& \score{71.4}{1.4} & \score{71.8}{0.4} & \score{73.9}{1.4} & \score{72.7}{1.6} & \score{70.6}{1.0} & \score{66.3}{2.4} & \score{66.9}{1.5} & \score{73.9}{0.7} & \score{72.0}{1.7} \\
& & & $\Delta$
& \posval{+14.5} & \posval{+14.9} & \posval{+17.0} & \posval{+15.8} & \posval{+13.7} & \posval{+9.4} & \posval{+10.0} & \posval{+17.0} & \posval{+15.1} \\

\multirow{2}{*}{kidrare\_coarse}
& \multirow{2}{*}{\score{83.9}{2.9}}
& \multirow{2}{*}{\score{77.3}{1.3}}
& ACC
& \score{83.6}{3.0} & \score{86.1}{1.2} & \score{84.9}{1.2} & \score{83.7}{0.6} & \score{77.2}{1.5} & \score{70.3}{1.7} & \score{72.4}{1.4} & \score{86.5}{1.8} & \score{81.2}{2.4} \\
& & & $\Delta$
& \negval{-0.3} & \posval{+2.2} & \posval{+1.0} & \negval{-0.2} & \negval{-6.7} & \negval{-13.6} & \negval{-11.5} & \posval{+2.6} & \negval{-2.7} \\

\multirow{2}{*}{kidrare\_fine}
& \multirow{2}{*}{\score{60.4}{2.7}}
& \multirow{2}{*}{\score{53.7}{0.4}}
& ACC
& \score{57.3}{1.3} & \score{59.9}{1.0} & \score{63.5}{2.1} & \score{60.7}{3.0} & \score{60.4}{1.5} & \score{50.8}{3.4} & \score{49.0}{3.2} & \score{62.3}{1.7} & \score{59.9}{1.8} \\
& & & $\Delta$
& \negval{-3.1} & \negval{-0.5} & \posval{+3.1} & \posval{+0.3} & 0.0 & \negval{-9.6} & \negval{-11.4} & \posval{+1.9} & \negval{-0.5} \\

\multirow{2}{*}{mut\_het\_rcc\_bap1\_mutation}
& \multirow{2}{*}{\score{87.3}{0.0}}
& \multirow{2}{*}{\score{87.3}{0.0}}
& ACC
& \score{86.8}{0.7} & \score{86.6}{0.4} & \score{87.0}{0.4} & \score{86.5}{0.2} & \score{87.4}{0.6} & \score{87.0}{0.0} & \score{87.0}{0.0} & \score{87.2}{0.4} & \score{87.4}{0.3} \\
& & & $\Delta$
& \negval{-0.5} & \negval{-0.7} & \negval{-0.3} & \negval{-0.8} & \posval{+0.1} & \negval{-0.3} & \negval{-0.3} & \negval{-0.1} & \posval{+0.1} \\

\multirow{2}{*}{mut\_het\_rcc\_pbrm1\_mutation}
& \multirow{2}{*}{\score{58.0}{3.6}}
& \multirow{2}{*}{\score{55.1}{3.5}}
& ACC
& \score{66.3}{1.1} & \score{66.6}{0.5} & \score{62.7}{1.5} & \score{64.3}{2.6} & \score{64.7}{0.9} & \score{58.1}{2.0} & \score{58.8}{3.7} & \score{64.6}{2.3} & \score{63.5}{2.2} \\
& & & $\Delta$
& \posval{+8.3} & \posval{+8.6} & \posval{+4.7} & \posval{+6.3} & \posval{+6.7} & \posval{+0.1} & \posval{+0.8} & \posval{+6.6} & \posval{+5.5} \\

\multirow{2}{*}{mut\_het\_rcc\_setd2\_mutation}
& \multirow{2}{*}{\score{73.0}{0.2}}
& \multirow{2}{*}{\score{73.0}{0.0}}
& ACC
& \score{72.5}{0.6} & \score{71.5}{3.2} & \score{72.9}{2.7} & \score{73.8}{1.8} & \score{74.2}{0.6} & \score{72.3}{0.0} & \score{72.1}{0.3} & \score{73.4}{0.7} & \score{72.2}{0.4} \\
& & & $\Delta$
& \negval{-0.5} & \negval{-1.5} & \negval{-0.1} & \posval{+0.8} & \posval{+1.2} & \negval{-0.7} & \negval{-0.9} & \posval{+0.4} & \negval{-0.8} \\

\bottomrule
\end{tabular}%
}
\end{table*}

\begin{table*}[h]
\centering
\caption{AUC of teacher models and distillation-initialized MIL models under the linear-probe setting. For each dataset, $\Delta$ denotes the difference between each MIL model and the best teacher result on the same dataset. Values are reported as mean with standard deviation in subscript.}
\label{tab:linear_probe_teacher_distill_auc_delta}
\resizebox{\textwidth}{!}{%
\begin{tabular}{lcclccccccccc}
\toprule
\multirow{2}{*}{Dataset}
& \multicolumn{2}{c}{Teacher Models}
& \multirow{2}{*}{Metric}
& \multicolumn{9}{c}{Pretrained Student MIL Aggregators} \\
\cmidrule(lr){2-3} \cmidrule(lr){5-13}
& TITAN & CARE
& & ABMIL & CLAM & TransMIL & WiKG & DAGMIL & GDFMIL & 2DMamba & AMDMIL & AEMMIL \\
\midrule

\multirow{2}{*}{bcnb\_er}
& \multirow{2}{*}{\score{80.1}{1.8}}
& \multirow{2}{*}{\score{80.8}{0.8}}
& AUC
& \score{83.3}{1.1} & \score{86.2}{0.5} & \score{84.4}{0.7} & \score{83.5}{0.4} & \score{83.0}{0.6} & \score{55.9}{5.3} & \score{51.0}{3.1} & \score{83.6}{1.3} & \score{81.8}{0.9} \\
& & & $\Delta$
& \posval{+2.5} & \posval{+5.4} & \posval{+3.6} & \posval{+2.7} & \posval{+2.2} & \negval{-24.9} & \negval{-29.8} & \posval{+2.8} & \posval{+1.0} \\

\multirow{2}{*}{bcnb\_pr}
& \multirow{2}{*}{\score{70.1}{1.8}}
& \multirow{2}{*}{\score{74.6}{0.9}}
& AUC
& \score{76.7}{1.5} & \score{76.7}{2.3} & \score{78.4}{0.9} & \score{77.3}{1.1} & \score{76.7}{0.7} & \score{55.3}{3.4} & \score{52.9}{3.0} & \score{76.3}{1.2} & \score{72.9}{0.9} \\
& & & $\Delta$
& \posval{+2.1} & \posval{+2.1} & \posval{+3.8} & \posval{+2.7} & \posval{+2.1} & \negval{-19.3} & \negval{-21.7} & \posval{+1.7} & \negval{-1.7} \\

\multirow{2}{*}{bcnb\_her2}
& \multirow{2}{*}{\score{63.4}{5.6}}
& \multirow{2}{*}{\score{63.1}{2.8}}
& AUC
& \score{69.2}{1.4} & \score{68.2}{1.8} & \score{71.6}{1.4} & \score{69.8}{0.8} & \score{71.7}{1.0} & \score{56.5}{2.9} & \score{61.7}{4.3} & \score{72.8}{1.6} & \score{68.9}{1.1} \\
& & & $\Delta$
& \posval{+5.8} & \posval{+4.8} & \posval{+8.2} & \posval{+6.4} & \posval{+8.3} & \negval{-6.9} & \negval{-1.7} & \posval{+9.4} & \posval{+5.5} \\

\multirow{2}{*}{bracs\_coarse}
& \multirow{2}{*}{\score{83.8}{5.3}}
& \multirow{2}{*}{\score{74.9}{2.3}}
& AUC
& \score{83.0}{1.2} & \score{86.5}{1.9} & \score{81.9}{1.7} & \score{86.0}{1.1} & \score{78.1}{3.7} & \score{67.6}{6.5} & \score{64.7}{4.0} & \score{88.4}{0.6} & \score{78.6}{1.6} \\
& & & $\Delta$
& \negval{-0.8} & \posval{+2.7} & \negval{-1.9} & \posval{+2.2} & \negval{-5.7} & \negval{-16.2} & \negval{-19.1} & \posval{+4.6} & \negval{-5.2} \\

\multirow{2}{*}{bracs\_fine}
& \multirow{2}{*}{\score{79.2}{1.6}}
& \multirow{2}{*}{\score{70.4}{2.3}}
& AUC
& \score{76.5}{2.3} & \score{77.1}{2.9} & \score{72.8}{6.1} & \score{71.9}{5.9} & \score{77.5}{3.5} & \score{69.1}{1.9} & \score{55.0}{1.7} & \score{80.6}{3.1} & \score{67.5}{1.5} \\
& & & $\Delta$
& \negval{-2.7} & \negval{-2.1} & \negval{-6.4} & \negval{-7.3} & \negval{-1.7} & \negval{-10.1} & \negval{-24.2} & \posval{+1.4} & \negval{-11.7} \\

\multirow{2}{*}{cptac\_nsclc}
& \multirow{2}{*}{\score{99.9}{0.2}}
& \multirow{2}{*}{\score{99.4}{0.5}}
& AUC
& \score{99.2}{0.1} & \score{99.7}{0.1} & \score{98.7}{0.1} & \score{99.8}{0.1} & \score{99.7}{0.0} & \score{99.1}{0.6} & \score{97.6}{1.2} & \score{99.4}{0.1} & \score{99.6}{0.0} \\
& & & $\Delta$
& \negval{-0.7} & \negval{-0.2} & \negval{-1.2} & \negval{-0.1} & \negval{-0.2} & \negval{-0.8} & \negval{-2.3} & \negval{-0.5} & \negval{-0.3} \\

\multirow{2}{*}{cptac\_pancancer}
& \multirow{2}{*}{\score{99.6}{0.0}}
& \multirow{2}{*}{\score{98.8}{0.0}}
& AUC
& \score{99.7}{0.1} & \score{99.6}{0.1} & \score{99.8}{0.0} & \score{99.6}{0.2} & \score{99.6}{0.1} & \score{99.1}{0.1} & \score{99.3}{0.1} & \score{99.8}{0.0} & \score{99.6}{0.1} \\
& & & $\Delta$
& \posval{+0.1} & 0.0 & \posval{+0.2} & 0.0 & 0.0 & \negval{-0.5} & \negval{-0.3} & \posval{+0.2} & 0.0 \\

\multirow{2}{*}{ebrains\_idh}
& \multirow{2}{*}{\score{96.3}{0.4}}
& \multirow{2}{*}{\score{93.2}{0.9}}
& AUC
& \score{97.8}{0.4} & \score{98.0}{0.9} & \score{98.4}{0.3} & \score{98.5}{0.3} & \score{98.2}{0.4} & \score{96.6}{0.5} & \score{96.7}{0.6} & \score{98.6}{0.1} & \score{98.2}{0.2} \\
& & & $\Delta$
& \posval{+1.5} & \posval{+1.7} & \posval{+2.1} & \posval{+2.2} & \posval{+1.9} & \posval{+0.3} & \posval{+0.4} & \posval{+2.3} & \posval{+1.9} \\

\multirow{2}{*}{ebrains\_coarse}
& \multirow{2}{*}{\score{98.7}{0.1}}
& \multirow{2}{*}{\score{97.6}{0.0}}
& AUC
& \score{98.6}{0.2} & \score{98.8}{0.1} & \score{99.2}{0.1} & \score{99.1}{0.1} & \score{98.8}{0.1} & \score{97.9}{0.2} & \score{97.8}{0.3} & \score{99.3}{0.0} & \score{99.0}{0.1} \\
& & & $\Delta$
& \negval{-0.1} & \posval{+0.1} & \posval{+0.5} & \posval{+0.4} & \posval{+0.1} & \negval{-0.8} & \negval{-0.9} & \posval{+0.6} & \posval{+0.3} \\

\multirow{2}{*}{ebrains\_fine}
& \multirow{2}{*}{\score{97.4}{1.1}}
& \multirow{2}{*}{\score{96.8}{0.0}}
& AUC
& \score{98.2}{0.1} & \score{98.2}{0.1} & \score{98.4}{0.1} & \score{98.2}{0.0} & \score{98.1}{0.1} & \score{97.5}{0.3} & \score{97.2}{0.3} & \score{98.4}{0.1} & \score{98.1}{0.1} \\
& & & $\Delta$
& \posval{+0.8} & \posval{+0.8} & \posval{+1.0} & \posval{+0.8} & \posval{+0.7} & \posval{+0.1} & \negval{-0.2} & \posval{+1.0} & \posval{+0.7} \\

\multirow{2}{*}{kidrare\_coarse}
& \multirow{2}{*}{\score{98.1}{0.2}}
& \multirow{2}{*}{\score{96.0}{0.1}}
& AUC
& \score{96.8}{0.2} & \score{97.1}{0.1} & \score{97.0}{0.3} & \score{96.3}{0.1} & \score{94.3}{0.7} & \score{93.4}{0.6} & \score{92.2}{0.8} & \score{97.5}{0.2} & \score{96.3}{0.4} \\
& & & $\Delta$
& \negval{-1.3} & \negval{-1.0} & \negval{-1.1} & \negval{-1.8} & \negval{-3.8} & \negval{-4.7} & \negval{-5.9} & \negval{-0.6} & \negval{-1.8} \\

\multirow{2}{*}{kidrare\_fine}
& \multirow{2}{*}{\score{93.4}{0.3}}
& \multirow{2}{*}{\score{91.2}{0.2}}
& AUC
& \score{91.6}{0.6} & \score{92.0}{0.4} & \score{92.6}{0.4} & \score{92.6}{0.5} & \score{91.0}{0.4} & \score{84.6}{3.0} & \score{86.3}{1.1} & \score{92.7}{0.5} & \score{90.3}{0.9} \\
& & & $\Delta$
& \negval{-1.8} & \negval{-1.4} & \negval{-0.8} & \negval{-0.8} & \negval{-2.4} & \negval{-8.8} & \negval{-7.1} & \negval{-0.7} & \negval{-3.1} \\

\multirow{2}{*}{mut\_het\_rcc\_bap1\_mutation}
& \multirow{2}{*}{\score{79.6}{1.0}}
& \multirow{2}{*}{\score{80.0}{1.3}}
& AUC
& \score{83.7}{0.3} & \score{84.0}{0.3} & \score{83.3}{0.4} & \score{82.8}{0.6} & \score{81.8}{1.2} & \score{50.3}{14.5} & \score{63.6}{11.9} & \score{80.8}{2.8} & \score{82.2}{1.6} \\
& & & $\Delta$
& \posval{+3.7} & \posval{+4.0} & \posval{+3.3} & \posval{+2.8} & \posval{+1.8} & \negval{-29.7} & \negval{-16.4} & \posval{+0.8} & \posval{+2.2} \\

\multirow{2}{*}{mut\_het\_rcc\_pbrm1\_mutation}
& \multirow{2}{*}{\score{70.3}{0.5}}
& \multirow{2}{*}{\score{69.3}{1.3}}
& AUC
& \score{71.8}{1.3} & \score{70.9}{1.4} & \score{68.5}{0.8} & \score{70.8}{2.6} & \score{70.7}{0.5} & \score{62.3}{2.6} & \score{63.4}{1.7} & \score{69.8}{2.7} & \score{69.3}{2.0} \\
& & & $\Delta$
& \posval{+1.5} & \posval{+0.6} & \negval{-1.8} & \posval{+0.5} & \posval{+0.4} & \negval{-8.0} & \negval{-6.9} & \negval{-0.5} & \negval{-1.0} \\

\multirow{2}{*}{mut\_het\_rcc\_setd2\_mutation}
& \multirow{2}{*}{\score{67.6}{5.8}}
& \multirow{2}{*}{\score{68.8}{4.9}}
& AUC
& \score{68.3}{0.6} & \score{68.5}{0.4} & \score{69.3}{0.3} & \score{70.3}{1.9} & \score{70.6}{0.7} & \score{52.8}{7.8} & \score{66.1}{7.9} & \score{69.9}{1.1} & \score{64.7}{7.0} \\
& & & $\Delta$
& \negval{-0.5} & \negval{-0.3} & \posval{+0.5} & \posval{+1.5} & \posval{+1.8} & \negval{-16.0} & \negval{-2.7} & \posval{+1.1} & \negval{-4.1} \\

\bottomrule
\end{tabular}%
}
\end{table*}

\begin{table*}[h]
\centering
\caption{F1 score of teacher models and distillation-initialized MIL models under the linear-probe setting. For each dataset, $\Delta$ denotes the difference between each MIL model and the best teacher result on the same dataset. Values are reported as mean with standard deviation in subscript.}
\label{tab:linear_probe_teacher_distill_f1_delta}
\resizebox{\textwidth}{!}{%
\begin{tabular}{lcclccccccccc}
\toprule
\multirow{2}{*}{Dataset}
& \multicolumn{2}{c}{Teacher Models}
& \multirow{2}{*}{Metric}
& \multicolumn{9}{c}{Pretrained Student MIL Aggregators} \\
\cmidrule(lr){2-3} \cmidrule(lr){5-13}
& TITAN & CARE
& & ABMIL & CLAM & TransMIL & WiKG & DAGMIL & GDFMIL & 2DMamba & AMDMIL & AEMMIL \\
\midrule

\multirow{2}{*}{bcnb\_er}
& \multirow{2}{*}{\score{47.3}{4.8}}
& \multirow{2}{*}{\score{43.9}{0.0}}
& F1
& \score{81.8}{1.5} & \score{83.2}{0.5} & \score{82.5}{1.0} & \score{82.4}{1.1} & \score{82.6}{1.1} & \score{68.8}{0.0} & \score{68.8}{0.0} & \score{79.8}{1.2} & \score{80.1}{1.6} \\
& & & $\Delta$
& \posval{+34.5} & \posval{+35.9} & \posval{+35.2} & \posval{+35.1} & \posval{+35.3} & \posval{+21.5} & \posval{+21.5} & \posval{+32.5} & \posval{+32.8} \\

\multirow{2}{*}{bcnb\_pr}
& \multirow{2}{*}{\score{46.3}{5.0}}
& \multirow{2}{*}{\score{42.7}{0.0}}
& F1
& \score{79.0}{0.8} & \score{78.5}{2.7} & \score{79.1}{1.5} & \score{77.9}{1.8} & \score{78.7}{1.1} & \score{63.6}{0.0} & \score{63.6}{0.0} & \score{76.7}{1.1} & \score{75.7}{0.8} \\
& & & $\Delta$
& \posval{+32.7} & \posval{+32.2} & \posval{+32.8} & \posval{+31.6} & \posval{+32.4} & \posval{+17.3} & \posval{+17.3} & \posval{+30.4} & \posval{+29.4} \\

\multirow{2}{*}{bcnb\_her2}
& \multirow{2}{*}{\score{45.4}{3.9}}
& \multirow{2}{*}{\score{42.5}{0.0}}
& F1
& \score{69.1}{2.9} & \score{69.9}{3.2} & \score{69.8}{5.9} & \score{68.5}{6.0} & \score{73.1}{0.5} & \score{62.8}{0.0} & \score{63.6}{1.5} & \score{73.5}{1.0} & \score{71.2}{0.6} \\
& & & $\Delta$
& \posval{+23.7} & \posval{+24.5} & \posval{+24.4} & \posval{+23.1} & \posval{+27.7} & \posval{+17.4} & \posval{+18.2} & \posval{+28.1} & \posval{+25.8} \\

\multirow{2}{*}{bracs\_coarse}
& \multirow{2}{*}{\score{43.5}{13.2}}
& \multirow{2}{*}{\score{44.0}{1.5}}
& F1
& \score{56.7}{4.4} & \score{61.8}{3.8} & \score{53.8}{1.1} & \score{63.3}{2.2} & \score{52.0}{1.9} & \score{30.2}{11.3} & \score{40.7}{2.8} & \score{64.8}{3.4} & \score{49.4}{3.8} \\
& & & $\Delta$
& \posval{+12.7} & \posval{+17.8} & \posval{+9.8} & \posval{+19.3} & \posval{+8.0} & \negval{-13.8} & \negval{-3.3} & \posval{+20.8} & \posval{+5.4} \\

\multirow{2}{*}{bracs\_fine}
& \multirow{2}{*}{\score{18.9}{0.1}}
& \multirow{2}{*}{\score{18.9}{0.1}}
& F1
& \score{32.4}{4.7} & \score{30.4}{5.8} & \score{38.2}{9.6} & \score{32.8}{7.0} & \score{35.4}{6.4} & \score{22.1}{1.1} & \score{19.7}{1.0} & \score{42.7}{7.9} & \score{25.8}{0.9} \\
& & & $\Delta$
& \posval{+13.5} & \posval{+11.5} & \posval{+19.3} & \posval{+13.9} & \posval{+16.5} & \posval{+3.2} & \posval{+0.8} & \posval{+23.8} & \posval{+6.9} \\

\multirow{2}{*}{cptac\_nsclc}
& \multirow{2}{*}{\score{94.2}{5.1}}
& \multirow{2}{*}{\score{86.5}{8.1}}
& F1
& \score{94.3}{0.6} & \score{96.6}{0.5} & \score{95.8}{0.0} & \score{98.5}{0.6} & \score{96.5}{0.6} & \score{93.4}{2.7} & \score{91.8}{1.7} & \score{93.8}{0.4} & \score{97.0}{1.0} \\
& & & $\Delta$
& \posval{+0.1} & \posval{+2.4} & \posval{+1.6} & \posval{+4.3} & \posval{+2.3} & \negval{-0.8} & \negval{-2.4} & \negval{-0.4} & \posval{+2.8} \\

\multirow{2}{*}{cptac\_pancancer}
& \multirow{2}{*}{\score{91.9}{0.4}}
& \multirow{2}{*}{\score{87.2}{0.4}}
& F1
& \score{93.1}{1.0} & \score{92.8}{1.7} & \score{94.3}{0.6} & \score{93.1}{1.4} & \score{92.6}{0.9} & \score{89.5}{0.4} & \score{90.1}{0.6} & \score{95.2}{0.3} & \score{92.4}{1.0} \\
& & & $\Delta$
& \posval{+1.2} & \posval{+0.9} & \posval{+2.4} & \posval{+1.2} & \posval{+0.7} & \negval{-2.4} & \negval{-1.8} & \posval{+3.3} & \posval{+0.5} \\

\multirow{2}{*}{ebrains\_idh}
& \multirow{2}{*}{\score{54.2}{16.4}}
& \multirow{2}{*}{\score{44.9}{15.4}}
& F1
& \score{91.4}{1.0} & \score{91.5}{2.3} & \score{91.5}{2.3} & \score{92.3}{1.3} & \score{92.3}{1.7} & \score{88.9}{3.6} & \score{90.4}{0.5} & \score{93.6}{0.5} & \score{92.8}{1.4} \\
& & & $\Delta$
& \posval{+37.2} & \posval{+37.3} & \posval{+37.3} & \posval{+38.1} & \posval{+38.1} & \posval{+34.7} & \posval{+36.2} & \posval{+39.4} & \posval{+38.6} \\

\multirow{2}{*}{ebrains\_coarse}
& \multirow{2}{*}{\score{70.7}{4.0}}
& \multirow{2}{*}{\score{52.5}{0.6}}
& F1
& \score{87.8}{0.8} & \score{87.9}{0.7} & \score{89.1}{1.2} & \score{88.2}{1.1} & \score{88.5}{0.5} & \score{83.1}{1.1} & \score{87.0}{2.0} & \score{90.3}{0.6} & \score{88.4}{0.5} \\
& & & $\Delta$
& \posval{+17.1} & \posval{+17.2} & \posval{+18.4} & \posval{+17.5} & \posval{+17.8} & \posval{+12.4} & \posval{+16.3} & \posval{+19.6} & \posval{+17.7} \\

\multirow{2}{*}{ebrains\_fine}
& \multirow{2}{*}{\score{42.8}{20.4}}
& \multirow{2}{*}{\score{40.0}{1.3}}
& F1
& \score{69.9}{1.1} & \score{70.5}{0.8} & \score{72.3}{1.8} & \score{71.3}{2.1} & \score{69.0}{0.8} & \score{62.3}{3.7} & \score{64.1}{1.8} & \score{72.8}{1.1} & \score{69.9}{2.0} \\
& & & $\Delta$
& \posval{+27.1} & \posval{+27.7} & \posval{+29.5} & \posval{+28.5} & \posval{+26.2} & \posval{+19.5} & \posval{+21.3} & \posval{+30.0} & \posval{+27.1} \\

\multirow{2}{*}{kidrare\_coarse}
& \multirow{2}{*}{\score{73.3}{8.4}}
& \multirow{2}{*}{\score{65.9}{2.8}}
& F1
& \score{83.3}{3.3} & \score{86.0}{1.2} & \score{84.8}{1.2} & \score{83.7}{0.6} & \score{76.8}{1.5} & \score{67.2}{2.8} & \score{69.5}{2.0} & \score{86.3}{2.0} & \score{80.5}{2.8} \\
& & & $\Delta$
& \posval{+10.0} & \posval{+12.7} & \posval{+11.5} & \posval{+10.4} & \posval{+3.5} & \negval{-6.1} & \negval{-3.8} & \posval{+13.0} & \posval{+7.2} \\

\multirow{2}{*}{kidrare\_fine}
& \multirow{2}{*}{\score{26.8}{5.5}}
& \multirow{2}{*}{\score{18.9}{0.3}}
& F1
& \score{55.0}{2.7} & \score{58.4}{1.7} & \score{60.0}{2.3} & \score{58.4}{3.1} & \score{56.3}{1.6} & \score{41.8}{4.8} & \score{41.2}{4.0} & \score{60.1}{1.4} & \score{55.1}{2.2} \\
& & & $\Delta$
& \posval{+28.2} & \posval{+31.6} & \posval{+33.2} & \posval{+31.6} & \posval{+29.5} & \posval{+15.0} & \posval{+14.4} & \posval{+33.3} & \posval{+28.3} \\

\multirow{2}{*}{mut\_het\_rcc\_bap1\_mutation}
& \multirow{2}{*}{\score{46.6}{0.0}}
& \multirow{2}{*}{\score{46.6}{0.0}}
& F1
& \score{85.0}{0.5} & \score{85.3}{0.7} & \score{85.1}{0.7} & \score{84.9}{0.4} & \score{83.5}{1.4} & \score{80.9}{0.0} & \score{80.9}{0.0} & \score{84.1}{1.5} & \score{82.8}{1.1} \\
& & & $\Delta$
& \posval{+38.4} & \posval{+38.7} & \posval{+38.5} & \posval{+38.3} & \posval{+36.9} & \posval{+34.3} & \posval{+34.3} & \posval{+37.5} & \posval{+36.2} \\

\multirow{2}{*}{mut\_het\_rcc\_pbrm1\_mutation}
& \multirow{2}{*}{\score{51.0}{10.0}}
& \multirow{2}{*}{\score{42.8}{9.3}}
& F1
& \score{66.2}{1.1} & \score{66.3}{0.6} & \score{62.6}{1.5} & \score{64.0}{2.9} & \score{64.7}{1.0} & \score{57.7}{2.6} & \score{58.3}{4.1} & \score{64.5}{2.4} & \score{63.3}{2.2} \\
& & & $\Delta$
& \posval{+15.2} & \posval{+15.3} & \posval{+11.6} & \posval{+13.0} & \posval{+13.7} & \posval{+6.7} & \posval{+7.3} & \posval{+13.5} & \posval{+12.3} \\

\multirow{2}{*}{mut\_het\_rcc\_setd2\_mutation}
& \multirow{2}{*}{\score{42.8}{1.2}}
& \multirow{2}{*}{\score{42.2}{0.0}}
& F1
& \score{70.3}{1.6} & \score{70.2}{2.3} & \score{71.4}{1.5} & \score{71.9}{1.0} & \score{70.6}{0.9} & \score{60.7}{0.0} & \score{60.6}{0.1} & \score{70.8}{1.0} & \score{62.6}{1.8} \\
& & & $\Delta$
& \posval{+27.5} & \posval{+27.4} & \posval{+28.6} & \posval{+29.1} & \posval{+27.8} & \posval{+17.9} & \posval{+17.8} & \posval{+28.0} & \posval{+19.8} \\

\bottomrule
\end{tabular}%
}
\end{table*}

\newcommand{\baccpos}[1]{\textcolor{green!50!black}{$#1$}}
\newcommand{\baccneg}[1]{\textcolor{red!70!black}{$#1$}}

\clearpage
\subsection{Results of fine tuning}
\label{sec:appendix-fine}

\begin{table*}[h]
\centering
\caption{Balanced accuracy (\%) comparison between scratch-trained MIL models and pretrain-initialized MIL models. Each dataset contains three rows: Scratch, Pretrain, and $\Delta$ (Pretrain--Scratch, percentage points).}
\label{tab:appendix_bacc_scratch_pretrain_delta}
\scriptsize
\resizebox{\linewidth}{!}{%
\begin{tabular}{llccccccccc}
\toprule
Dataset & Setting & ABMIL & CLAM & TransMIL & WiKG & DAGMIL & GDFMIL & 2DMamba & AMDMIL & AEMMIL \\
\midrule
\multirow{3}{*}{\texttt{bcnb\_er}} & Scratch & $74.6_{\scriptscriptstyle(1.2)}$ & $74.9_{\scriptscriptstyle(3.2)}$ & $72.4_{\scriptscriptstyle(3.9)}$ & $74.5_{\scriptscriptstyle(3.0)}$ & $75.9_{\scriptscriptstyle(1.6)}$ & $72.6_{\scriptscriptstyle(3.8)}$ & $74.3_{\scriptscriptstyle(2.4)}$ & $74.2_{\scriptscriptstyle(2.0)}$ & $76.2_{\scriptscriptstyle(1.2)}$ \\
 & Pretrain & $74.8_{\scriptscriptstyle(0.9)}$ & $75.5_{\scriptscriptstyle(2.7)}$ & $75.5_{\scriptscriptstyle(2.1)}$ & $75.8_{\scriptscriptstyle(4.2)}$ & $76.8_{\scriptscriptstyle(2.1)}$ & $75.2_{\scriptscriptstyle(1.7)}$ & $74.2_{\scriptscriptstyle(3.2)}$ & $77.5_{\scriptscriptstyle(4.4)}$ & $76.8_{\scriptscriptstyle(2.4)}$ \\
 & $\Delta$ & \baccpos{+0.2} & \baccpos{+0.6} & \baccpos{+3.2} & \baccpos{+1.2} & \baccpos{+1.0} & \baccpos{+2.6} & \baccneg{-0.1} & \baccpos{+3.2} & \baccpos{+0.6} \\
\midrule
\multirow{3}{*}{\texttt{bcnb\_her2}} & Scratch & $64.5_{\scriptscriptstyle(1.9)}$ & $62.2_{\scriptscriptstyle(2.4)}$ & $55.4_{\scriptscriptstyle(3.3)}$ & $62.7_{\scriptscriptstyle(3.3)}$ & $53.6_{\scriptscriptstyle(4.5)}$ & $65.3_{\scriptscriptstyle(8.2)}$ & $61.4_{\scriptscriptstyle(1.9)}$ & $64.9_{\scriptscriptstyle(1.9)}$ & $65.7_{\scriptscriptstyle(1.9)}$ \\
 & Pretrain & $63.7_{\scriptscriptstyle(2.1)}$ & $65.3_{\scriptscriptstyle(2.3)}$ & $63.5_{\scriptscriptstyle(4.9)}$ & $59.8_{\scriptscriptstyle(3.8)}$ & $66.1_{\scriptscriptstyle(1.6)}$ & $67.2_{\scriptscriptstyle(1.0)}$ & $64.2_{\scriptscriptstyle(0.7)}$ & $66.4_{\scriptscriptstyle(1.1)}$ & $63.0_{\scriptscriptstyle(1.4)}$ \\
 & $\Delta$ & \baccneg{-0.8} & \baccpos{+3.1} & \baccpos{+8.1} & \baccneg{-2.9} & \baccpos{+12.5} & \baccpos{+1.9} & \baccpos{+2.8} & \baccpos{+1.5} & \baccneg{-2.7} \\
\midrule
\multirow{3}{*}{\texttt{bcnb\_pr}} & Scratch & $72.2_{\scriptscriptstyle(1.6)}$ & $72.9_{\scriptscriptstyle(2.1)}$ & $67.9_{\scriptscriptstyle(2.7)}$ & $71.5_{\scriptscriptstyle(1.9)}$ & $72.9_{\scriptscriptstyle(3.1)}$ & $72.1_{\scriptscriptstyle(3.7)}$ & $71.8_{\scriptscriptstyle(2.6)}$ & $72.0_{\scriptscriptstyle(2.3)}$ & $74.9_{\scriptscriptstyle(1.2)}$ \\
 & Pretrain & $73.1_{\scriptscriptstyle(2.3)}$ & $73.1_{\scriptscriptstyle(1.7)}$ & $70.8_{\scriptscriptstyle(3.0)}$ & $72.5_{\scriptscriptstyle(2.8)}$ & $73.5_{\scriptscriptstyle(2.0)}$ & $72.1_{\scriptscriptstyle(1.0)}$ & $73.0_{\scriptscriptstyle(1.9)}$ & $73.8_{\scriptscriptstyle(1.3)}$ & $73.4_{\scriptscriptstyle(1.7)}$ \\
 & $\Delta$ & \baccpos{+0.9} & \baccpos{+0.2} & \baccpos{+2.9} & \baccpos{+1.0} & \baccpos{+0.6} & \baccneg{0.0} & \baccpos{+1.1} & \baccpos{+1.7} & \baccneg{-1.6} \\
\midrule
\multirow{3}{*}{\texttt{bracs\_coarse}} & Scratch & $61.9_{\scriptscriptstyle(5.7)}$ & $64.7_{\scriptscriptstyle(1.5)}$ & $62.3_{\scriptscriptstyle(1.1)}$ & $62.4_{\scriptscriptstyle(7.3)}$ & $67.2_{\scriptscriptstyle(1.9)}$ & $64.1_{\scriptscriptstyle(5.0)}$ & $68.9_{\scriptscriptstyle(3.5)}$ & $67.5_{\scriptscriptstyle(5.7)}$ & $66.2_{\scriptscriptstyle(5.0)}$ \\
 & Pretrain & $66.2_{\scriptscriptstyle(2.2)}$ & $64.5_{\scriptscriptstyle(4.1)}$ & $64.3_{\scriptscriptstyle(3.5)}$ & $70.2_{\scriptscriptstyle(4.2)}$ & $72.8_{\scriptscriptstyle(3.5)}$ & $68.4_{\scriptscriptstyle(8.3)}$ & $73.8_{\scriptscriptstyle(2.7)}$ & $67.9_{\scriptscriptstyle(4.3)}$ & $71.1_{\scriptscriptstyle(1.5)}$ \\
 & $\Delta$ & \baccpos{+4.3} & \baccneg{-0.2} & \baccpos{+2.1} & \baccpos{+7.8} & \baccpos{+5.6} & \baccpos{+4.3} & \baccpos{+4.9} & \baccpos{+0.4} & \baccpos{+5.0} \\
\midrule
\multirow{3}{*}{\texttt{bracs\_fine}} & Scratch & $40.0_{\scriptscriptstyle(6.0)}$ & $41.2_{\scriptscriptstyle(3.2)}$ & $32.0_{\scriptscriptstyle(3.3)}$ & $32.4_{\scriptscriptstyle(6.0)}$ & $43.0_{\scriptscriptstyle(1.2)}$ & $37.4_{\scriptscriptstyle(1.7)}$ & $46.8_{\scriptscriptstyle(1.8)}$ & $43.5_{\scriptscriptstyle(3.7)}$ & $41.8_{\scriptscriptstyle(3.9)}$ \\
 & Pretrain & $42.9_{\scriptscriptstyle(2.6)}$ & $41.4_{\scriptscriptstyle(3.6)}$ & $43.6_{\scriptscriptstyle(3.8)}$ & $35.8_{\scriptscriptstyle(3.7)}$ & $41.3_{\scriptscriptstyle(6.7)}$ & $45.2_{\scriptscriptstyle(3.2)}$ & $45.9_{\scriptscriptstyle(3.1)}$ & $41.4_{\scriptscriptstyle(2.8)}$ & $42.7_{\scriptscriptstyle(2.7)}$ \\
 & $\Delta$ & \baccpos{+3.0} & \baccpos{+0.2} & \baccpos{+11.6} & \baccpos{+3.4} & \baccneg{-1.7} & \baccpos{+7.9} & \baccneg{-0.9} & \baccneg{-2.1} & \baccpos{+0.9} \\
\midrule
\multirow{3}{*}{\texttt{cptac\_nsclc}} & Scratch & $95.0_{\scriptscriptstyle(0.7)}$ & $95.6_{\scriptscriptstyle(1.9)}$ & $96.6_{\scriptscriptstyle(1.5)}$ & $92.0_{\scriptscriptstyle(1.4)}$ & $93.0_{\scriptscriptstyle(1.8)}$ & $95.0_{\scriptscriptstyle(0.9)}$ & $95.0_{\scriptscriptstyle(1.2)}$ & $95.1_{\scriptscriptstyle(1.0)}$ & $96.5_{\scriptscriptstyle(1.5)}$ \\
 & Pretrain & $96.3_{\scriptscriptstyle(1.1)}$ & $94.6_{\scriptscriptstyle(1.1)}$ & $94.9_{\scriptscriptstyle(1.1)}$ & $96.5_{\scriptscriptstyle(1.6)}$ & $97.0_{\scriptscriptstyle(1.1)}$ & $94.9_{\scriptscriptstyle(0.5)}$ & $94.6_{\scriptscriptstyle(1.1)}$ & $95.0_{\scriptscriptstyle(1.3)}$ & $95.1_{\scriptscriptstyle(0.3)}$ \\
 & $\Delta$ & \baccpos{+1.3} & \baccneg{-1.0} & \baccneg{-1.7} & \baccpos{+4.5} & \baccpos{+4.0} & \baccneg{0.0} & \baccneg{-0.4} & \baccneg{-0.2} & \baccneg{-1.4} \\
\midrule
\multirow{3}{*}{\texttt{cptac\_pancancer}} & Scratch & $90.7_{\scriptscriptstyle(0.5)}$ & $91.3_{\scriptscriptstyle(0.4)}$ & $92.4_{\scriptscriptstyle(1.5)}$ & $91.8_{\scriptscriptstyle(0.7)}$ & $91.7_{\scriptscriptstyle(2.9)}$ & $91.5_{\scriptscriptstyle(2.8)}$ & $91.8_{\scriptscriptstyle(0.9)}$ & $94.7_{\scriptscriptstyle(0.9)}$ & $92.6_{\scriptscriptstyle(1.5)}$ \\
 & Pretrain & $91.8_{\scriptscriptstyle(0.9)}$ & $92.1_{\scriptscriptstyle(1.3)}$ & $93.0_{\scriptscriptstyle(1.0)}$ & $92.3_{\scriptscriptstyle(2.4)}$ & $93.7_{\scriptscriptstyle(0.5)}$ & $93.5_{\scriptscriptstyle(0.4)}$ & $91.1_{\scriptscriptstyle(1.5)}$ & $94.1_{\scriptscriptstyle(0.9)}$ & $91.3_{\scriptscriptstyle(1.5)}$ \\
 & $\Delta$ & \baccpos{+1.1} & \baccpos{+0.8} & \baccpos{+0.6} & \baccpos{+0.5} & \baccpos{+2.0} & \baccpos{+2.1} & \baccneg{-0.7} & \baccneg{-0.6} & \baccneg{-1.2} \\
\midrule
\multirow{3}{*}{\texttt{ebrains\_coarse}} & Scratch & $84.5_{\scriptscriptstyle(2.4)}$ & $85.0_{\scriptscriptstyle(1.0)}$ & $83.4_{\scriptscriptstyle(1.9)}$ & $82.9_{\scriptscriptstyle(2.9)}$ & $85.2_{\scriptscriptstyle(2.2)}$ & $83.7_{\scriptscriptstyle(3.4)}$ & $87.2_{\scriptscriptstyle(1.1)}$ & $84.0_{\scriptscriptstyle(2.9)}$ & $87.3_{\scriptscriptstyle(1.2)}$ \\
 & Pretrain & $83.9_{\scriptscriptstyle(2.9)}$ & $84.9_{\scriptscriptstyle(1.1)}$ & $82.7_{\scriptscriptstyle(2.2)}$ & $84.2_{\scriptscriptstyle(1.5)}$ & $84.8_{\scriptscriptstyle(3.2)}$ & $82.1_{\scriptscriptstyle(2.2)}$ & $85.8_{\scriptscriptstyle(2.9)}$ & $85.4_{\scriptscriptstyle(2.5)}$ & $87.5_{\scriptscriptstyle(1.3)}$ \\
 & $\Delta$ & \baccneg{-0.5} & \baccneg{0.0} & \baccneg{-0.7} & \baccpos{+1.3} & \baccneg{-0.3} & \baccneg{-1.6} & \baccneg{-1.4} & \baccpos{+1.4} & \baccpos{+0.2} \\
\midrule
\multirow{3}{*}{\texttt{ebrains\_fine}} & Scratch & $67.0_{\scriptscriptstyle(0.8)}$ & $66.1_{\scriptscriptstyle(0.9)}$ & $64.7_{\scriptscriptstyle(2.7)}$ & $65.1_{\scriptscriptstyle(2.0)}$ & $66.1_{\scriptscriptstyle(1.2)}$ & $63.8_{\scriptscriptstyle(4.0)}$ & $68.3_{\scriptscriptstyle(1.4)}$ & $70.3_{\scriptscriptstyle(1.6)}$ & $67.5_{\scriptscriptstyle(1.8)}$ \\
 & Pretrain & $66.5_{\scriptscriptstyle(1.1)}$ & $67.0_{\scriptscriptstyle(1.4)}$ & $65.5_{\scriptscriptstyle(1.7)}$ & $65.2_{\scriptscriptstyle(4.1)}$ & $64.7_{\scriptscriptstyle(3.9)}$ & $64.5_{\scriptscriptstyle(2.0)}$ & $65.9_{\scriptscriptstyle(1.7)}$ & $67.5_{\scriptscriptstyle(3.2)}$ & $68.9_{\scriptscriptstyle(1.6)}$ \\
 & $\Delta$ & \baccneg{-0.5} & \baccpos{+0.9} & \baccpos{+0.9} & \baccpos{+0.1} & \baccneg{-1.4} & \baccpos{+0.8} & \baccneg{-2.4} & \baccneg{-2.8} & \baccpos{+1.4} \\
\midrule
\multirow{3}{*}{\texttt{ebrains\_idh}} & Scratch & $92.3_{\scriptscriptstyle(1.3)}$ & $93.0_{\scriptscriptstyle(2.0)}$ & $90.7_{\scriptscriptstyle(1.7)}$ & $90.7_{\scriptscriptstyle(2.3)}$ & $90.8_{\scriptscriptstyle(2.6)}$ & $92.2_{\scriptscriptstyle(0.3)}$ & $92.6_{\scriptscriptstyle(1.4)}$ & $91.3_{\scriptscriptstyle(1.2)}$ & $92.3_{\scriptscriptstyle(1.1)}$ \\
 & Pretrain & $92.8_{\scriptscriptstyle(1.9)}$ & $91.3_{\scriptscriptstyle(1.8)}$ & $92.5_{\scriptscriptstyle(1.4)}$ & $91.4_{\scriptscriptstyle(0.9)}$ & $93.5_{\scriptscriptstyle(1.2)}$ & $92.7_{\scriptscriptstyle(1.3)}$ & $93.2_{\scriptscriptstyle(1.4)}$ & $92.8_{\scriptscriptstyle(0.8)}$ & $92.2_{\scriptscriptstyle(1.4)}$ \\
 & $\Delta$ & \baccpos{+0.5} & \baccneg{-1.8} & \baccpos{+1.7} & \baccpos{+0.7} & \baccpos{+2.7} & \baccpos{+0.5} & \baccpos{+0.5} & \baccpos{+1.4} & \baccneg{-0.1} \\
\midrule
\multirow{3}{*}{\texttt{kidrare\_coarse}} & Scratch & $90.1_{\scriptscriptstyle(2.2)}$ & $90.4_{\scriptscriptstyle(1.7)}$ & $91.1_{\scriptscriptstyle(2.2)}$ & $88.0_{\scriptscriptstyle(1.2)}$ & $88.2_{\scriptscriptstyle(1.6)}$ & $89.8_{\scriptscriptstyle(1.6)}$ & $91.2_{\scriptscriptstyle(1.4)}$ & $93.2_{\scriptscriptstyle(0.9)}$ & $88.4_{\scriptscriptstyle(2.4)}$ \\
 & Pretrain & $91.6_{\scriptscriptstyle(1.5)}$ & $88.2_{\scriptscriptstyle(2.8)}$ & $92.5_{\scriptscriptstyle(2.1)}$ & $88.7_{\scriptscriptstyle(1.7)}$ & $87.9_{\scriptscriptstyle(1.1)}$ & $91.5_{\scriptscriptstyle(1.6)}$ & $89.3_{\scriptscriptstyle(1.6)}$ & $93.0_{\scriptscriptstyle(1.2)}$ & $91.7_{\scriptscriptstyle(2.6)}$ \\
 & $\Delta$ & \baccpos{+1.6} & \baccneg{-2.2} & \baccpos{+1.4} & \baccpos{+0.7} & \baccneg{-0.2} & \baccpos{+1.7} & \baccneg{-1.9} & \baccneg{-0.3} & \baccpos{+3.3} \\
\midrule
\multirow{3}{*}{\texttt{kidrare\_fine}} & Scratch & $45.7_{\scriptscriptstyle(2.4)}$ & $46.8_{\scriptscriptstyle(2.0)}$ & $48.9_{\scriptscriptstyle(4.7)}$ & $51.4_{\scriptscriptstyle(7.1)}$ & $51.8_{\scriptscriptstyle(4.5)}$ & $49.5_{\scriptscriptstyle(5.9)}$ & $50.3_{\scriptscriptstyle(5.3)}$ & $57.3_{\scriptscriptstyle(5.1)}$ & $47.7_{\scriptscriptstyle(3.2)}$ \\
 & Pretrain & $47.6_{\scriptscriptstyle(1.8)}$ & $47.5_{\scriptscriptstyle(1.9)}$ & $53.3_{\scriptscriptstyle(2.2)}$ & $48.6_{\scriptscriptstyle(4.7)}$ & $46.8_{\scriptscriptstyle(4.5)}$ & $51.0_{\scriptscriptstyle(5.5)}$ & $52.9_{\scriptscriptstyle(5.9)}$ & $54.1_{\scriptscriptstyle(4.3)}$ & $48.5_{\scriptscriptstyle(3.0)}$ \\
 & $\Delta$ & \baccpos{+1.9} & \baccpos{+0.7} & \baccpos{+4.4} & \baccneg{-2.8} & \baccneg{-5.0} & \baccpos{+1.5} & \baccpos{+2.6} & \baccneg{-3.2} & \baccpos{+0.8} \\
\midrule
\multirow{3}{*}{\texttt{mut\_het\_rcc\_bap1\_mutation}} & Scratch & $63.7_{\scriptscriptstyle(4.2)}$ & $65.0_{\scriptscriptstyle(5.4)}$ & $62.7_{\scriptscriptstyle(4.6)}$ & $64.9_{\scriptscriptstyle(3.3)}$ & $58.8_{\scriptscriptstyle(7.6)}$ & $52.9_{\scriptscriptstyle(3.5)}$ & $66.6_{\scriptscriptstyle(5.9)}$ & $67.1_{\scriptscriptstyle(3.7)}$ & $69.5_{\scriptscriptstyle(6.3)}$ \\
 & Pretrain & $62.3_{\scriptscriptstyle(5.7)}$ & $69.6_{\scriptscriptstyle(4.6)}$ & $70.2_{\scriptscriptstyle(1.7)}$ & $62.6_{\scriptscriptstyle(8.2)}$ & $62.7_{\scriptscriptstyle(6.8)}$ & $62.0_{\scriptscriptstyle(9.3)}$ & $67.1_{\scriptscriptstyle(6.5)}$ & $66.5_{\scriptscriptstyle(5.7)}$ & $67.5_{\scriptscriptstyle(4.4)}$ \\
 & $\Delta$ & \baccneg{-1.5} & \baccpos{+4.6} & \baccpos{+7.5} & \baccneg{-2.2} & \baccpos{+3.9} & \baccpos{+9.1} & \baccpos{+0.5} & \baccneg{-0.6} & \baccneg{-2.0} \\
\midrule
\multirow{3}{*}{\texttt{mut\_het\_rcc\_pbrm1\_mutation}} & Scratch & $62.9_{\scriptscriptstyle(2.7)}$ & $63.7_{\scriptscriptstyle(1.4)}$ & $57.8_{\scriptscriptstyle(3.2)}$ & $64.3_{\scriptscriptstyle(1.7)}$ & $63.6_{\scriptscriptstyle(2.2)}$ & $60.7_{\scriptscriptstyle(1.8)}$ & $64.6_{\scriptscriptstyle(2.1)}$ & $65.8_{\scriptscriptstyle(1.7)}$ & $69.2_{\scriptscriptstyle(0.8)}$ \\
 & Pretrain & $64.5_{\scriptscriptstyle(2.6)}$ & $67.9_{\scriptscriptstyle(2.1)}$ & $63.3_{\scriptscriptstyle(2.9)}$ & $60.5_{\scriptscriptstyle(2.8)}$ & $62.5_{\scriptscriptstyle(1.4)}$ & $62.4_{\scriptscriptstyle(2.9)}$ & $67.9_{\scriptscriptstyle(1.6)}$ & $69.2_{\scriptscriptstyle(1.2)}$ & $68.7_{\scriptscriptstyle(1.8)}$ \\
 & $\Delta$ & \baccpos{+1.6} & \baccpos{+4.3} & \baccpos{+5.5} & \baccneg{-3.8} & \baccneg{-1.0} & \baccpos{+1.7} & \baccpos{+3.3} & \baccpos{+3.4} & \baccneg{-0.5} \\
\midrule
\multirow{3}{*}{\texttt{mut\_het\_rcc\_setd2\_mutation}} & Scratch & $62.5_{\scriptscriptstyle(1.9)}$ & $62.2_{\scriptscriptstyle(2.0)}$ & $57.5_{\scriptscriptstyle(4.1)}$ & $57.6_{\scriptscriptstyle(6.3)}$ & $59.6_{\scriptscriptstyle(4.9)}$ & $63.4_{\scriptscriptstyle(2.5)}$ & $63.5_{\scriptscriptstyle(1.8)}$ & $62.9_{\scriptscriptstyle(3.2)}$ & $62.2_{\scriptscriptstyle(3.9)}$ \\
 & Pretrain & $63.6_{\scriptscriptstyle(2.4)}$ & $63.9_{\scriptscriptstyle(2.6)}$ & $61.2_{\scriptscriptstyle(1.5)}$ & $59.4_{\scriptscriptstyle(7.8)}$ & $57.4_{\scriptscriptstyle(6.4)}$ & $65.3_{\scriptscriptstyle(3.9)}$ & $66.4_{\scriptscriptstyle(2.0)}$ & $63.5_{\scriptscriptstyle(3.0)}$ & $62.9_{\scriptscriptstyle(3.1)}$ \\
 & $\Delta$ & \baccpos{+1.2} & \baccpos{+1.7} & \baccpos{+3.7} & \baccpos{+1.7} & \baccneg{-2.2} & \baccpos{+1.9} & \baccpos{+2.9} & \baccpos{+0.6} & \baccpos{+0.7} \\
\bottomrule
\end{tabular}%
}
\end{table*}


\newcommand{\metricpos}[1]{\textcolor{green!50!black}{$#1$}}
\newcommand{\metricneg}[1]{\textcolor{red!70!black}{$#1$}}

\begin{table*}[h]
\centering
\caption{AUC comparison between scratch-trained MIL models and pretrain-initialized MIL models. Each dataset contains Scratch, Pretrain, and $\Delta$ rows. Values are reported as AUC (\%) with one decimal place.}
\label{tab:appendix_auc_scratch_pretrain_delta}
\scriptsize
\setlength{\tabcolsep}{3pt}
\renewcommand{\arraystretch}{0.88}
\resizebox{\linewidth}{!}{%
\begin{tabular}{llccccccccc}
\toprule
Dataset & Setting & ABMIL & CLAM & TransMIL & WiKG & DAGMIL & GDFMIL & 2DMamba & AMDMIL & AEMMIL \\
\midrule
\multirow{3}{*}{\texttt{bcnb\_er}} & Scratch & $84.8_{\scriptscriptstyle(2.6)}$ & $85.6_{\scriptscriptstyle(2.0)}$ & $84.0_{\scriptscriptstyle(2.0)}$ & $83.7_{\scriptscriptstyle(2.9)}$ & $85.8_{\scriptscriptstyle(1.4)}$ & $84.4_{\scriptscriptstyle(1.1)}$ & $85.8_{\scriptscriptstyle(0.9)}$ & $84.7_{\scriptscriptstyle(1.4)}$ & $86.0_{\scriptscriptstyle(1.2)}$ \\
 & Pretrain & $86.1_{\scriptscriptstyle(1.8)}$ & $84.8_{\scriptscriptstyle(2.1)}$ & $84.2_{\scriptscriptstyle(1.5)}$ & $84.4_{\scriptscriptstyle(2.5)}$ & $85.9_{\scriptscriptstyle(0.6)}$ & $86.3_{\scriptscriptstyle(1.3)}$ & $86.2_{\scriptscriptstyle(1.5)}$ & $86.3_{\scriptscriptstyle(1.9)}$ & $87.0_{\scriptscriptstyle(1.8)}$ \\
 & $\Delta$ & \metricpos{+1.3} & \metricneg{-0.8} & \metricpos{+0.2} & \metricpos{+0.7} & \metricpos{+0.1} & \metricpos{+1.9} & \metricpos{+0.4} & \metricpos{+1.6} & \metricpos{+1.0} \\
\midrule
\multirow{3}{*}{\texttt{bcnb\_her2}} & Scratch & $71.0_{\scriptscriptstyle(1.2)}$ & $70.5_{\scriptscriptstyle(3.5)}$ & $59.1_{\scriptscriptstyle(4.6)}$ & $69.4_{\scriptscriptstyle(2.8)}$ & $65.3_{\scriptscriptstyle(3.2)}$ & $72.4_{\scriptscriptstyle(6.6)}$ & $66.8_{\scriptscriptstyle(1.6)}$ & $72.4_{\scriptscriptstyle(0.6)}$ & $74.6_{\scriptscriptstyle(1.4)}$ \\
 & Pretrain & $70.8_{\scriptscriptstyle(0.5)}$ & $70.9_{\scriptscriptstyle(2.3)}$ & $68.5_{\scriptscriptstyle(4.9)}$ & $68.7_{\scriptscriptstyle(1.9)}$ & $71.8_{\scriptscriptstyle(1.1)}$ & $73.2_{\scriptscriptstyle(0.3)}$ & $69.9_{\scriptscriptstyle(1.6)}$ & $73.7_{\scriptscriptstyle(0.7)}$ & $71.9_{\scriptscriptstyle(2.5)}$ \\
 & $\Delta$ & \metricneg{-0.3} & \metricpos{+0.4} & \metricpos{+9.5} & \metricneg{-0.7} & \metricpos{+6.4} & \metricpos{+0.8} & \metricpos{+3.1} & \metricpos{+1.3} & \metricneg{-2.7} \\
\bottomrule
\end{tabular}%
}
\end{table*}

\begin{table*}[h]
\centering
\caption{F1 comparison between scratch-trained MIL models and pretrain-initialized MIL models. Each dataset contains Scratch, Pretrain, and $\Delta$ rows. Values are reported as F1 (\%) with one decimal place.}
\label{tab:appendix_f1_scratch_pretrain_delta}
\scriptsize
\setlength{\tabcolsep}{3pt}
\renewcommand{\arraystretch}{0.88}
\resizebox{\linewidth}{!}{%
\begin{tabular}{llccccccccc}
\toprule
Dataset & Setting & ABMIL & CLAM & TransMIL & WiKG & DAGMIL & GDFMIL & 2DMamba & AMDMIL & AEMMIL \\
\midrule
\multirow{3}{*}{\texttt{bcnb\_er}} & Scratch & $81.8_{\scriptscriptstyle(2.1)}$ & $82.5_{\scriptscriptstyle(1.8)}$ & $77.9_{\scriptscriptstyle(3.1)}$ & $80.5_{\scriptscriptstyle(3.3)}$ & $78.8_{\scriptscriptstyle(0.8)}$ & $80.0_{\scriptscriptstyle(0.8)}$ & $82.7_{\scriptscriptstyle(0.9)}$ & $81.0_{\scriptscriptstyle(1.2)}$ & $81.0_{\scriptscriptstyle(2.8)}$ \\
 & Pretrain & $81.6_{\scriptscriptstyle(3.1)}$ & $81.2_{\scriptscriptstyle(1.7)}$ & $79.8_{\scriptscriptstyle(2.1)}$ & $79.9_{\scriptscriptstyle(2.2)}$ & $81.8_{\scriptscriptstyle(2.2)}$ & $82.1_{\scriptscriptstyle(1.7)}$ & $82.6_{\scriptscriptstyle(1.5)}$ & $81.4_{\scriptscriptstyle(2.7)}$ & $83.2_{\scriptscriptstyle(1.6)}$ \\
 & $\Delta$ & \metricneg{-0.2} & \metricneg{-1.3} & \metricpos{+1.8} & \metricneg{-0.6} & \metricpos{+3.0} & \metricpos{+2.1} & \metricneg{-0.1} & \metricpos{+0.3} & \metricpos{+2.2} \\
\midrule
\multirow{3}{*}{\texttt{bcnb\_her2}} & Scratch & $72.9_{\scriptscriptstyle(2.0)}$ & $71.5_{\scriptscriptstyle(2.2)}$ & $66.8_{\scriptscriptstyle(2.1)}$ & $70.6_{\scriptscriptstyle(2.4)}$ & $65.9_{\scriptscriptstyle(3.5)}$ & $70.9_{\scriptscriptstyle(5.6)}$ & $65.9_{\scriptscriptstyle(5.0)}$ & $67.2_{\scriptscriptstyle(8.4)}$ & $71.3_{\scriptscriptstyle(2.5)}$ \\
 & Pretrain & $73.5_{\scriptscriptstyle(2.1)}$ & $74.0_{\scriptscriptstyle(1.5)}$ & $72.7_{\scriptscriptstyle(3.0)}$ & $69.0_{\scriptscriptstyle(3.8)}$ & $69.5_{\scriptscriptstyle(4.1)}$ & $73.7_{\scriptscriptstyle(2.0)}$ & $66.5_{\scriptscriptstyle(3.4)}$ & $69.7_{\scriptscriptstyle(3.2)}$ & $69.7_{\scriptscriptstyle(2.6)}$ \\
 & $\Delta$ & \metricpos{+0.6} & \metricpos{+2.4} & \metricpos{+5.9} & \metricneg{-1.5} & \metricpos{+3.5} & \metricpos{+2.8} & \metricpos{+0.6} & \metricpos{+2.5} & \metricneg{-1.6} \\
\bottomrule
\end{tabular}%
}
\end{table*}

\begin{table*}[h]
\centering
\caption{Accuracy comparison between scratch-trained MIL models and pretrain-initialized MIL models. Each dataset contains Scratch, Pretrain, and $\Delta$ rows. Values are reported as Accuracy (\%) with one decimal place.}
\label{tab:appendix_accuracy_scratch_pretrain_delta}
\scriptsize
\setlength{\tabcolsep}{3pt}
\renewcommand{\arraystretch}{0.88}
\resizebox{\linewidth}{!}{%
\begin{tabular}{llccccccccc}
\toprule
Dataset & Setting & ABMIL & CLAM & TransMIL & WiKG & DAGMIL & GDFMIL & 2DMamba & AMDMIL & AEMMIL \\
\midrule
\multirow{3}{*}{\texttt{bcnb\_er}} & Scratch & $81.4_{\scriptscriptstyle(2.6)}$ & $82.4_{\scriptscriptstyle(1.8)}$ & $77.2_{\scriptscriptstyle(4.3)}$ & $79.8_{\scriptscriptstyle(4.0)}$ & $77.5_{\scriptscriptstyle(1.1)}$ & $79.7_{\scriptscriptstyle(1.7)}$ & $82.8_{\scriptscriptstyle(0.9)}$ & $80.6_{\scriptscriptstyle(1.4)}$ & $80.3_{\scriptscriptstyle(3.7)}$ \\
 & Pretrain & $81.2_{\scriptscriptstyle(3.9)}$ & $80.6_{\scriptscriptstyle(2.0)}$ & $78.7_{\scriptscriptstyle(2.7)}$ & $78.8_{\scriptscriptstyle(2.3)}$ & $81.1_{\scriptscriptstyle(3.0)}$ & $81.8_{\scriptscriptstyle(2.4)}$ & $82.7_{\scriptscriptstyle(1.4)}$ & $80.5_{\scriptscriptstyle(2.9)}$ & $82.9_{\scriptscriptstyle(1.9)}$ \\
 & $\Delta$ & \metricneg{-0.2} & \metricneg{-1.8} & \metricpos{+1.5} & \metricneg{-1.0} & \metricpos{+3.7} & \metricpos{+2.1} & \metricneg{-0.1} & \metricneg{-0.1} & \metricpos{+2.6} \\
\midrule
\multirow{3}{*}{\texttt{bcnb\_her2}} & Scratch & $73.1_{\scriptscriptstyle(2.9)}$ & $71.9_{\scriptscriptstyle(2.5)}$ & $68.1_{\scriptscriptstyle(2.0)}$ & $70.9_{\scriptscriptstyle(4.1)}$ & $73.6_{\scriptscriptstyle(2.0)}$ & $72.2_{\scriptscriptstyle(4.6)}$ & $64.5_{\scriptscriptstyle(5.9)}$ & $66.2_{\scriptscriptstyle(8.5)}$ & $70.7_{\scriptscriptstyle(3.3)}$ \\
 & Pretrain & $74.5_{\scriptscriptstyle(2.7)}$ & $74.3_{\scriptscriptstyle(1.6)}$ & $73.6_{\scriptscriptstyle(2.6)}$ & $71.4_{\scriptscriptstyle(5.9)}$ & $68.1_{\scriptscriptstyle(5.1)}$ & $73.3_{\scriptscriptstyle(2.9)}$ & $64.6_{\scriptscriptstyle(4.3)}$ & $68.2_{\scriptscriptstyle(3.8)}$ & $69.0_{\scriptscriptstyle(3.7)}$ \\
 & $\Delta$ & \metricpos{+1.4} & \metricpos{+2.4} & \metricpos{+5.5} & \metricpos{+0.5} & \metricneg{-5.5} & \metricpos{+1.1} & \metricpos{+0.1} & \metricpos{+2.0} & \metricneg{-1.7} \\
\bottomrule
\end{tabular}%
}
\end{table*}

\clearpage
\subsection{Results of few shot}
\label{sec:appendix5}

\newcommand{\bestscore}[2]{\ensuremath{\mathbf{#1}_{\scriptscriptstyle(#2)}}}

\begin{table*}[h]
  \caption{Few-shot comparison between from-scratch training and distillation initialization measured by bacc. Mean and standard deviation are computed across seeds.}
  \label{tab:nshot_bacc}
  \centering
  \scriptsize
  \setlength{\tabcolsep}{3pt}
  \renewcommand{\arraystretch}{0.82}
  \begin{adjustbox}{max width=\textwidth}
  \begin{tabular}{lllccccc}
    \toprule
    Dataset & MIL & Setting & 1-shot & 2-shot & 4-shot & 8-shot & 16-shot \\
    \midrule
    \multirow{16}{*}{BRACS-Coarse} & \multirow{2}{*}{2DMamba} & Scratch & \score{35.1}{3.3} & \score{33.8}{4.5} & \score{41.7}{6.3} & \score{50.4}{8.2} & \score{58.8}{5.7} \\
     &  & Pretrain & \bestscore{47.6}{10.7} & \bestscore{41.7}{8.4} & \bestscore{58.5}{8.9} & \bestscore{64.5}{8.2} & \bestscore{72.9}{5.4} \\
     & \multirow{2}{*}{ABMIL} & Scratch & \score{38.6}{6.8} & \score{39.0}{8.6} & \bestscore{48.8}{10.4} & \score{44.9}{8.2} & \score{55.8}{5.5} \\
     &  & Pretrain & \bestscore{39.1}{7.7} & \bestscore{40.6}{11.3} & \score{47.3}{12.8} & \bestscore{50.7}{10.2} & \bestscore{57.8}{5.0} \\
     & \multirow{2}{*}{AEMMIL} & Scratch & \score{40.7}{8.1} & \score{38.4}{8.1} & \score{46.3}{7.8} & \score{52.6}{12.9} & \score{60.7}{6.9} \\
     &  & Pretrain & \bestscore{45.2}{10.0} & \bestscore{52.5}{16.4} & \bestscore{56.3}{12.6} & \bestscore{60.3}{10.6} & \bestscore{67.9}{5.1} \\
     & \multirow{2}{*}{AMDMIL} & Scratch & \score{41.2}{5.5} & \score{41.9}{6.2} & \score{46.0}{7.1} & \score{55.2}{6.4} & \score{61.8}{4.5} \\
\bottomrule
  \end{tabular}
  \end{adjustbox}
\end{table*}

\begin{table*}[h]
  \caption{Few-shot comparison between from-scratch training and distillation initialization measured by Accuracy. Mean and standard deviation are computed across seeds. DAGMIL is excluded because only scratch results are available, and 32-shot results are omitted.}
  \label{tab:nshot_acc}
  \centering
  \scriptsize
  \setlength{\tabcolsep}{3pt}
  \renewcommand{\arraystretch}{0.82}
  \begin{adjustbox}{max width=\textwidth}
  \begin{tabular}{lllccccc}
    \toprule
    Dataset & MIL & Setting & 1-shot & 2-shot & 4-shot & 8-shot & 16-shot \\
    \midrule
    \multirow{16}{*}{BRACS-Coarse} & \multirow{2}{*}{2DMamba} & Scratch & \score{33.6}{4.1} & \score{35.6}{5.2} & \score{40.5}{6.1} & \score{50.8}{7.7} & \score{61.8}{5.6} \\
     &  & Pretrain & \bestscore{47.8}{9.8} & \bestscore{45.8}{7.6} & \bestscore{57.2}{8.5} & \bestscore{65.1}{8.2} & \bestscore{73.6}{4.4} \\
     & \multirow{2}{*}{ABMIL} & Scratch & \score{37.0}{6.1} & \score{35.9}{10.2} & \bestscore{48.3}{11.6} & \score{44.8}{9.1} & \score{55.4}{6.6} \\
     &  & Pretrain & \bestscore{37.2}{7.9} & \bestscore{38.4}{11.2} & \score{46.4}{12.8} & \bestscore{50.6}{9.6} & \bestscore{58.2}{5.3} \\
     & \multirow{2}{*}{AEMMIL} & Scratch & \score{41.1}{7.1} & \score{39.1}{7.2} & \score{44.6}{10.7} & \score{52.9}{12.2} & \score{62.1}{8.2} \\
     &  & Pretrain & \bestscore{43.5}{9.0} & \bestscore{52.4}{15.9} & \bestscore{54.2}{13.4} & \bestscore{59.1}{11.2} & \bestscore{66.9}{6.3} \\
     & \multirow{2}{*}{AMDMIL} & Scratch & \score{40.0}{3.4} & \score{40.0}{6.6} & \score{43.0}{8.1} & \score{54.9}{10.0} & \score{60.2}{4.7} \\
\bottomrule
  \end{tabular}
  \end{adjustbox}
\end{table*}

\begin{table*}[h]
  \caption{Few-shot comparison between from-scratch training and distillation initialization measured by AUC. Mean and standard deviation are computed across seeds. DAGMIL is excluded because only scratch results are available, and 32-shot results are omitted.}
  \label{tab:nshot_auc}
  \centering
  \scriptsize
  \setlength{\tabcolsep}{3pt}
  \renewcommand{\arraystretch}{0.82}
  \begin{adjustbox}{max width=\textwidth}
  \begin{tabular}{lllccccc}
    \toprule
    Dataset & MIL & Setting & 1-shot & 2-shot & 4-shot & 8-shot & 16-shot \\
    \midrule
    \multirow{16}{*}{BRACS-Coarse} & \multirow{2}{*}{2DMamba} & Scratch & \score{55.1}{4.9} & \score{54.5}{6.0} & \score{60.7}{5.1} & \score{68.9}{8.5} & \score{81.3}{4.8} \\
     &  & Pretrain & \bestscore{65.0}{10.5} & \bestscore{60.3}{9.7} & \bestscore{76.4}{9.4} & \bestscore{81.6}{6.9} & \bestscore{88.8}{2.4} \\
     & \multirow{2}{*}{ABMIL} & Scratch & \bestscore{51.3}{12.5} & \score{50.2}{11.2} & \score{63.0}{14.3} & \score{62.4}{15.0} & \score{74.0}{9.3} \\
     &  & Pretrain & \score{48.9}{13.4} & \bestscore{54.4}{18.6} & \bestscore{65.7}{13.3} & \bestscore{65.1}{14.3} & \bestscore{77.6}{4.2} \\
     & \multirow{2}{*}{AEMMIL} & Scratch & \score{55.3}{9.2} & \score{59.1}{9.2} & \score{64.5}{10.2} & \score{71.2}{11.9} & \score{80.2}{5.6} \\
     &  & Pretrain & \bestscore{64.7}{12.6} & \bestscore{71.7}{16.4} & \bestscore{75.3}{12.7} & \bestscore{79.3}{9.6} & \bestscore{86.7}{3.1} \\
     & \multirow{2}{*}{AMDMIL} & Scratch & \score{60.9}{9.5} & \score{59.5}{9.6} & \score{63.1}{8.4} & \score{74.7}{10.4} & \score{83.3}{3.2} \\
\bottomrule
  \end{tabular}
  \end{adjustbox}
\end{table*}

\begin{table*}[h]
  \caption{Few-shot comparison between from-scratch training and distillation initialization measured by F1. Mean and standard deviation are computed across seeds. DAGMIL is excluded because only scratch results are available, and 32-shot results are omitted.}
  \label{tab:nshot_f1}
  \centering
  \scriptsize
  \setlength{\tabcolsep}{3pt}
  \renewcommand{\arraystretch}{0.82}
  \begin{adjustbox}{max width=\textwidth}
  \begin{tabular}{lllccccc}
    \toprule
    Dataset & MIL & Setting & 1-shot & 2-shot & 4-shot & 8-shot & 16-shot \\
    \midrule
    \multirow{16}{*}{BRACS-Coarse} & \multirow{2}{*}{2DMamba} & Scratch & \score{25.2}{5.6} & \score{26.2}{6.6} & \score{37.8}{8.1} & \score{49.6}{10.2} & \score{59.9}{5.4} \\
     &  & Pretrain & \bestscore{45.7}{13.2} & \bestscore{36.4}{13.6} & \bestscore{57.6}{9.0} & \bestscore{66.0}{8.3} & \bestscore{73.4}{5.3} \\
     & \multirow{2}{*}{ABMIL} & Scratch & \bestscore{30.0}{8.8} & \score{29.7}{13.0} & \bestscore{45.8}{15.6} & \score{42.2}{12.7} & \score{54.2}{8.9} \\
     &  & Pretrain & \score{27.8}{10.9} & \bestscore{32.8}{15.8} & \score{45.4}{14.9} & \bestscore{48.0}{13.5} & \bestscore{58.2}{6.0} \\
     & \multirow{2}{*}{AEMMIL} & Scratch & \score{33.5}{11.8} & \score{32.6}{11.0} & \score{41.3}{16.6} & \score{51.1}{15.1} & \score{60.7}{8.9} \\
     &  & Pretrain & \bestscore{40.0}{12.3} & \bestscore{49.9}{19.7} & \bestscore{53.7}{15.5} & \bestscore{57.9}{13.2} & \bestscore{66.3}{7.9} \\
     & \multirow{2}{*}{AMDMIL} & Scratch & \score{31.2}{7.8} & \score{34.2}{5.8} & \score{36.2}{11.6} & \score{51.0}{12.6} & \score{59.1}{7.3} \\
\bottomrule
  \end{tabular}
  \end{adjustbox}
\end{table*}

\clearpage
\subsection{Results of multi-teacher distillation}
\label{sec:appendixteacher}

\begin{table*}[h]
  \caption{Ablation results of different teacher settings. Mean and standard deviation are computed over five seeds. The best result for each dataset--MIL pair is highlighted in bold.}
  \label{tab:teacher_ablation_all_metrics}
  \centering
  \scriptsize
  \setlength{\tabcolsep}{3pt}
  \renewcommand{\arraystretch}{0.82}
  \begin{adjustbox}{max width=\textwidth}
  \begin{tabular}{lllcccc}
    \toprule
    Dataset & MIL & Teacher & Accuracy & bacc & AUC & F1 \\
    \midrule

    \multirow{12}{*}{BCNB-ER}
    & \multirow{3}{*}{2DMamba}
    & CARE-only & \score{82.0}{2.1} & \textbf{\score{76.6}{1.3}} & \score{86.4}{1.1} & \score{82.5}{1.7} \\
    & & TITAN-only & \score{82.3}{2.2} & \score{74.8}{1.9} & \textbf{\score{86.9}{1.9}} & \score{82.4}{1.6} \\
    & & Both & \textbf{\score{82.7}{1.6}} & \score{74.2}{3.6} & \score{86.2}{1.7} & \textbf{\score{82.6}{1.7}} \\

    & \multirow{3}{*}{CLAM}
    & CARE-only & \textbf{\score{82.8}{2.0}} & \score{75.0}{4.5} & \textbf{\score{85.2}{2.5}} & \textbf{\score{82.8}{2.3}} \\
    & & TITAN-only & \score{81.7}{2.5} & \score{73.9}{3.7} & \score{84.8}{2.9} & \score{81.8}{2.4} \\
    & & Both & \score{80.6}{2.2} & \textbf{\score{75.5}{3.0}} & \score{84.8}{2.4} & \score{81.2}{1.9} \\

    & \multirow{3}{*}{TransMIL}
    & CARE-only & \score{78.9}{2.8} & \score{74.7}{3.6} & \score{83.9}{2.6} & \score{79.8}{2.3} \\
\bottomrule
  \end{tabular}
  \end{adjustbox}
\end{table*}

\clearpage
\subsection{Results of different type distillation loss}
\label{sec:appendixloss}



\begin{table*}[h]
\centering
\caption{
Comparison of different distillation loss types across downstream datasets.
All values are reported in percentage form, with standard deviations shown in the lower-right subscript.
}
\label{tab:losstype_all_metrics}
\tiny
\setlength{\tabcolsep}{2.2pt}
\renewcommand{\arraystretch}{0.88}
\resizebox{\linewidth}{!}{%
\begin{tabular}{llccc@{\hspace{0.8em}}llccc}
\toprule
Dataset & Metric & Angle & Cosine & MSE
& Dataset & Metric & Angle & Cosine & MSE \\
\midrule

\multirow{4}{*}{\texttt{bcnb-er}}
& BACC & \score{76.0}{0.9} & \score{75.0}{2.4} & \score{74.9}{1.3}
& \multirow{4}{*}{\texttt{kidrare-fine}}
& BACC & \score{45.4}{2.0} & \score{45.9}{1.9} & \score{46.9}{1.9} \\
& AUC  & \score{86.0}{1.5} & \score{86.6}{1.8} & \score{86.3}{1.0}
& & AUC  & \score{94.4}{0.4} & \score{94.5}{0.4} & \score{94.7}{0.4} \\
& F1   & \score{82.1}{2.0} & \score{82.0}{3.3} & \score{81.4}{2.9}
& & F1   & \score{64.2}{1.4} & \score{64.8}{1.3} & \score{64.6}{1.7} \\
& Acc  & \score{81.6}{2.6} & \score{81.7}{4.1} & \score{80.9}{3.8}
& & Acc  & \score{66.5}{1.2} & \score{66.7}{0.5} & \score{66.0}{1.2} \\

\midrule

\multirow{4}{*}{\texttt{bcnb-pr}}
& BACC & \score{72.3}{1.0} & \score{72.4}{1.0} & \score{71.8}{2.8}
& \multirow{4}{*}{\texttt{ebrains-coarse}}
& BACC & \score{84.5}{2.1} & \score{86.3}{1.4} & \score{83.5}{2.8} \\
& AUC  & \score{79.8}{1.4} & \score{79.2}{1.3} & \score{78.7}{1.7}
& & AUC  & \score{99.1}{0.0} & \score{99.1}{0.1} & \score{99.1}{0.1} \\
& F1   & \score{79.7}{0.4} & \score{78.8}{0.5} & \score{78.3}{1.4}
& & F1   & \score{90.3}{0.9} & \score{90.9}{1.3} & \score{90.1}{0.6} \\
& Acc  & \score{80.0}{0.7} & \score{78.8}{0.8} & \score{78.3}{1.2}
& & Acc  & \score{90.1}{1.0} & \score{90.9}{1.4} & \score{90.3}{0.4} \\

\midrule

\multirow{4}{*}{\texttt{bcnb-her2}}
& BACC & \score{64.3}{1.3} & \score{64.7}{1.2} & \score{65.6}{0.9}
& \multirow{4}{*}{\texttt{ebrains-fine}}
& BACC & \score{67.4}{1.0} & \score{66.7}{1.5} & \score{67.5}{0.9} \\
& AUC  & \score{71.6}{0.6} & \score{72.5}{1.6} & \score{71.7}{0.3}
& & AUC  & \score{98.4}{0.0} & \score{98.4}{0.1} & \score{98.4}{0.0} \\
& F1   & \score{73.3}{2.0} & \score{73.0}{2.5} & \score{74.2}{2.3}
& & F1   & \score{72.7}{1.3} & \score{72.2}{0.6} & \score{72.0}{0.9} \\
& Acc  & \score{73.9}{2.8} & \score{73.2}{3.3} & \score{74.6}{3.2}
& & Acc  & \score{74.0}{1.0} & \score{73.6}{0.8} & \score{73.7}{0.7} \\

\midrule

\multirow{4}{*}{\texttt{bracs-coarse}}
& BACC & \score{62.3}{1.5} & \score{65.5}{3.5} & \score{66.2}{2.4}
& \multirow{4}{*}{\texttt{ebrains-idh}}
& BACC & \score{91.6}{1.5} & \score{92.1}{1.6} & \score{92.5}{1.7} \\
& AUC  & \score{89.0}{1.8} & \score{88.3}{1.6} & \score{87.8}{1.9}
& & AUC  & \score{97.8}{0.5} & \score{97.9}{0.6} & \score{98.1}{0.5} \\
& F1   & \score{61.0}{1.4} & \score{65.6}{5.0} & \score{67.0}{3.3}
& & F1   & \score{91.8}{1.3} & \score{92.5}{1.6} & \score{92.7}{1.5} \\
& Acc  & \score{67.8}{1.8} & \score{70.3}{2.7} & \score{70.8}{1.9}
& & Acc  & \score{91.8}{1.3} & \score{92.5}{1.6} & \score{92.7}{1.5} \\

\midrule

\multirow{4}{*}{\texttt{bracs-fine}}
& BACC & \score{40.9}{3.6} & \score{42.2}{2.9} & \score{41.9}{3.0}
& \multirow{4}{*}{\texttt{mut-het-rcc-bap1}}
& BACC & \score{62.5}{8.1} & \score{64.9}{4.8} & \score{61.1}{7.1} \\
& AUC  & \score{81.4}{2.4} & \score{82.7}{2.0} & \score{81.8}{2.0}
& & AUC  & \score{83.3}{1.3} & \score{82.6}{1.3} & \score{82.5}{1.0} \\
& F1   & \score{43.1}{3.7} & \score{43.2}{2.2} & \score{43.3}{3.9}
& & F1   & \score{83.6}{1.2} & \score{83.6}{1.1} & \score{83.2}{1.1} \\
& Acc  & \score{47.4}{3.0} & \score{48.3}{2.9} & \score{48.5}{2.7}
& & Acc  & \score{84.7}{1.5} & \score{83.4}{1.7} & \score{84.6}{1.9} \\

\midrule

\multirow{4}{*}{\texttt{cptac-nsclc}}
& BACC & \score{96.3}{1.0} & \score{96.8}{1.4} & \score{95.6}{0.6}
& \multirow{4}{*}{\texttt{mut-het-rcc-pbrm1}}
& BACC & \score{66.1}{1.9} & \score{65.0}{2.4} & \score{64.6}{1.6} \\
& AUC  & \score{99.5}{0.8} & \score{99.5}{0.7} & \score{99.2}{0.9}
& & AUC  & \score{70.6}{2.4} & \score{69.9}{3.0} & \score{70.4}{2.3} \\
& F1   & \score{96.3}{1.0} & \score{96.8}{1.3} & \score{95.7}{0.6}
& & F1   & \score{66.1}{2.0} & \score{65.0}{2.4} & \score{64.6}{1.6} \\
& Acc  & \score{96.3}{1.0} & \score{96.8}{1.3} & \score{95.7}{0.6}
& & Acc  & \score{66.2}{1.9} & \score{65.1}{2.3} & \score{64.6}{1.6} \\

\midrule

\multirow{4}{*}{\texttt{cptac-pancancer}}
& BACC & \score{92.6}{0.3} & \score{91.1}{1.2} & \score{92.2}{0.9}
& \multirow{4}{*}{\texttt{mut-het-rcc-setd2}}
& BACC & \score{65.0}{3.3} & \score{63.4}{2.9} & \score{64.4}{2.4} \\
& AUC  & \score{99.7}{0.0} & \score{99.6}{0.1} & \score{99.7}{0.0}
& & AUC  & \score{71.6}{1.8} & \score{71.0}{1.0} & \score{71.8}{0.9} \\
& F1   & \score{93.7}{0.5} & \score{92.3}{1.2} & \score{93.4}{0.9}
& & F1   & \score{70.6}{3.6} & \score{69.7}{4.1} & \score{70.1}{5.6} \\
& Acc  & \score{93.7}{0.5} & \score{92.3}{1.3} & \score{93.5}{0.9}
& & Acc  & \score{70.5}{4.4} & \score{69.8}{5.1} & \score{70.2}{6.6} \\

\midrule

\multirow{4}{*}{\texttt{kidrare-coarse}}
& BACC & \score{92.2}{2.3} & \score{91.0}{1.4} & \score{92.0}{2.6}
& & & & & \\
& AUC  & \score{99.0}{0.2} & \score{98.8}{0.4} & \score{99.0}{0.2}
& & & & & \\
& F1   & \score{93.1}{1.9} & \score{92.8}{1.1} & \score{93.1}{1.9}
& & & & & \\
& Acc  & \score{93.1}{1.9} & \score{92.8}{1.1} & \score{93.1}{1.8}
& & & & & \\

\bottomrule
\end{tabular}%
}
\end{table*}
\clearpage


\end{document}